\documentclass{clv3}

\makeatletter
\def\@fnsymbol#1{\ensuremath{\ifcase#1\or *\or 1\or 2\or
   3\or 4\or \|\or **\or \dagger\dagger
   \or \ddagger\ddagger \else\@ctrerr\fi}}
\makeatother

\usepackage{hyperref}
\usepackage{xcolor}
\usepackage{times}
\usepackage{latexsym}
\usepackage{booktabs}
\usepackage{subcaption}
\usepackage{graphicx}
\usepackage{color,soul}
\usepackage{amsmath,amssymb}
\usepackage{multicol}
\usepackage{multirow}
\usepackage[normalem]{ulem}
\usepackage{pifont}% http://ctan.org/pkg/pifont
\usepackage{todonotes}
\usepackage{tikz-dependency}
\usepackage{diagbox}
\usepackage{array}
\newcolumntype{H}{@{}>{\lrbox0}l<{\endlrbox}}

\definecolor{darkblue}{rgb}{0, 0, 0.5}
\hypersetup{colorlinks=true,citecolor=darkblue, linkcolor=darkblue, urlcolor=darkblue}

\usepackage{lipsum}

\newcommand{\reals}{\mathbb{R}}

\issue{1}{1}{2016}

\dochead{}

\runningtitle{Linguistic Representations in NMT}

\runningauthor{Belinkov, Durrani et al.}

\begin{document}

\title{On the Linguistic Representational Power of Neural Machine Translation Models}

% Alternatives for title

% 1)  \title{How rich are the Neural MT representations? \\ A linguistic perspective}

\author{Yonatan Belinkov\thanks{Authors contributed equally.}\thanks{Computer Science and Artificial Intelligence Laboratory, MIT, Cambridge, MA 02139, USA}\thanks{John F.\ Paulson School of Engineering and Applied Sciences, Harvard University, Cambridge, MA 02138, USA}}
\affil{Massachusetts Institute of Technology \\ Harvard University}

\author{Nadir Durrani\textsuperscript{*}\thanks{Qatar Computing Research Institute, HBKU Research Complex, Doha 5825, Qatar}}
\affil{Qatar Computing Research Institute}

\author{Fahim Dalvi\textsuperscript{3}}
\affil{Qatar Computing Research Institute}

\author{Hassan Sajjad\textsuperscript{3}}
\affil{Qatar Computing Research Institute}

%\author{Llu\'{i}s M\`{a}rquez}
%\affil{QCRI}

\author{James Glass\textsuperscript{1}}
\affil{Massachusetts Institute of Technology}

\historydates{Submission received:            30 November 2018;
revised version received:               21 July 2019;
accepted for publication:               17 September 2019.}

\maketitle

\begin{abstract}

Despite the recent success of deep neural networks in %the field of 
natural language processing (NLP) and other spheres of artificial intelligence (AI), their interpretability remains a challenge. %and their ``black-box'' nature presents a significant barrier in understanding and predicting model's behavior. 
%There have been several efforts to debug the process and to diagnose what is learned within intermediate representations of these models. 
%In this paper, we 
We analyze the representations learned by neural machine translation (NMT) models at various levels of granularity and %empirically 
evaluate their quality %of the representations 
through relevant extrinsic properties. %, %that are 
%important for the task of machine translation. 
In particular, we seek answers to the following questions: (i) How accurately is \textbf{word-structure} captured within the learned representations, %a property that is especially important when 
which is an important aspect in translating %languages with rich 
\textbf{morphologically-rich} languages? (ii) %if the models are able to 
Do the representations capture long-range dependencies, and effectively handle  \textbf{syntactically divergent} languages? (iii) Do the representations capture lexical \textbf{semantics}? %and how does the model map a sequence of subword units to a meaning representation? 
We conduct a thorough investigation along several parameters: (i) Which layers in the architecture capture each of these linguistic phenomena; (ii) How does the choice of translation unit (word, character, or subword unit) impact the linguistic properties captured by the underlying representations? (iii) Do the encoder and decoder learn differently and independently? (iv) Do the representations learned by multilingual NMT models capture the same amount of linguistic information as their bilingual counterparts?
%and v) how robust they are towards noise, such as spelling errors in the input.
%State-of-the-art NMT systems address data-sparsity using subword or character units. However, 
%All previous work on interpreting deep NLP models, analyzed word-based representations only. We perform a qualitative comparison of the representations learned %when the NMT models are trained using words/character or BPE-based subword units, in terms of morphology, syntax and semantics. We additionally analyze how robust these representations are towards noise, such as spelling errors in the input. 
%Our methodology is simple but effective. We generate feature representations from different components of a trained model and use the activations to train an auxiliary classification task (e.g. morphological tagging). The quality of the trained classifier is considered a proxy for judging the quality of the representations, with respect to the auxiliary task.  
Our data-driven, quantitative evaluation illuminates important aspects in NMT models and their ability to capture various linguistic phenomena. We show that deep NMT models trained in an end-to-end fashion, without being provided any direct supervision during the %initial 
training process, learn a non-trivial amount of linguistic information. Notable findings include the following observations:
i) Word morphology and part-of-speech information are captured at the lower layers of the model; (ii) In contrast, %global %high level 
%properties such as 
lexical semantics or non-local syntactic and semantic dependencies are better represented at the higher layers of the model; (iii) Representations learned using characters are more informed about word-morphology compared to those learned using subword units; and (iv) Representations learned by multilingual models are richer compared to  bilingual models. %\todo{YB: I'm wondering if we should add more observations to be more complete, or alternatively not give any specific observations in the abstract. ND: I think it would be sufficient to highlight only important results. But feel free to add any other finding that you think is interesting enough}  %; and (iv) and are more robust in sparse %and noisy data conditions, v) subword units provide a good balance between handling word-morphology (like character units) and capturing non-local dependencies (like word units).

\end{abstract}

\section{Introduction}
\label{sec:intro}

Deep neural networks (DNNs) have quickly become the predominant approach to most tasks in artificial intelligence (AI), including machine translation (MT). Compared to their traditional counterparts, these models are trained in an end-to-end fashion, providing a simple yet elegant mechanism. This simplicity, however, comes at the price of opaqueness. Unlike traditional systems that contain specialized modules carrying specific sub-tasks, neural MT (NMT) systems train one large network, optimized towards the overall task. For example, non-neural statistical MT systems %\todo{YB: I don't like the distinction b/w statistical MT and neural MT, because neural is also statistical. So I added ``non-neural'' here.}
have sub-components to handle fluency~\cite{kenlm}, lexical generation~\cite{Koehn:2003:SPT}, word reordering~\cite{galley-manning:2008:EMNLP,durrani-etal-2015-operation}, rich morphology~\cite{D07-1091,durrani-EtAl:2014:Coling}, and a smorgasbord of features \cite{N09-1025} for modeling different phenomena. Neural MT systems, on the other hand, contain a single model based on an encoder-decoder mechanism \cite{sutskever2014sequence}  with attention~\cite{bahdanau2014neural}. Despite its simplicity, neural MT surpassed non-neural statistical MT within a few years of its emergence. Human evaluation and error analysis revealed that the improvements were obtained through more fluent outputs \cite{toral-sanchezcartagena:2017:EACLlong} and better handling of morphology and non-local dependencies \cite{bentivogli-EtAl:2016:EMNLP2016}. However, it is not clear what the role of different components in the network is, what kind of information is learned during the training process, and how different components interact. Consequently, MT systems trained using neural networks are often thought of as a ``black-box''---i.e., they map inputs to outputs, but the internal machinery is opaque and difficult to interpret. Gaining a better understanding of these systems is necessary for improving the design choices and performance. In current practice, their development is often limited to a trial-and-error process, without gaining a real understanding of what the system has learned. We aim to increase model transparency by analyzing the representations learned by NMT models at different levels of granularity in light of various linguistic phenomena---at morphological, syntactic, and semantic levels---that are considered important for the task of machine translation and for learning complex NLP problems. We thus strive for post-hoc decomposability, in the sense of \citet{lipton2016mythos}. That is, we analyze models after they have been trained, to uncover what linguistic phenomena %(for example word morphology)
are captured within the underlying representations. More specifically, we aim to address the following questions in this paper:

\begin{itemize}
\item What linguistic information is captured in deep learning models? 
\begin{itemize}
\item Do the NMT representations capture word morphology?
\item Do the NMT models, being trained on flat sequences of words, still acquire structural %syntactic 
information? 
\item Do the NMT models learn informative semantic representations?
\end{itemize}
\item Is the language information well distributed across the network or are designated parts (different layers, encoder vs.\ decoder)  more focused on a particular linguistic property?
\item What impact does the choice of translation unit (characters, subword units, or words) have on the learned representations in terms of different linguistic phenomena? %, and when handling %noisy inputs or 
%out-of-vocabulary words?
\item How does translating into different target languages affect the representations on the (encoder) source-side? % What is the effect of varying the target language on source-side (encoder) representations? 
\item How do the representations acquired by multilingual %NMT
models compare with those acquired by bilingual models?
\end{itemize}

To this end, we follow a simple and effective procedure with three steps: (i) train an NMT system; (ii) use the trained model to generate feature representations for source/target language words; and (iii) train a classifier using the generated features to make predictions for  a relevant auxiliary task. We then evaluate the quality of the trained classifier on the given task as a proxy to the quality of the trained NMT model. In this way, we obtain a quantitative measure of how well the original NMT system learns features that are relevant to the given task. 
This procedure has become common for analyzing various neural NLP models~\cite{belinkov2019analysis}.
In this work, 
we analyze NMT representations through several linguistic annotation tasks:
part-of-speech (POS) tagging and morphological tagging for morphological
knowledge; CCG supertagging and syntactic dependency labeling for syntactic
knowledge; lexical semantic tagging and semantic dependency labelling for
semantic knowledge.
%We use %part-of-speech (POS) tagging, 
%morphological, syntactic, and semantic tagging %and semantic tagging 
%to evaluate word-level properties. We also use syntactic and semantic dependency labeling to evaluate compositional information.

We experiment with several languages with varying degrees of morphological richness and syntactic divergence (compared to English): French, German, Czech, Russian, Arabic, and Hebrew. Our analyses reveal interesting insights such as:
\begin{itemize}
\item NMT models trained in an end-to-end fashion learn a non-trivial amount of linguistic information without being provided with direct supervision during the initial training process.
\item Linguistic information tends to be organized in a modular manner, whereby different parts of the neural network generate representations with varying amounts and types of linguistic properties.
\item  A hierarchy of language representations emerges in networks trained
on the complex tasks studied in this paper. The lower layers of the network focus on local, low-level linguistic properties (morphology, parts-of-speech, local relations), while higher layers are more concerned with global, high level
properties (lexical semantics, long-range relations).
\item Character-based representations are better for learning morphology, especially for unknown and low-frequency %and noisy 
input words. In contrast, representations learned using subword units are better for handling syntactic and semantic dependencies. %long-distance dependencies. %The two are fairly similar in predicting semantics.  
\item The target language impacts the kind of information learned by the MT system. For example, translating into morphologically-poorer languages leads to better source-side word representations. This effect is especially apparent in smaller data regimes. 
\item Representations learned by multilingual NMT models are richer in terms of learning different linguistic phenomena and benefit from shared learning.\footnote{The learned parameters are implicitly shared by all the language pairs being
modeled}%\todo{YB: ``benefit from shared learning'' is not clear}
\end{itemize}

%This paper extends our previously published work \cite{belinkov:2017:acl, belinkov:2017:ijcnlp,dalvi:2017:ijcnlp} in the following ways: (i) Previous work on analyzing NMT representations has been limited to the analysis of word representations only, where there is a one-to-one mapping from translation units (words) and their NMT representations (hidden states) to their linguistic annotations (e.g., part-of-speech tags). In the present work, we conduct an experimental investigation and a comparison of the representations learned based on word, character, and subword units in NMT systems. (ii) We move beyond word-level properties to study syntactic and semantic dependency relations, representing important compositional properties.  (iii) Previous research  has shown that a single multilingual model can be trained to achieve comparable performance to the individual models \cite{TACL1081}. We analyze representations extracted from multilingual NMT systems using the specified linguistic properties.

This article is organized into the following sections. Section \ref{sec:relatedWork} provides an account of the related work. Section \ref{sec:properties} describes the linguistic properties and the representative tasks used to carry out the analysis study.  Section \ref{sec:methodology} describes the methodology taken for analyzing the NMT representations. Section \ref{sec:expSetup} describes data, annotations and, experimental details. Sections \ref{sec:results-morphology}, \ref{sec:results-syntax}, and \ref{sec:results-semantics}  provide empirical results and analysis to evaluate the quality of NMT representations with respect to morphology, syntax, and semantics, respectively, and Section \ref{sec:multilingual} does the same for the multilingual NMT models. Section \ref{sec:discussion} sheds light on the overall patterns that arise from the experimental results from several angles. Section \ref{sec:conclusion} concludes the paper.
An open-source implementation of our analysis code is available through the NeuroX toolkit \cite{neuroX:aaai19:demo}. 
%at \url{}. 
%\todo{YB: add links to the code used ND: Fahim can you address this?}

%\newpage 

\section{Related Work}
\label{sec:relatedWork}

The work related to this paper can be divided into several groups:

\subsection{Analysis of Neural Networks}

The first group of related work aims at demystifying what information is learned within the neural network black-box. One line of work visualizes hidden unit activations in recurrent neural networks (RNNs) that are trained for a given task~\cite{elman1991distributed,karpathy2015visualizing,kadar2016representation}. While such visualizations illuminate the inner workings of the network, they are often qualitative in nature and somewhat anecdotal. 
Other work aims to evaluate systems on specific linguistic phenomena represented in so-called challenge sets. 
Prominent examples include older work on MT evaluation~\cite{C90-2037}, %,isahara1995jeida,koh2001test}, 
as well as more recent evaluations via contrastive translation pairs~\cite{sennrich:2017:EACLshort,W17-4705,W17-4702,N18-1118}. The latter line of work constructs minimal pairs of translations that differ by a known linguistic property, and evaluates whether the MT system  assigns a higher score to the correct translation. 
The challenge set evaluation may produce informative results on the quality of the overall model for some linguistic property, but it does not directly assess the learned representations.

A different approach tries to provide a quantitative analysis by correlating parts of the neural network with linguistic properties, for example by training a classifier to predict a feature of interest~\cite{adi2016fine,hupkes2017visualisation,conneau2018you}. Such an analysis has been conducted on word embeddings~\cite{kohn:2015:EMNLP,qian-qiu-huang:2016:P16-11}, sentence embeddings \cite{adi2016fine,Ganesh:2017:IST:3110025.3110083,conneau2018you}, and RNN states ~\cite{qian-qiu-huang:2016:EMNLP2016,wu2016investigating,wang2017gate}. The language properties mainly analyzed are morphological \cite{qian-qiu-huang:2016:P16-11,vylomova2016word,belinkov:2017:acl, dalvi:2017:ijcnlp}, semantic \cite{qian-qiu-huang:2016:P16-11,belinkov:2017:ijcnlp} and syntactic \cite{tran2018importance,kohn:2015:EMNLP,
conneau2018you}. Recent studies carried a more fine-grained neuron-level analysis for NMT and LM~\cite{dalvi:2019:AAAI,indivdualneuron:iclr,lakretz2019emergence}.
In contrast to all of this work, we focus on the representations learned in neural machine translation in  light of various linguistic properties (morphological, syntactic, and semantic) and phenomena such as handling low frequency words. %\alert{and noisy inputs} <-- left out in this paper. 
Our work is most similar to \namecite{shi-padhi-knight:2016:EMNLP2016} and \namecite{vylomova2016word}. The former used hidden vectors from a neural MT encoder to predict syntactic properties on the English source side, whereas we study multiple language properties in different languages.  \namecite{vylomova2016word} analyzed different representations for morphologically rich languages in MT, but they did not directly measure the quality of the learned representations.
Surveying the %increasingly more popular 
work on analyzing  neural networks in NLP is beyond the scope of the present paper.  We have highlighted here several of the more relevant studies and refer to \citet{belinkov2019analysis} for a recent survey on the topic.

\subsection{Subword Units}
%\todo{should we add our ACL short paper? -- lets do it in the camera-ready}

One of the major challenges in training NMT systems is  handling less frequent and out-of-vocabulary words. 
To address this issue, researchers have resorted to using subword units for training the neural network models.  
%The second group of work focuses on addressing the limitations of the word-based approach by exploring subword units. 
\namecite{luong2016achieving} trained a hybrid system that integrates character-level representation within a word-based framework. \namecite{lingTDB15} used a bidirectional long short-term memory network~\cite[LSTM;][]{hochreiter1997long} to compose word embeddings from the character embeddings. \namecite{costajussa-fonollosa:2016:P16-2} and \namecite{renduchintala2018character} combined convolutional and
highway layers to replace the standard lookup-based word representations  in NMT systems with character-aware representations.\footnote{Character-based systems have been used previously in phrase-based MT for handling morphologically-rich \cite{Luong:D10-1015} and closely related language pairs \cite{durrani-EtAl:2010:ACL,Nakov:Tiedemann:2012} or for transliterating unknown words \cite{durrani-EtAl:2014:EACL}.}
\namecite{sennrich-haddow-birch:2016:P16-12} used {\it byte-pair encoding} (BPE), a data-compression algorithm, to segment words into smaller units. A variant of this method known as a \emph{wordpiece} model is used by Google \cite{google-nmt:2016}.
\namecite{shapiro2018bpe} used a similar convolutional architecture on top of BPE. 
\namecite{chung-cho-bengio:2016:P16-1} used a combination of BPE-based encoder and character-based decoder to improve translation quality. Motivated by their findings, \namecite{TACL1051} explored using fully character representations (with no word boundaries) on both the source and target sides. As BPE segmentation is not linguistically motivated, an alternative of using morpheme-based segmentation has been explored in \namecite{bradbury-socher:2016:WMT}. It is important to address what using different translation units (word, BPE, morpheme, character) entails. \namecite{sennrich:2017:EACLshort} performed a comparative evaluation of character- and BPE-based systems on carefully crafted synthetic tests and found that character-based models are effective in handling unknown words, but perform worse in capturing long-distance dependencies. Our work contributes to this body of research by analyzing how models based on different units capture various linguistic properties. We analyze the representations obtained by training systems on word, character, and BPE-based units.

%\subsection{Robustness of Neural Models}

%The last group of related work deals with assessing the robustness of  neural models with respect to noise. Recent work has shown that a small amount of perturbations can cause a significant deterioration in the performance of machine learning models \cite{szegedy2013intriguing,Goodfellow:ICLR}, including  NLP models 
%\cite{papernot2016crafting,samanta2017towards,liang2017deep,ebrahimi2017hotflip,gao2018black,jia-liang:2017:EMNLP2017}. Most relevant to the present paper, \namecite{heigold2017robust} and \namecite{belinkovNoise:ICLR} evaluated the robustness of character-based NMT on noisy data. They both found a degradation in translation quality with noisier texts. 
%However,  these studies did not evaluate systems based on different translation units in a comparable setup: \namecite{belinkovNoise:ICLR} used competition-level systems with different architectures and training data, while \namecite{heigold2017robust} kept word segmentation and added a charLSTM %an LSTM on characters 
%to generate word representations. In contrast, we use the same architecture and training data for all systems. More importantly, we analyze what linguistic information is captured by systems with different translation units. 

\section{Linguistic Properties}
\label{sec:properties}

%Once the NMT models have been trained, our goal is to analyze the representations against various linguistic phenomenon that are considered important for the task of machine translation, and that we believe are intrinsically learned in the model to effectively perform the complex task of translation. We explore three such linguistic phenomenon:

In this section, we describe the linguistic phenomena for which we analyze  NMT representations. We focus on linguistic properties that are considered important for the task of machine translation, and that we believe are intrinsically learned in the model to effectively perform the complex task of translation. We consider properties from the realms of morphology, syntax, and semantics. %In each case, we define a supervised tagging task that aims to capture the relevant property. 
In each case, we describe linguistic properties of interest and define relevant classification tasks that aim to capture them.

\begin{table*}[t]
	\centering
	%\footnotesize
    % The CCG supertags are taken from \citep{W17-4707}; POS and semantic tags are our own annotation.}
	\resizebox{\columnwidth}{!}{
	\begin{tabular}{c|cccccccc}
		Words & Obama  & receives & Netanyahu  & in  & the & capital & of  & USA \\ %& reading \\
		\midrule
		POS  & NP & VBZ & NP & IN & DT & NN & IN & NP \\ % & VBG \\
		SEM  & PER & ENS & PER & REL & DEF & REL & REL & GEO \\ %& EXS \\ 
		CCG & NP & ((S[dcl]$\backslash$NP)/PP)/NP & NP  & PP/NP  & NP/N  & N  & (NP$\backslash$NP)/NP  & NP
        \\
        \midrule
        %\midrule
        Words & Obama  & empf{\"a}ngt & Netanyahu  & in  & der & Hauptstadt & der  & USA \\ %& reading \\
        \midrule
        MORPH & nn.nom.sg.neut & vvfin.3.sg.pres.ind & ne.nom.sg.* & appr.-- & art.dat.sg.fem & nn.dat.sg.fem & art.gen.pl.* & ne.gen.pl.* \\
   %     MORPH & NN.Nom.Sg.Neut & VVFIN.3.Sg.Pres.Ind & NE.Nom.Sg.* & APPR.-- & ART.Dat.Sg.Fem & NN.Dat.Sg.Fem & ADJD.Pos & NE.Dat.Pl.* \\
		\bottomrule
	\end{tabular}
	}
    \caption{Example sentence with different word-level annotations. The CCG supertags are taken from \citet{W17-4707}. POS and semantic tags are our own annotation, as well as the German translation and its morphological tags.}
	\label{tab:example-annotation}
\end{table*}

\begin{figure}[t]
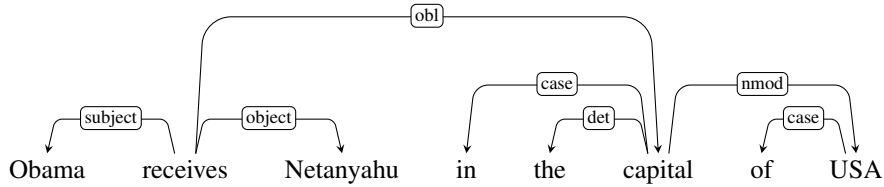
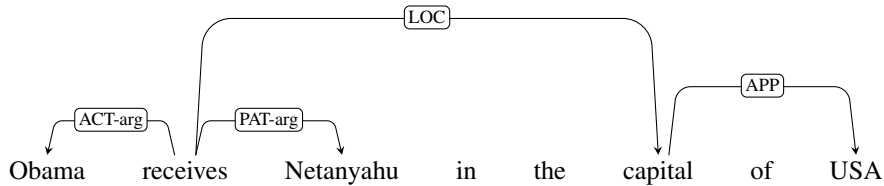

\centering
  \begin{subfigure}{\textwidth}
  \centering
    \begin{dependency}[theme=default]%[arc edge, text only label, label style={above}]
      \begin{deptext}[column sep=.6cm, font=\normalsize]
      Obama \& receives \& Netanyahu \& in \& the \& capital \& of \& USA \\
      \end{deptext}
      \depedge{2}{1}{subject} %nsubj}
      \depedge{2}{3}{object}
      \depedge{6}{4}{case}% mark}
      \depedge{6}{5}{det} %obj}
      \depedge{2}{6}{obl} %conj}
      \depedge{8}{7}{case}% cc}
      \depedge{6}{8}{nmod}
    \end{dependency}
  \caption{Syntactic relations according to the Universal Dependencies formalism. Here ``Obama'' and ``Netanyahu'' are the subject and object of ``receives'', respectively, \texttt{obl} refers to an oblique relation of the locative modifier, \texttt{nmod} denotes the genitive relation, the prepositions ``in'' and ``of'' are treated as case-marking elements, and ``the'' is a determiner. See \url{https://universaldependencies.org/guidelines.html} for detailed definitions. }
  \label{fig:syn}  
  \end{subfigure}
  \begin{subfigure}{\textwidth}
  \centering
    \begin{dependency}[theme=default]%[arc edge, text only label, label style={above}]
      \begin{deptext}[column sep=.6cm, font=\normalsize]
      Obama \& receives \& Netanyahu \& in \& the \& capital \& of \& USA \\
      \end{deptext}
      \depedge{2}{1}{ACT-arg} 
      \depedge{2}{3}{PAT-arg}
      %\depedge{4}{6}{case}% mark}
      %\depedge{5}{6}{det} %obj}
      \depedge{2}{6}{LOC} %conj}
      %\depedge{7}{8}{case}% cc}
      \depedge{6}{8}{APP}
    \end{dependency}
  \caption{Semantic relations according to the PSD formalism. Here \texttt{ACT-arg} and \texttt{PAT-arg} refer respectively to the originator and affected arguments of ``receives'', \texttt{LOC} is the location, and \texttt{APP} is the thing that ``capital'' belongs to. For detailed definitions, see \citet{tectogrammatical-reference-book}. }
  \label{fig:sem}  
  \end{subfigure}  
  \caption{Example sentence with syntactic and semantic relations. }
  \label{fig:syn-sem}
\end{figure}

\begin{table*}[t]
\centering
%\footnotesize
\begin{tabular}{c|c}
\toprule
Words & Professor admits to shooting his girlfriend \\
\midrule
BPE  & Professor admits to sho@@ oting his gir@@ l@@ friend \\
\midrule
Morfessor  & Professor admit@@ s to shoot@@ ing his girl@@ friend  \\
\midrule
Characters  & P r o f e s s o r \_ a d m i t s \_ t o \_ s h o o t i n g \_ h i s \_ g i r l f r i e n d \\
\bottomrule
\end{tabular}
\caption{Example sentence with different segmentations: words, byte-pair encoding (BPE) subwords~\cite{sennrich-haddow-birch:2016:P16-12}, Morfessor-based subwords~\cite{smit-EtAl:2014:Demos}, and characters. Notice that BPE subwords do not necessarily conform to morphemes (``shooting'' $\rightarrow$ ``sho@@'' and ``oting''), while Morfessor tends to have a more morphological segmentation (``shoot@@'', ``ing'').  ``@@''  indicates a split subword unit and ``\_'' marks a word boundary.}
\label{tab:example-segmentation}
%\vspace{-5pt}
\end{table*}

\subsection{Morphology}

Modeling the structure of words and their relationship with other words in the sentence is a fundamental task in any NLP application. Languages vary in the way they encode information within words. Some languages exhibit grammatical relations such as subject/object/predicate or gender agreement by only changing the word form, others achieve the same through word order or addition of particles.  Morphology (aka word structure), poses an exigent problem in machine translation and is at the heart of dealing with the challenge of data-sparsity. While English is limited in morphology, other languages such as Czech, Arabic, and Russian have highly inflected morphology. This entails that for each lemma many possible word variants could exist, thus causing an out-of-vocabulary word problem. For example, \namecite{E17-2059} found only one morphological variant of the Czech word ``\v c\"e\u ska" (plural of English ``kneecap") in a corpus of 50K parallel sentences. It required 50M sentences, a size of parallel corpus only available for a handful of language pairs, for them to observe all possible variants of the word. Even if such a dataset is available, the computational complexity requires NMT systems to limit the vocabulary size. It is therefore important for an MT system to model word-structure with the available data and vocabulary size limitation. In traditional statistical machine translation, this is often addressed by splitting tokens in morphologically-rich languages into constituents in a preprocessing step, using word segmentation in Arabic \cite{PASHA14.593.L14-1479, abdelali-EtAl:2016:N16-3} or compound splitting in German \cite{E03-1076}. Previous work also explored generative morphological models, known as \emph{Factored Translation Models}, that explicitly integrate additional linguistic markup at the word level to learn morphology \cite{D07-1091}. In  NMT training, using subword units such as byte-pair encoding \cite{sennrich-haddow-birch:2016:P16-12} has become a \emph{de facto} standard in training competition grade systems \cite{sennrich-EtAl:2017:WMT,pinnis-EtAl:2017:WMT}. A few have tried morpheme-based segmentation \cite{bradbury-socher:2016:WMT}, and several even used character-based systems \cite{chung-cho-bengio:2016:P16-1,TACL1051} %and have achieved 
to achieve similar performance %to 
as the BPE-segmented systems. 

 Table~\ref{tab:example-segmentation} shows an example of each representation unit. \emph{BPE} splits words into symbols (a symbol is a sequence of characters) and then iteratively replaces the most frequent sequences of symbols with a new merged symbol. In essence, frequent character $n$-gram sequences %will be merged 
merge to form one symbol. The number of merge operations is controlled by a hyper-parameter {\tt OP}, which directly affects the granularity of segmentation:
a high value of {\tt OP} means coarse segmentation and a low value means fine-grained segmentation. %For \emph{morphologically segmented units}, we use an unsupervised morphological segmenter, Morfessor. 
Note that although BPE and Morfessor (unsupervised morpheme-based segmentation) segment words at a similar level of granularity, the segmentation generated by Morfessor \cite{smit-EtAl:2014:Demos} is linguistically motivated. For example, it splits the gerund verb \emph{shooting} into the base verb \emph{shoot} and the suffix \emph{ing}. In comparison the BPE segmentation \emph{sho} + \emph{oting} has no linguistic justification. At the extreme, the fully \emph{character-level} units treat each word as a sequence of characters. %\todo{YB: are we going to mention charCNN results at all? ND: No, i think we should just do with char results}

\paragraph{Tagging tasks}
In this paper, we study how effective are neural MT representations in learning word-morphology and what  different translation units offer in this regard. To answer such questions, we focus on the tasks of part-of-speech (POS) and full morphological tagging, which is the identification of all pertinent morphological features for every word. See Table \ref{tab:example-annotation}. For example,  the morphological tag {\tt vvfin.3.sg.pres.ind} for the word ``empf{\"a}ngt'' (English ``receives'') marks that it is a finite verb, third person, singular gender, present tense, and indicative mood.

%\todo{YB: add example of morphological tagging to table \ref{tab:example-annotation} and refer already here. ND: We don't have Morph tags for English }

\subsection{Syntax}
\label{sec:syntax}

Linguistic theories argue that words are hierarchically organized in syntactic constituents referred to as syntactic trees. It is therefore natural to think that translation models should be 
based on trees rather than a flat sequence representation of sentences. For more than a decade of research in machine translation,  a tremendous amount of effort has been put into syntax-based machine translation \cite{P02-1039,Chiang:2005,galley06:acl,zhang2007tree,shen2010string,P14-2024}, with notable success in languages such as Chinese and German, which are syntactically divergent compared to English. However, the sequence-to-sequence NMT systems were able to surpass the performance of the state-of-the-art syntax-based systems in recent MT competitions \cite{bojar-EtAl:2016:WMT1}. %\todo{YB: maybe add refs on SOTA NMT on these pairs} 
The LSTM-based RNN model with the help of the attention mechanism is able to handle long-distance dependencies. %\todo{YB: too strong claim? ND: I think we are fine} 
There have also been recent attempts to integrate syntax into NMT \cite{P17-2021,D17-1304,P16-1078,P16-2049,P17-1065}, but sequence-to-sequence NMT models without explicit syntax are the state-of-the-art at the moment \cite{sennrich-EtAl:2017:WMT,pinnis-EtAl:2017:WMT}. 

\paragraph{Tagging tasks}
In this paper, we analyze if NMT models trained on flat sequences acquire structural syntactic information. %\sout{and language hierarchy}. 
To answer this, we use two tagging tasks. First, we 
use combinatory categorial grammar (CCG) supertagging, which captures global syntactic information locally at the word level by assigning a label to each word annotating its syntactic role in the sentence. The process is almost equivalent to parsing \cite{J99-2004}. For example, 
the syntactic tag {\tt PP/NP} (in Table \ref{tab:example-annotation}) can be thought of as a function that takes a noun phrase on the right (``the capital of USA'') and returns a prepositional phrase (``in the capital of USA'').\footnote{Refer to  \citet{doi:10.1002/9781444395037.ch5} and \citet{C04-1041} for more information on CCG supertagging.}
%Table \ref{tab:example-annotation} shows an example sentence with CCG supertags. 

Second, we use syntactic dependency labeling, the task of assigning a  type to each arc in a syntactic dependency tree. 
In dependency grammar, sentence structure is represented by a labeled directed graph whose vertices are words and
whose edges are relations, or dependencies, between the words \cite{melvcuk1988dependency,NivreDGA05}. A dependency
is a directed bi-lexical relation between a head and its dependent, or modifier. Dependency structures are attractive to study for three main reasons. First, dependency
formalisms have become increasingly popular in NLP in recent years, and much work has
been devoted to developing large annotated datasets for these formalisms. The Universal Dependencies dataset \cite{UD-11234/1-1983} that is used in this paper has been especially influential. Second, there is a fairly rich history of using dependency structures in machine translation, although much work has focused on using constituency structures \cite{williams2016syntax}. Third, as dependencies are bi-lexical relations between words, it is straightforward to obtain representations for them from an NMT model. This makes them amenable to the general methodology followed in this paper. 
Figure~\ref{fig:syn} shows an example sentence with syntactic dependencies.

%\todo{this paragraph can perhaps be removed and its content integrated into the preceding subsections} Table \ref{tab:example-annotation} shows an example sentence with annotations of each task. The morphological tags capture word structure, semantic tags show semantic property, and syntax tags (CCG super tags) captures global syntactic information locally at the lexical level.  For example in Table \ref{tab:example-annotation}, -- the morphological tag {\tt VBZ} for the word ``receives'', marks that it is a verb with non-third person singular present property, the semantic tag {\tt ENS} describes a {\tt present simple event} category, and the syntactic tag {\tt S[dcl]$\backslash$NP)/NP} indicates that the preposition ``in" attaches to the verb. 

%\alert{ND: Yonatan, can you provide Figures 4-1 and 4-2 of your thesis?} \alert{YB: it's in the file I shared with you, but I'm not sure 4-1 is needed. I added syntactic dependencies of the example sentence instead.}

\subsection{Semantics} 
\label{sec:semantics}

The holy grail in machine translation has long been to achieve an interlingua-based translation model, where the goal is to capture the meaning of the source sentence and generate a target sentence with the same meaning. It is believed since the inception of machine translation that without acquiring such meaning representations it will be impossible to generate human like translations \cite{weaver1955translation}. Traditional statistical MT systems are  
weak at capturing meaning representations (e.g., ``who does what
to whom'', i.e., what are the agent, the action, and the patient in the sentence \cite{C12-1083}). 
Although neural MT systems are also trained only on parallel data, without providing any direct supervision of word meaning, they are a continuous space model, and are believed to capture word meaning. \namecite{johnson2016google}, for example, %visually demonstrated 
found preliminary evidence that the shared architecture in their multilingual NMT systems learns a universal interlingua. 
There have also been some recent efforts to incorporate such information in NMT systems, either explicitly \cite{W17-4702} or implicitly \cite{liu2018handling}. 

\paragraph{Tagging task}

In this paper, we study how semantic information
is captured in NMT through two tasks: lexical semantic tagging and semantic dependency labeling. 
First, we utilize the lexical semantic (SEM) tagging task introduced by \namecite{bjerva-plank-bos:2016:COLING}. It is a sequence labeling task: given a sentence, the goal is to assign to each word a tag representing a semantic class. This is a good task to use as a starting point for investigating semantics because: (i) tagging words with semantic labels is very simple, compared to building complex relational semantic structures; (ii) it provides a large supervised dataset to train on, in contrast to most of the available datasets on word sense disambiguation, lexical substitution, and lexical similarity; and (iii) the proposed SEM tagging task is an abstraction over part-of-speech (POS) tagging\footnote{For instance, proximal and distal demonstratives (e.g.,
``this'' and ``that'') are typically assigned the same POS tag (\texttt{DT}) but receive different SEM tags
({\tt PRX} and {\tt DST}, respectively), and proper nouns are disambiguated into several classes such
as geo-political entity, location, organization, person, and artifact.} aimed at being language-neutral, and oriented to multilingual semantic parsing, all relevant aspects to machine translation.
Table~\ref{tab:example-annotation} shows an example sentence annotated with SEM tags. The semantic tag {\tt ENS} describes a present-simple event category.

The second semantic task is semantic dependency labeling, the task of assigning a type to each arc in a semantic dependency graph. Such dependencies are also known as predicate-argument relations, and may be seen as a first step towards semantic structure. They capture different aspects from syntactic relations, as can be noticed by the different graph structure (compare Figure~\ref{fig:sem} to Figure~\ref{fig:syn}). Predicate-argument relations have also been used in many (non-neural) MT systems~\cite{P12-1095,I11-1004,komachi2006phrase,N13-1060}.  
Figure~\ref{fig:sem} shows an example sentence annotated with Prague Semantic Dependencies (PSD), a reduction of the tectogrammatical annotation in the Prague Czech-English dependency treebank~\cite{tectogrammatical-reference-book,biblio:CiToTectogrammaticalAnnotation2009}, which was made available as part of the Semantic Dependency Parsing shared tasks in SemEval~\cite{S14-2008,S15-2153}.

\section{Methodology}
\label{sec:methodology}

We follow a 3-step process for studying linguistic information learned by the trained neural
MT systems. The steps include: (i) training a neural MT system; (ii) using the trained model to generate feature representations for words in a language of interest; and (iii) training a classifier using generated features to make predictions for the different linguistic tagging tasks. 
The quality of the trained classifier on the given task serves as a proxy to the quality of the generated representations. It thus provides a quantitative measure of how well the original MT system learns features that are relevant to the given task. 

%\subsection{Our Approach}

\begin{figure}[t]
	\centering
	\includegraphics[width=0.85\linewidth]{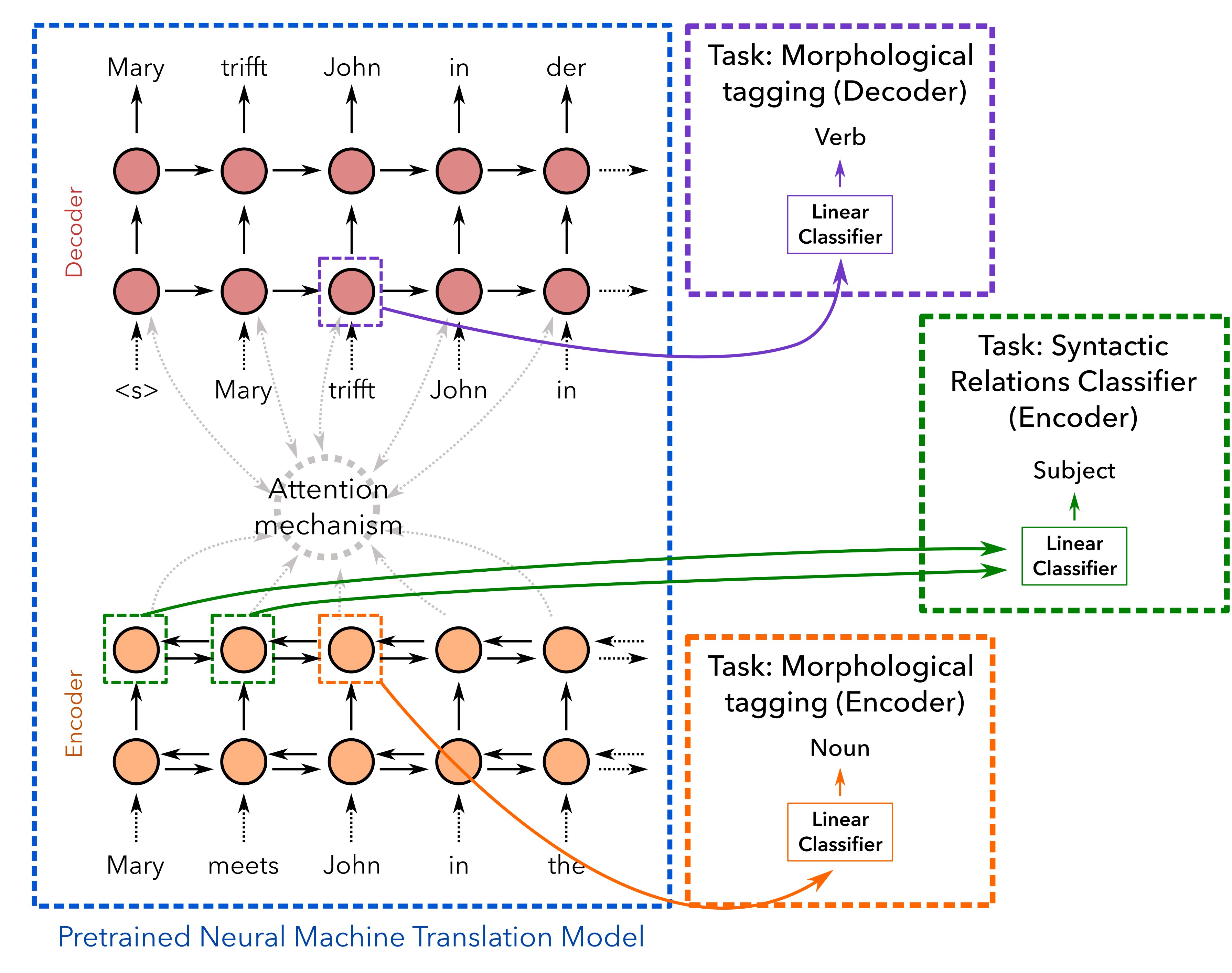}
	\caption{Illustration of our approach: After training an NMT system on parallel data, we extract activations as features from the encoder/decoder and use these along with the labels to train an external classifier. For Mmrphological tagging, we consider the activations for the word alone, while for syntactic/semantic relations we concatenate the activations for the two words involved in the relation.} 
	\label{fig:approach-encoder}
\end{figure}

In this work, we focus on neural MT systems trained using the sequence-to-sequence with attention architecture \cite{bahdanau2014neural}, where an \textit{encoder} network first encodes the source sentence, followed by an \textit{attention} mechanism to compute a weighted average of the encoder states which the \textit{decoder} network uses to generate the target sentence. Both the \textit{encoder} and the \textit{decoder} networks are recurrent neural networks in our case. Several other architectures, for example the Transformer models \cite{NIPS2017_7181} have recently been proposed for neural MT. %but the sequence-to-sequence model still remains a dominant architecture for this task. 
We discuss %some of the other architectures
these briefly in Section \ref{sec:nmt-arch}. 

Formally, let $s = \{s_1, s_2, ..., s_N\}$ denote a source sentence, and  $t=\{t_1, t_2, ..., t_M\}$ denote a target sentence, where $s_i$ and $t_i$ are words. We first describe the simple case where we have word-level model and linguistic properties. Later we extend this scenario to subword units and to linguistic properties that involve multiple words. 

We first use the encoder (Equation\ \ref{eq:enc}) to compute a set of hidden states $h = \{h_1, h_2, ..., h_N\}$, where $h_i$ represents the hidden state for word $s_i$. The encoder is a stacked LSTM with $L$ layers, where the output of layer $l-1$ is passed as input to layer $l$ (at each timestep). We then use an attention mechanism to compute a weighted average of these hidden states from the previous decoder state ($d_{i-1}$), known as the context vector $c_i$ (Equation\ \ref{eq:attn}). The context vector is a real valued vector of $k$ dimensions, which is set to be the same as the hidden states in our case. The attention model computes a weight $w_{h_i}$ for each hidden state of the encoder, thus giving soft alignment for each target word. The context vector is then used by the decoder (Equation \ \ref{eq:dec}), which is also a stacked LSTM, to generate the next word in the target sequence: 

\small
\begin{align} 
&\texttt{ENC}_{s_i}: s_i, e_{i-1} \mapsto h_i \hphantom{.......................................} (1\leq i \leq N) \label{eq:enc} \\
&\texttt{ATTN}_i: \{h_1, ..., h_N\}, d_{i-1},t_{i-1} \mapsto c_i \in \reals^k \hphantom{.......}  (1\leq i \leq M) \label{eq:attn} \\
&\texttt{DEC}_{t_i} : c_i, d_{i-1}, t_{i-1} \mapsto t_{i} \hphantom{................................} (1\leq i \leq M) \label{eq:dec}
\end{align}
\normalsize

% \begin{figure}[t]
% 	\centering
% 	\includegraphics[width=0.50\linewidth]{approach-decoder.png}
% 	\caption{Features for the word \emph{Nun} ($\texttt{DEC}_{t_1}$) are extracted from the decoder of a pre-trained NMT system and provided to the classifier for training} 
% 	\label{fig:approach-decoder}
% \end{figure}

\noindent After training the NMT system, we freeze the parameters of the network and use the encoder or the decoder as a feature extractor to generate vectors representing words in the sentence. Let $\texttt{ENC}_{s_i}^{l}$ denote the representation of a source word $s_i$ at layer $l$ in our stacked LSTM.  We use $\texttt{ENC}_{s_i}^{l}$ from a particular layer $l$ or concatenate all layer representations to train the external classifier for predicting a linguistic tag for $s_i$. The quality of the representation can be deduced from our ability to train a good classifier. For word representations on the target side, we feed our word of interest $t_i$ as the previously predicted word, and extract the representation $\texttt{DEC}_{t_i}$ %from the higher layers 
(see Figure \ref{fig:approach-encoder} for illustration). %Note that in the decoder, the target word representations $\texttt{DEC}_{t_i}$ are not learned for predicting the word $t_i$, but the next word ($t_{i+1}$). 
%\todo{YB: consider adding here symbols for different layers}
%\todo{YB: somewhere we should say that we usually use the top layer hidden states, unless we investigate layer-wise performance, and then we mention explicitly which layer. ND: we now have layer-wise performance}

\paragraph{Generating representations for dependency labeling}

We used dependency structures to evaluate the syntactic and semantic quality of the learned NMT representations (See Section \ref{sec:syntax} and \ref{sec:semantics} for details). Given two words that are known
to participate in a relation, a classifier is trained to predict the relation type. 
%The classifier is the same as before, that is, a one-hidden layer neural network. But 
For the relation labeling task, the input to the classifier is a concatenation
of encoder representations for two words in a relation, %$h^{s,l}_{i}$ 
$\texttt{ENC}_{s_i}^{l}$ and $\texttt{ENC}_{s_j}^{l}$,%\todo{YB: are these meant to be source side, $l$-th layer states? This needs to be defined earlier} 
where ($s_i$, $s_j$) is a known dependency pair with head $s_i$ and modifier $s_j$. Again, we perform experiments with both representations from a particular layer $l$ and the concatenated representation from all layers. Note that this formulation assumes that the order of the dependency is known.
%Here we use a concatenation of the representations of two words, as an input to the classifier, to predict the relation tag. 
This formulation can be seen as a dependency labeling problem, where dependency labels are predicted independently. While limited in scope, this formulation captures a basic notion of structural relations between words.\footnote{It is also not unrealistic, as dependency parsers often work in two stages, first predicting an unlabeled dependency tree, and then labeling its edges \cite{J11-1007,W06-2932}. More complicated formulations can be conceived, from predicting the existence of dependencies independently to solving the full parsing task, but dependency labeling is a simple basic task to begin with.} 

\paragraph{Generating representations with subword and character units}

\begin{figure}[t]
	\centering
	\includegraphics[width=0.85\linewidth]{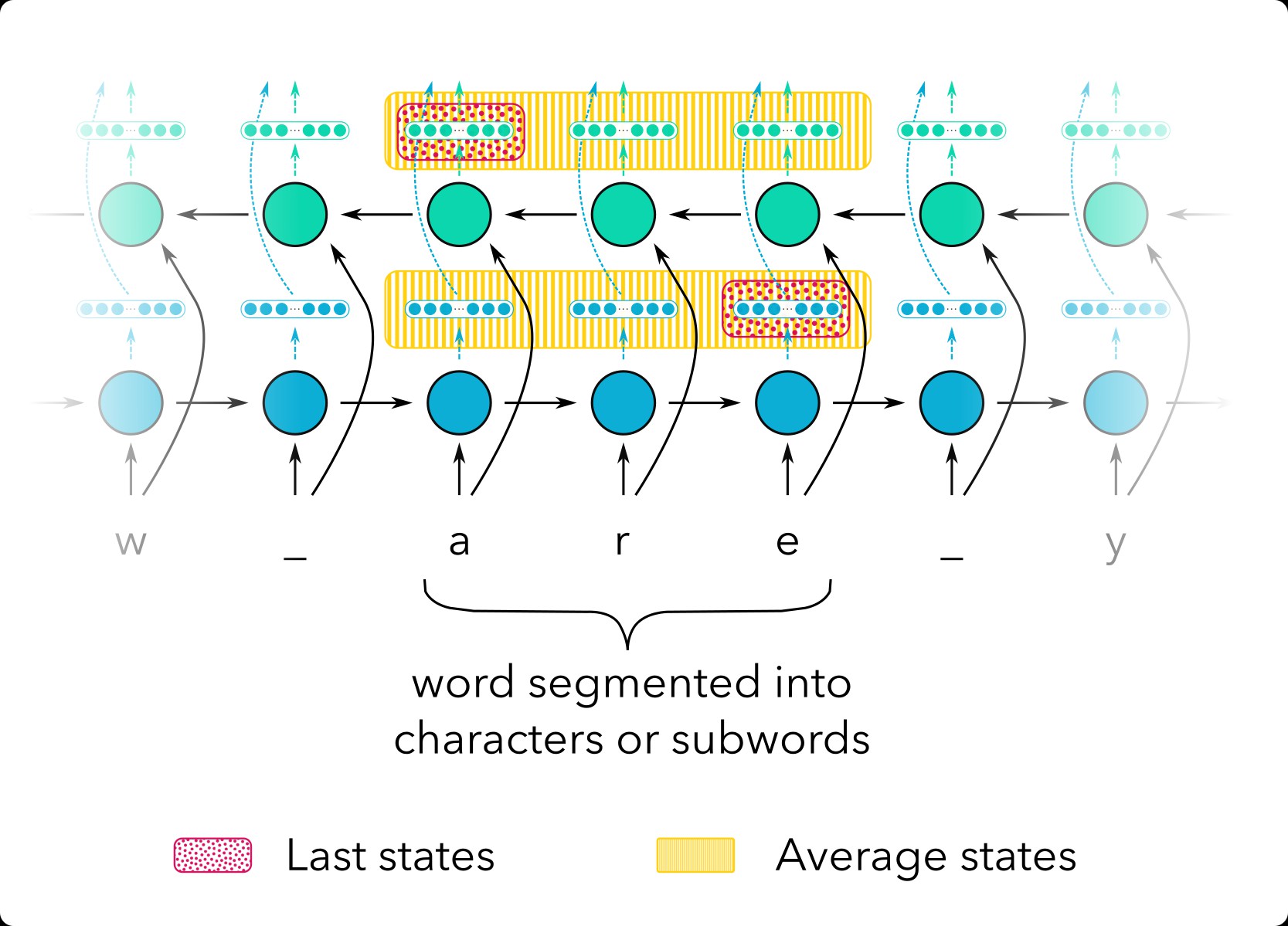}
	\caption{Illustration of a bidirectional layer. Representations from the forward and backward layers are concatenated. For the \textit{average} method, all of the hidden states corresponding to subwords or characters of a given word are averaged together for each layer. For the \textit{last} method, only the hidden state of the final subword or character is considered.}
	\label{fig:extraction-methods}
\end{figure}

Previous work on analyzing NMT representations has been limited to the analysis of word representations only, where there is a one-to-one mapping from translation units (words) and their NMT representations (hidden states) to their linguistic annotations (e.g., part-of-speech tags).\footnote{Although we studied representations from a charCNN \cite{kim2015character} in \namecite{belinkov:2017:acl}, the extracted features were still based on word representations produced by the charCNN.  As a result, in that work we could not analyze and compare subword and character-based models that do not assume a segmentation into words.}  In the case of character- or BPE-based systems, each word is split into multiple translation units, and each unit has its own representation. It is less trivial to define which representations should be evaluated when predicting a linguistic property such as the part-of-speech.   In this work, we consider two simple approximations, illustrated in Figure \ref{fig:extraction-methods}:

\begin{enumerate}[label=\emph{\roman*}),itemsep=0pt]
	\item %i) 
	\textbf{Average}: for every source (or target) word, %we 
	average the activation values of all the subwords (or characters) comprising it. In the case of a bidirectional encoder, we concatenate the averages from the forward and backward encoders' activations on the subwords (or characters) that represent the current word.\footnote{One could envision more sophisticated averages, such as weighting via an attention mechanism.}
	\item %ii) 
	\textbf{Last}: 
	%we 
	consider the activation of the last subword (or character) as the representation of the word. For the bidirectional encoder,  concatenate the forward encoder's activation on the last subword unit with the backward encoder's activation on the first subword unit. % (See Figure \ref{fig:extraction-methods}).   
\end{enumerate}
This formalization allows us to analyze the quality of character- and subword-based representations via prediction tasks, which has not been explored before.

\section{Experimental Setup}
\label{sec:expSetup}

\subsection{NMT Training Data} \label{sec:nmt-data}

We experiment with several languages with varying degrees of morphological richness and syntactic divergence (compared to English): Spanish (es), French (fr), German (de), Czech (cs), Arabic (ar), Russian (ru), and Hebrew (he). We trained NMT systems using data made available by the two popular machine translation campaigns, namely, WMT \cite{bojar-EtAl:2017:WMT1} and IWSLT \cite{iwslt-2016}. The MT models were trained using a concatenation of NEWS, TED and Europarl training data (\mbox{$\approx$ 2.5M} sentence pairs). The multilingual systems were trained by simply concatenating data from different language pairs (a total of $\approx$10M sentence pairs) and training a shared encoder-decoder pipeline. We used German, French, Spanish, and Czech to/from English to train multilingual systems. Language codes were added as prefixes before each sentence. We used official TED test sets to report translation quality \cite{P02-1040}.  %In some of our experiments, 
We also used the fully-aligned United Nations corpus \cite{ZIEMSKI16.1195} for training the models in some of our experiments. It includes six languages: Arabic, Chinese, English, French, Spanish, and Russian. This dataset has the benefit of multiple alignment of the several languages, which allows for comparable cross-linguistic analysis, for example studying the effect of only changing the target language. We used the first 2 million sentences of the training set, using the official training/development/test split.

\subsection{Neural MT Systems}
\vspace{2mm}
\subsubsection{Preprocessing}

We used standard Moses \cite{koehn2007moses} pipeline for preprocessing the data, which includes tokenization, filtering for length, and true-casing. %We used Farasa \cite{abdelali-EtAl:2016:N16-3} for segmenting Arabic. 
The systems were trained with a a maximum sentence length of 80 words. For the BPE systems, we used a vocabulary size using 50,000 operations. In the case of multilingual systems, we used 90,000 operations. For the character-based systems, we simply split the words into characters.\footnote{We also explored charCNN \cite{kim2015character,costajussa-fonollosa:2016:P16-2} models in our preliminary experiments, and found the charCNN variant to perform poorly, compared to the simple char-based LSTM model both in translation quality and when comparing classifier accuracy. Therefore, we decided to leave them out for brevity. See Appendix for the results.} We used Morfessor \cite{smit-EtAl:2014:Demos} with default settings to get morpheme-segmented data. %\todo{YB: this is the first time morpheme-segmented is mentioned. If it's a central point, it should be mentioned earlier with the BPE subword units. ND: Fixed} 
The subword (BPE and Morfessor) and character-based systems were trained with a maximum sentence length of 100, 100--120 and 400--550 units respectively.\footnote{The sentence length was varied across different configurations, to keep the training data sizes the same for all systems.} 

\subsubsection{Model Training}

We used the \texttt{seq2seq-attn} implementation \cite{kim2016} with the following default settings: word embeddings and LSTM states with 500 dimensions, initialized with default {\tt Torch} initialization, SGD with an initial learning rate of $1.0$ and decay rate of $0.5$ (after the 9th epoch), and dropout rate of $0.3$. We used 2--4 hidden layers for both the encoder and the decoder. The NMT system was trained for 20 epochs, and the model with the best validation loss was used for generating features for the external classifier. %We also tried a character convolution neural network (charCNN) model with a highway network over characters \cite{kim2015character} with 1000 feature maps and a kernel width of 6 characters. 
%We used a vocabulary size of 50000 on both the source and target side. For the multilingual settings, we used a shared vocabulary of 90,000. 
These are the settings that we have generally used for the experiments reported in this paper. We will explicitly mention in the individual sections where we digress from the prescribed settings.

\subsection{Classifier Settings}

The classifier is a logistic regression model whose input is either hidden states in word-based models, or \textbf{Last} or \textbf{Average} representations in character- and subword-based models. Since we concatenate forward and backward states from all layers, this ends up being 4000/2000 dimensions when classifying the encoder/decoder: 500 dimensions$\times$4 layers$\times$2 directions (1 for decoder).  The objective function is categorical cross-entropy, optimized by Adam \cite{kingma2014adam}. Training is run with shuffled mini-batches of size 512 and stopped after 20 epochs.

%The classifier is modeled as a simple feed-forward neural network with one hidden layer, dropout ($\rho$ = 0.5), a rectified linear unit (ReLU) non-linearity, and an output layer mapping to the tag set (followed by a Softmax). The size of the hidden layer is set to be identical to the size of the encoder/decoder's hidden state (typically 500 dimensions). The objective function is cross-entropy, optimized by Adam \cite{kingma2014adam} with the recommended parameters ($\alpha$ = 0.001, $\beta_1$ = 0.9, $\beta_2$  = 0.999, $\epsilon$  = $e^{-8}$). Training is run with shuffled mini-batches of size 16 and stopped once the loss on the development set stops improving (allowing a patience of 5 epochs). 

The choice of classifier is motivated by two considerations. First, the classifier takes features only from the current word (or word-pair), without additional context. The goal is to evaluate how well the word representation itself captures pertinent information, potentially including contextual information through the NMT LSTM encoder or decoder. 
Second, using a linear classifier enables focusing on the quality of the representations learned by the NMT system, rather than obtaining state-of-the-art prediction performance. 
In the literature on analyzing neural representations by classification tasks, simple  linear classifiers are a popular choice~\cite{belinkov2019analysis}. Using a stronger classifier may lead to better overall numbers, but does not typically change the relative quality of different representations~\cite[Chapter D.1]{qian-qiu-huang:2016:P16-11,belinkovthesis}, which is our main concern in this work. 

%\todo{YB: need to be more careful about the following sentences. Also, later in the Results we say ``linear''...}
% A note on the choice of classifier. If the goal is to obtain state-of-the-art accuracy,
% %the best results on predicting morphology, 
% then a powerful classifier might be desirable (for instance, an LSTM over encoder states). However, using a linear classifier enables focusing on the quality of the representations learned by the machine translation system rather than obtaining state-of-the-art prediction performance. Arguably, if the learned representations
% are good, then a %non-
% linear classifier should be able to extract useful information
% from them.\footnote{Note that in a few controlled experiments, a linear classifier produced similar trends to the non-linear one, but overall lower results; \namecite{qian-qiu-huang:2016:P16-11} reported similar findings.}

\subsection{Supervised Data and Annotations} \label{sec:supervised-data}

\begin{table}[t]
\centering
				\begin{tabular}{l l  r r r r r r}
					\toprule
				&	& \multicolumn{1}{c}{de} & \multicolumn{1}{c}{en} & \multicolumn{1}{c}{cs} & \multicolumn{1}{c}{ru} &  \multicolumn{1}{c}{fr} & \multicolumn{1}{c}{es}\\
					\midrule
					POS Tags & Train & 14498 & 14498 & 14498 & 11824 & 11495 & 14006  \\
					         & Test &  8172 & 8172 & 8172 & 5999 & 3003 & 5640 \\
					         \midrule
					Morph Tags & Train & 14498 & 14498 & 14498 & 11824 & 11495 & 14006  \\
					        & Test &  8172 & 8172 & 8172 & 5999 & 3003 & 5640 \\	
					        \midrule
					CCG Tags & Train & -- & 41586 & -- & -- & -- & --  \\
					        & Test &  -- & 2407 & -- & -- & -- & -- \\
					        \midrule
					Syntactic Dependency & Train & 14118 & 12467 & 14553 & 3848 & -- & --  \\
					        & Test &  1776 & 4049 & 1894 & 1180 & -- & -- \\
				 \midrule
					Semantic Tags & Train & 1490 & 14084 & -- & -- & -- & --  \\
					        & Test &  373 & 12168 & -- & -- & -- & -- \\
					        \midrule
				    Semantic Dependency & Train & -- & 12000 & 11999 & -- & -- & --  \\
					        & Test &  -- & 9692 & 10010 & -- & -- & -- \\
					\bottomrule
				\end{tabular}
				\caption{Train and test data (number of sentences) used to train MT classifiers to predict different tasks. We used automated tools to annotate data for the morphology  tasks and gold annotations for syntactic and semantics tasks.}
                \label{tab:classifierData}
\end{table}

\begin{table}[t]
\centering
				\begin{tabular}{l  r r r r r r r}
					\toprule
					& \multicolumn{1}{c}{de} & \multicolumn{1}{c}{cs} & \multicolumn{1}{c}{ru} & \multicolumn{1}{c}{en} & \multicolumn{1}{c}{ar}  & \multicolumn{1}{c}{fr} & \multicolumn{1}{c}{es}\\
					\midrule
					POS Tags & 54 & -- & -- & 42  & 42 & 33 & -- \\
					\midrule
                    Morphological Tags & 509 & 1004 & 602 & --  & 1969 & 183 & 212\\
                    \midrule
					Semantic Tags & 69 & -- & -- & 66  & -- & -- & \\
					\midrule
					CCG tags & -- & -- & -- & 1272  & -- & -- & \\
					\midrule
                    Syntactic Dependency labels & 35 & 41 & 40 & --  & -- & 40 &  32 \\
                    \midrule
                    Semantic Dependency labels & -- & 64 & -- & 87  & -- & --  & --\\
					\bottomrule
				\end{tabular}

                                \caption{Number of tags (for word-level tasks) and labels (for relation-level tasks) per task in different languages.}
                \label{tab:numTags}

\end{table}

We make use of  gold-standard annotations wherever available, but in some cases we have to rely on using automatic taggers to get the annotations. In particular, to analyze the representations on the decoder side, we require parallel sentences.\footnote{We need source sentences to generate encoder states which in turn are required for obtaining the decoder states that we want to analyze.} It is difficult to get gold-standard data with parallel sentences, so we rely on automatic annotation tools. An advantage of using automatic annotations, though, is that we can reduce the effect of domain mismatch and high out-of-vocabulary (OOV) rate in analyzing these representations. %Please see Tables \ref{tab:oovRate} and Table \ref{tab:locMaj-Syn} for OOV rates across the tasks. In the former we annotated in-domain test-sets and in the latter, we used gold annotated data.

We used Tree-Tagger \cite{schmid:2004:PAPERS} for annotating Russian and MADAMIRA tagger \cite{PASHA14.593.L14-1479} for annotating Arabic. For the remaining languages (French, German, Spanish, and Czech) we used RDRPOST \cite{nguyen-EtAl:2014:Demos}, a  state-of-the-art morphological tagger. %for many language pairs,\todo{YB: but which specific languages were tagged with RDRPOST here?} to annotate train and test data for the classifier. For Russian, we used Tree-Tagger \cite{schmid:2004:PAPERS} and for Arabic we used MADAMIRA \cite{PASHA14.593.L14-1479}. 
For experiments using gold tags, we used the Arabic Treebank for Arabic (with the versions and splits described in the MADAMIRA manual) and the Tiger corpus for German.\footnote{\url{http://www.ims.uni-stuttgart.de/forschung/ressourcen/korpora/tiger.html}} 
For semantic tagging, we used the  semantic tags from the Groningen Parallel Meaning Bank \cite{abzianidze-EtAl:2017:EACLshort}. For syntactic  relation labeling we used the Universal Dependencies dataset \cite{UD-11234/1-1983}. %See Table \ref{tab:synDep} for statistics. 
%We also used 
For CCG supertagging we used the English CCGBank \cite{hockenmaier2007ccgbank}.%, which contains 41586/2407 train/test sentences.
\footnote{There are no available CCG banks for the other languages we experiment with, except for a German CCG bank which is not publicly available \cite{hockenmaier2006creating}.} 
For semantic dependency labeling we used 
PSD, which is a reduction of the tectogrammatical analysis layer of the Prague Czech-English Dependency Treebank, and is made available as part of the Semantic Dependency Parsing dataset \cite{S14-2008,S15-2153}. Most of the PSD dependency labels mark semantic roles of arguments, which are called functors in the Prague dependency treebank.\footnote{% 
The main differences between PSD and the original tectogrammatical annotation are the omission of elided elements, such that all nodes are surface tokens; the inclusion of functional and punctuation tokens; ignoring most cases of function word attachments to content words; ignoring coreference links; and ignoring grammatemes (tectogrammatical correlates of morphological categories). As a side effect, these simplifications make it straightforward to generate representations for surface tokens participating in dependency relations under the PSD formalism. See \url{http://sdp.delph-in.net} for more information on PSD and refer to \citet{biblio:CiToTectogrammaticalAnnotation2009} for details on the original tectogrammatical annotations.}
PSD annotations are available in English and Czech.  
Table \ref{tab:classifierData} provides the amount of data used to train the MT classifiers for different NLP tasks. Table \ref{tab:numTags} details the number of tags (or labels) in each task across different languages. %, which highlights the complexity of each task.
%\todo{YB: we're not consistent in the stats, giving number of tags for all tasks, but number of sentences/tokens for syntactic relations. I think we need to give everything in this paper.}

%\begin{table}[t]
%\centering
%				\begin{tabular}{l|rrr|rrr}
%					\toprule
 %                   					& \multicolumn{3}{c}{Sentences} &  \multicolumn{3}{c}{Tokens/Relations} \\           
  %                   \midrule
	%				& \multicolumn{1}{c}{Train} & \multicolumn{1}{c}{Dev} & \multicolumn{1}{c}{Test} & \multicolumn{1}{c}{Train} & \multicolumn{1}{c}{Dev}  & \multicolumn{1}{c}{Test} \\
	%				\midrule
	%				Arabic &  6075 & 909 & 680 & 218K  & 29K & 28K\\
     %               English & 12543 & 2002 & 2077 & 192K  & 23K & 23K \\
		%			French & 14553 & 1478 & 416 & 342K  & 34K & 10K\\
		%			Russian & 3850 & 579 & 601 & 72K  & 11K & 11K \\
         %           Spanish & 14187 & 1400 & 426 & 368K  & 36K & 12K \\
			%		\bottomrule
			%	\end{tabular}
             %   \caption{Statistics for datasets of syntactic relations, extracted from the Universal Dependencies datasets}
          %      \label{tab:synDep}
% \end{table}

\section{Morphology Results}
\label{sec:results-morphology}

\begin{table}[t]
\centering
				\begin{tabular}{ll r r r r r}
					\toprule
			&	& \multicolumn{1}{c}{de} & \multicolumn{1}{c}{cs} & \multicolumn{1}{c}{ru} & \multicolumn{1}{c}{en} &  \multicolumn{1}{c}{fr} \\
					\midrule
					%\multicolumn{7}{c}{POS Tags} \\ 
					 \midrule %\cmidrule(lr){1-2}
					 POS Tags & MT Classifier  & 94.0 & -- & -- & 95.8   & 96.3 \\
					\midrule
           & Majority & 88.4 & -- & -- & 90.1  & 92.6\\
        Baselines             & char-to-POS & 98.3 & -- & -- & 97.7  & 99.2 \\
                     & POS Classifier & 95.4 & -- & -- & 96.0  & 98.5 \\
\midrule
\midrule
                    %\multicolumn{7}{c}{Morphological Tags} \\ 
                     Morphological Tags & MT Classifier  & 80.5 & 85.2 & 87.7 & -- & 88.2 \\
                    \midrule
           & Majority & 68.3 & 70.4 & 74.8 & --  & 84.7\\
      Baselines              	 & char-to-Morph & 92.7 & 95.7 & 94.2 & --  & 98.6\\
                    	 & Morph Classifier  & 89.6 & 90.5 & 90.5 & --  & 95.8 \\
                    
					\bottomrule
				\end{tabular}
                \caption{POS and morphological tagging results: Comparing classifier trained on char-based NMT representations with several baselines: (i) local majority baseline (most frequent tag), (ii) character-to-tag trained using sequence-to-sequence model on the same training data as the MT systems, (iii) Classifier trained on representations extracted from (ii) to match the MT generated representations). NMT systems used here to extract representations are character-based models,
                 %Here the results are with the top layer activations of the encoder in word-based NMT models, which are 
                 trained on translating each language to English (and English to German).
                The classifier results are substantially above the majority baseline, indicating that NMT representations learn non-trivial amounts of morphological information.}
                \label{tab:wordClassifier}
\end{table}

%To this end, we trained sequence-to-sequence {\tt char-to-tag} models using the same data and configuration that we used to train our char-based MT systems. Following Reviewer C's advice, we additionally trained actual classifiers using the features extracted from the trained ({\tt char-to-tag}) models. This allows us to exactly compare representations learned for the task of translation versus the representations that are directly optimized towards the task (POS or morphological tagging, for example).

In this section, we investigate what kind of morphological information is captured within NMT models, using the tasks of POS and morphological tagging. To probe this, we annotated a subset of the training data (see Table \ref{tab:classifierData}) using POS or morphological taggers. We then generated features from the trained NMT models and trained a linear classifier using these features to predict the POS or morphological tags. 

While our goal is not to surpass state-of-the-art tagging performance,  %it would be good to have 
we still wanted to compare against several reference points to assess the quality of the underlying representations. To this end we report several baselines: (i) A simple local majority baseline where each word is assigned its most frequent tag and unknown words are assigned the most frequent global tag. (ii) We annotated the data used to train NMT models using the tools mentioned above and trained {\tt char-to-tag} models using the same sequence-to-sequence regime we used to train our MT systems. This can be seen as a skyline reference. (iii)  To have a closer comparison with our MT classifier, we generate features from the trained {\tt char-to-tag} models and train a linear classifier using these features. This allows us to exactly compare representations learned for the task of translation versus the representations that are directly optimized towards the task (POS or morphological tagging, for example). 

Table \ref{tab:wordClassifier} shows the prediction accuracy of the classifiers trained on the \emph{encoder-side} representations. MT classifiers always outperform the majority baseline which entails that the representations contain non-trivial linguistic information about language morphology.  The accuracy is high when the language is morphologically poor (e.g., English) or the task is simpler (fewer tags to predict;  see Table \ref{tab:numTags}). On the contrary, the accuracy in the case of a morphologically-rich language such as Czech is lower. %The overall accuracy trend shows that the representations possess information to predict language morphology.\footnote{The classifiers achieve higher accuracies compared to the local majority baseline i.e., selecting the most frequent tag for each word and the most frequent global tag for the unknown words (See Table \ref{tab:wordClassifier}).} %\todo{YB: can we have the local majority baseline numbers in the table? ND: done}
The {\tt char-to-POS/Morph} baselines seems to give much higher numbers compared to ours, but remember that these models are trained on a lot more data (the entire data on which the MT models were trained) and with a more sophisticated bilingual LSTM with attention model, compared to the MT classifier which is trained on a small subset of neural activations using a simple logistic regression. A much closer skyline reference is the POS/Morph classifiers that are trained on the same data and model architecture as the MT classifier, with the difference that former is trained on the representations optimized for the task itself whereas the latter is trained on the representations optimized towards the task of machine translation. Therefore, this is still comparing apples to oranges, but provides a more exact reference for the quality of MT representations with respect to learning morphology.

We now proceed with answering more specific questions regarding several aspects of the NMT systems: (i) How do the representations trained from different translation units (word vs.\ character vs.\ subword units) compare? (ii) How do the representations trained from the encoder and decoder compare? (iii) What kind of information different layers capture? and (iv) How does the target language affect the learned source language representations? 

%\todo{YB: should we refer to table \ref{tab:bleu-scores} somewhere? ND: Done} 

\subsection{Impact of Translation Unit on Learning Morphology} 

\begin{table}[t]
\centering
				\begin{tabular}{lcccccc}
					\toprule
			%Lang		
            & \multicolumn{2}{c}{de} & \multicolumn{2}{c}{cs} & \multicolumn{2}{c}{ru} \\ 				
            \cmidrule(lr){2-3} \cmidrule(lr){4-5} \cmidrule(lr){6-7}
            %\midrule
             %Heuristic 
             & \multicolumn{1}{c}{subword} & \multicolumn{1}{c}{char} & \multicolumn{1}{c}{subword} & \multicolumn{1}{c}{char} & \multicolumn{1}{c}{subword} & \multicolumn{1}{c}{char} \\ 					\midrule
					Last & 78.5 & 80.5& 78.6 & 88.3 & 80.0 & 88.8  \\
                    Average & 76.3 & 79.2 & 76.4 & 84.9 & 78.3 & 84.4  \\
					\bottomrule
				\end{tabular}
                \caption{Classification accuracy for morphological tags using representations generated by aggregating BPE subword or character representations using either the average or the last LSTM state for each word. Here the representations are obtained by concatenating the encoding layers of NMT models trained on translating each language to English. Using the last hidden state consistently outperforms the average state. }
                \label{tab:lastVsAvg}
\end{table}

\begin{figure*}[ht]
\centering
\includegraphics[width=\linewidth]{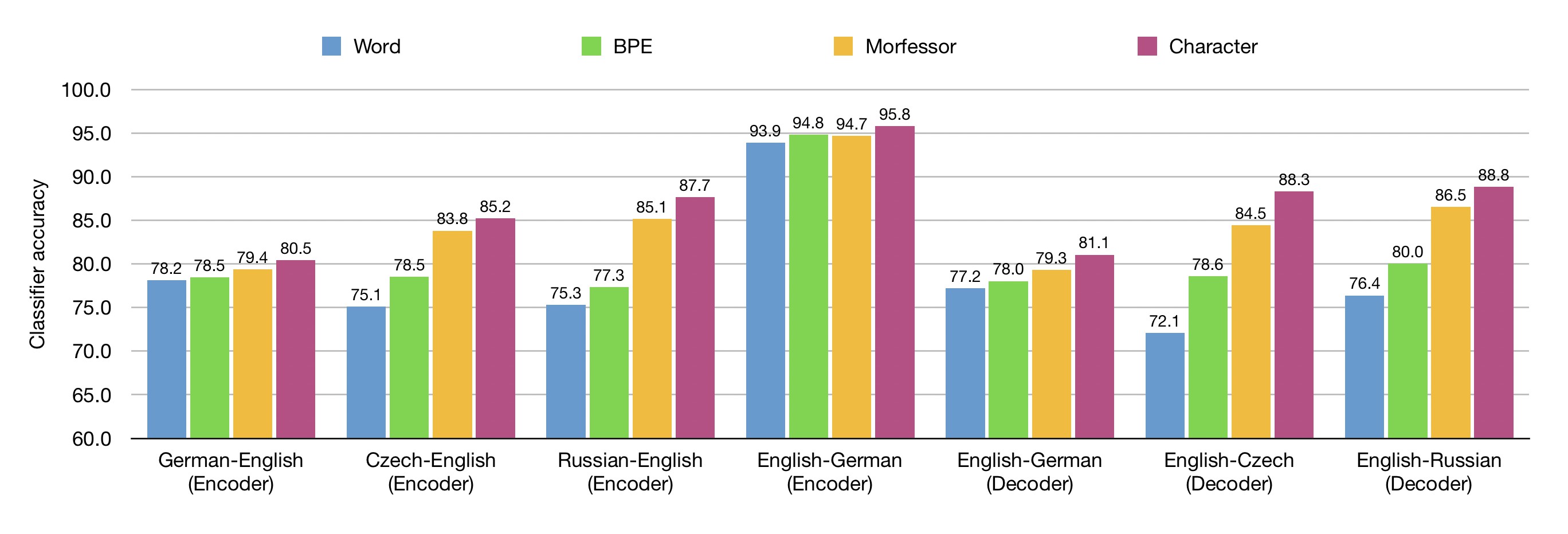}
\caption{Morphological classification accuracy with different translation units and language pairs. 
When comparing encoder (decoder) representations, 
we train NMT models with different source (target) side translation units---words, BPE subwords, Morfessor subwords, or characters---and hold the target (source) side unit fixed as BPE subwords.  %Encoder models are trained with BPE as target and Decoder models with BPE as source.
}
\label{fig:morph-results}
\end{figure*}

We trained NMT systems with different translation units: word, character, and subword units, of which we tried two, namely BPE \cite{sennrich-haddow-birch:2016:P16-12} and morphological segmentation \cite{smit-EtAl:2014:Demos}. For subword and character units, we found that the activation of the last subword/character unit of a word performed consistently better than using the average of all activations, so we present the results using the \textbf{Last} method throughout the paper (see Table \ref{tab:lastVsAvg} for comparison). %\todo{YB: if average isn't used, why mention it at all? Can we give at least a little table comparing them and then say from hence on we only use last? ND: Done} 

Figure \ref{fig:morph-results} summarizes the results of predicting morphology with representations learned by different models. %\todo{YB: results with which MT data? In general, we should either always specify the datasets, or designate one (WMT+IWSLT?) as the default and only specify when we diverge (UN?) ND: WMT} 
The character-based representations consistently outperformed other representations on all language pairs, while the word-based representations achieved the lowest accuracy. The differences are more significant in the case of languages with relatively complex morphology, notably Czech and Russian. We see a  difference of up to 14\% in favor of using character-based representations when compared with the word-based representations. The improvement is minimal in the case of English (1.2\%), which is a morphologically-simpler language. Comparing subword units as obtained using Morfessor and BPE, we found Morfessor to give much better morphological tagging performance, especially in the case of the morphologically-richer languages, Czech and Russian. %This is due to the 
The representations learned from morpheme-segmented units %which 
were found helpful in learning language morphology. These findings are also somewhat reflected in the translation quality (Table \ref{tab:bleu-scores}). The  character-based segmentation gave higher BLEU scores compared to a BPE-based system in the case of the  morphologically-rich language Czech, but character-based models performed poorly in the case of German, which requires handling long-distance dependencies. Our results (discussed later in Section \ref{sec:results-syntax}) show that character-based representations are %poor 
less effective at handling syntactic dependencies.%\todo{YB: maybe soften this claim. ND: done} 

\begin{table}[t]
	\begin{minipage}{0.50\textwidth}
		\centering
				\resizebox{\columnwidth}{!}{
			\begin{tabular}{l cccc}
				\toprule
				& \multicolumn{1}{c}{de-en} & \multicolumn{1}{c}{cs-en} & \multicolumn{1}{c}{ru-en} & \multicolumn{1}{c}{en-de} \\
				\midrule
				word  & 34.0 & 27.5 & 20.9 & 29.7 \\
				bpe  & 35.6 & 28.4 & 22.4 & 30.2 \\
				morfessor  & 35.5 & 28.5 & 22.5 & 29.9 \\
				char  & 34.9 & 29.0 & 21.3 & 30.0 \\
				\bottomrule
			\end{tabular}
		}
        \caption{BLEU scores across language pairs with different translation units on the source side (the target side is held fixed as BPE). The NMT models are trained on NEWS+TED data.         }
		\label{tab:bleu-scores}
        \end{minipage}
	\hfill
	\begin{minipage}{0.45\textwidth}
		\centering
		\resizebox{\columnwidth}{!}{
			\begin{tabular}{l cccc}
				\toprule
				& \multicolumn{1}{c}{de-en} & \multicolumn{1}{c}{cs-en} & \multicolumn{1}{c}{ru-en} & \multicolumn{1}{c}{en-de} \\
				\midrule
				MT  & 3.42 & 6.46 & 6.86 & 0.82 \\
				Classifier & 4.42 & 6.13 & 6.61 & 2.09 \\
				\bottomrule
			\end{tabular}
		}
        \caption{Out-of-vocabulary (OOV) rate (\%) in the (source-side) MT and morphological classification test sets. The morphologically richer Czech (cs)  and Russian (ru) have higher OOV rates. 
        }
         \label{tab:oovRate}
	\end{minipage}
\end{table}

\begin{figure*}[t]
\includegraphics[width=0.98\linewidth]{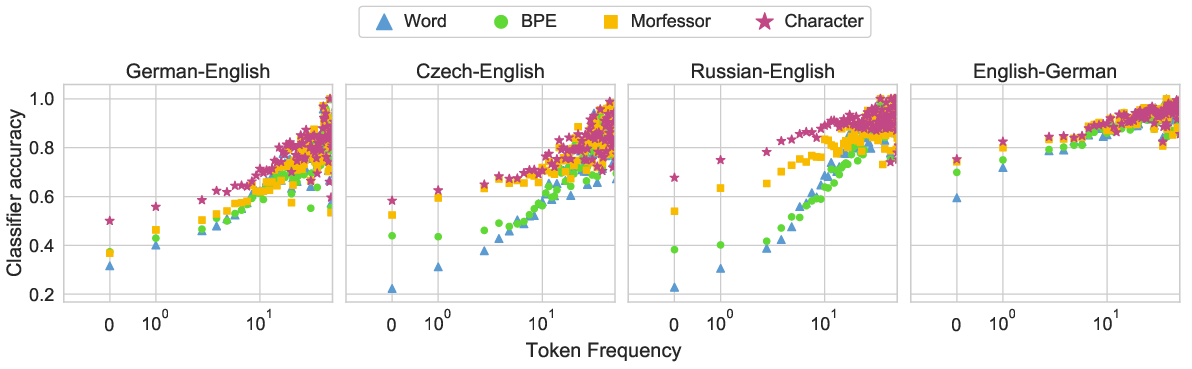}
\caption{Morphological tagging accuracy vs.\ word frequency for different translation units on the encoder-side. The target side is held fixed as BPE. The representations for training the morphological classifier are obtained from the top layer of the encoder. Character representations perform better than other ones especially in low-frequency regimes. 
%\alert{YB: Fahim, can you improve figure readability at the high-freq end using opaque or empty symbols, or something similar?}
}
\label{fig:frequency-analysis}
\end{figure*}

\subsubsection{Handling Unknown and Low Frequency Words} 
We further investigated whether the performance difference between various representations is due to the difference in modeling infrequent and out-of-vocabulary words. As Table \ref{tab:oovRate} shows, the morphologically-richer languages have higher OOV rates.  Figure \ref{fig:frequency-analysis} reveals that the gap between different representations is inversely related to  the frequency of the word in the training data: character-based models perform much better than others on less frequent and OOV words. 
The ranking of different units in low frequency regimes is consistent with the overall results in Figure \ref{fig:morph-results} -- characters perform best, followed by Morfessor subwords, BPE subwords, and words.  

\subsection{Encoder versus Decoder Representations}
 
The decoder \texttt{DEC} is a crucial part in an MT system with access to both source-side representations and partially generated target-side representations which it uses to generate the next target word. We now examine if the representations learned on the decoder-side possess the same amount of morphological knowledge as the encoder side. To probe this, we flipped the language direction and trained NMT systems with English$\rightarrow$\{German, Czech,  Russian\} configurations. Then, we use the trained model to encode a source sentence and generate features for words in the target sentence. These features are used to train a classifier on morphological tagging on the target side. Note that in this case the decoder is given the correct target words one-by-one, similar to the usual NMT training regime. The right hand-side of Figure \ref{fig:morph-results} shows a similar performance trend as in the case of encoder-side representations, with character units performing the best while word units performing the worst. Again, morphological units performed better than the BPE-based units.  

Comparing encoder representations with decoder representations, it is interesting to see that in several cases the decoder-side representations performed better than the encoder-side representations, even though they are trained using  a uni-directional LSTM only. Since we did not see any notable trends in differences between encoder and decoder side representations, we only present the encoder-side results in the rest of the paper. %\todo{YB: should we say these results supersede the ACL+IJCNLP papers? Also, should we go into the question of representing the previous word or the following word (as our buggy implementation did)? ND: I think the paper is already very dense and we shouldn't go into this}  

\subsection{Effect of Network Depth}

\begin{figure}[t]
\centering
\includegraphics[width=0.9\linewidth]{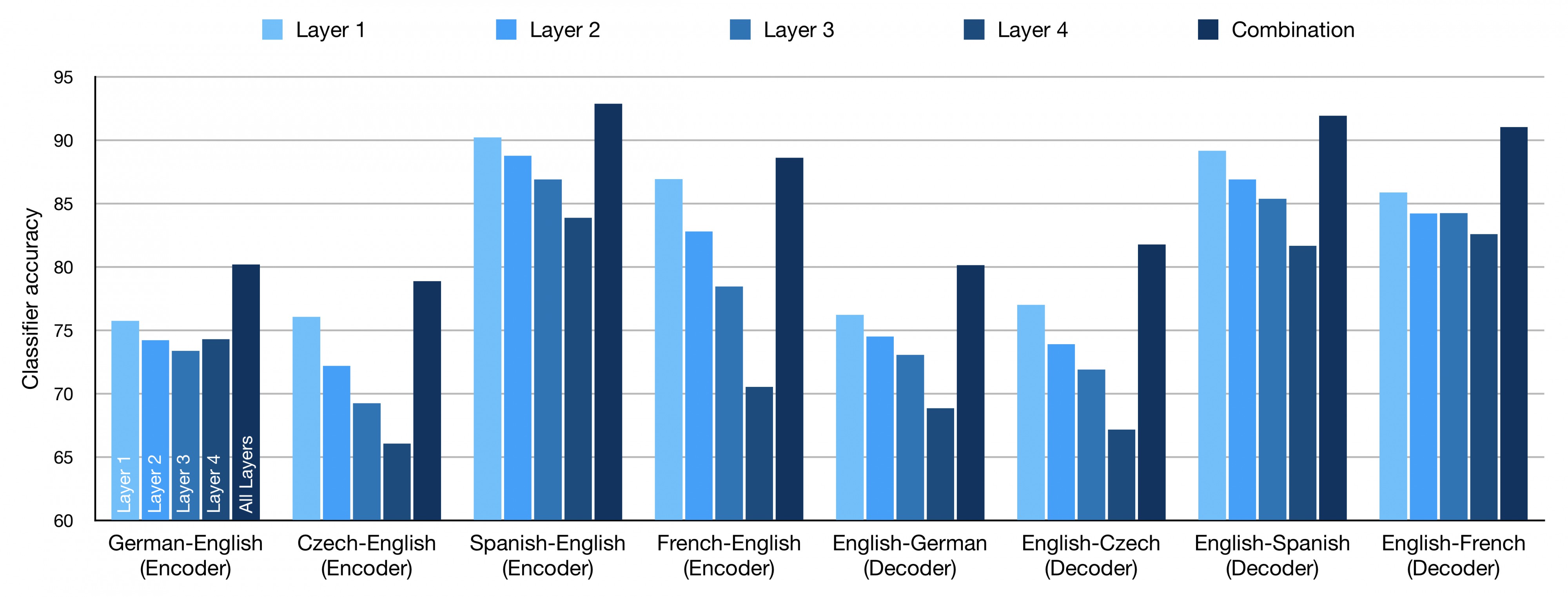}
\caption{Morphological tagging accuracy using representations from layers 1 to 4, taken from encoders and decoders of different language pairs. Here the NMT models were trained with BPE units.  Layer 1 generates the best representations and in most cases there is a gradual decrease with each layer. The combination of representations improves beyond layer 1, indicating that some morphological information is distributed among layers. 
}
\label{fig:layer-effect-all-langs}
\end{figure}

Modern NMT systems use very deep architectures \cite{wu2016google,TACL863}. %with up to 8 or 16 layers \cite{wu2016google,TACL863}. 
We are interested in understanding what kind of information different layers capture. Given a trained NMT model with multiple layers, we extract feature representations from the different layers in the encoder. We trained 4-layered models (using (NEW+TED+Europarl data).

%Let $\texttt{ENC}^l_i(s)$ denote the encoded representation of word $w_i$ after the $l$-th layer. \todo{YB: the notation may be unified or avoided} We can vary $l$ and train different classifiers to predict POS or morphological tags.  Here we focus on the case of a 4-layer encoder-decoder model for simplicity ($l \in \{1,2,3,4\}$). 

Figure~\ref{fig:layer-effect-all-langs} shows morphological tagging results using representations from different encoder and decoder layers across five language pairs. %We found that the representations generated from Layer 1 are consistently better compared to the other layers in predicting word morphology. 
The general trend %is that passing word vectors through the NMT encoder improves POS/morphological tagging,\todo{YB: improves compared to what? we don't give layer 0 anymore. We can show baseline numbers here as horizontal lines, which would make the point} 
%which can be explained by the % 
%contextual information contained in the representations after one layer. However, it turns out 
shows that representations from the first layer are better than those from the higher layers, for the purpose of capturing morphology. We found this observation to be true in multi-layered decoder as well (see the right side of Figure \ref{fig:layer-effect-all-langs}).
We verified these findings with models trained using 2, 3 and 4 layers. Layer 1 was consistently found to give better accuracy on the task of POS tagging and morphology learning. %\todo{YB: let's give numbers in the appendix} %\todo{YB: we may be required to give the numbers, ND: We have numbers} 
We also found the pattern to hold for representations trained on other units (for example character-based units).%\todo{YB: let's give numbers in the appendix}  %\todo{YB: again, we may be asked for the numbers, ND: We have numbers} 

Another interesting result to note is that concatenating representations from all the layers gave significantly better results compared to any individual layer (see \textbf{Combination} bars in Figure \ref{fig:layer-effect-all-langs}). This implies that although much of the information related to morphology is captured at the lower layer, some of it is also distributed to the higher layers. We analyzed individual tags across layers and found that open class categories such as  \emph{verbs} and \emph{nouns} are distributed across several layers, although the majority of the learning of these phenomena is still done at layer 1. Please refer to \cite{dalvi:2019:AAAI} %. and Table \ref{tab:open-class-percentage} in the supplementary 
for further information.

%\todo{YB: let's expand on this analysis of tags across layers for the camera-ready. ND: Fahim is working on this} 

\subsection{Effect of Target Language}

The task of machine translation involves translating from one language into another. While translating from morphologically-rich languages is a challenging task, translating into such languages is even harder. How does the target language affect the learned source language representations? Does translating into a morphologically-rich language require more knowledge about source language morphology? To investigate these questions, we trained NMT models by keeping the source side constant and using different target languages. To make a fair comparison, the models are trained on the intersection of the training data based on the source language. %In this way the experimental setup is completely identical: the models are trained on the same source sentences with translations into different languages. 

\begin{figure}[t]
\centering
\includegraphics[width=0.65\linewidth]{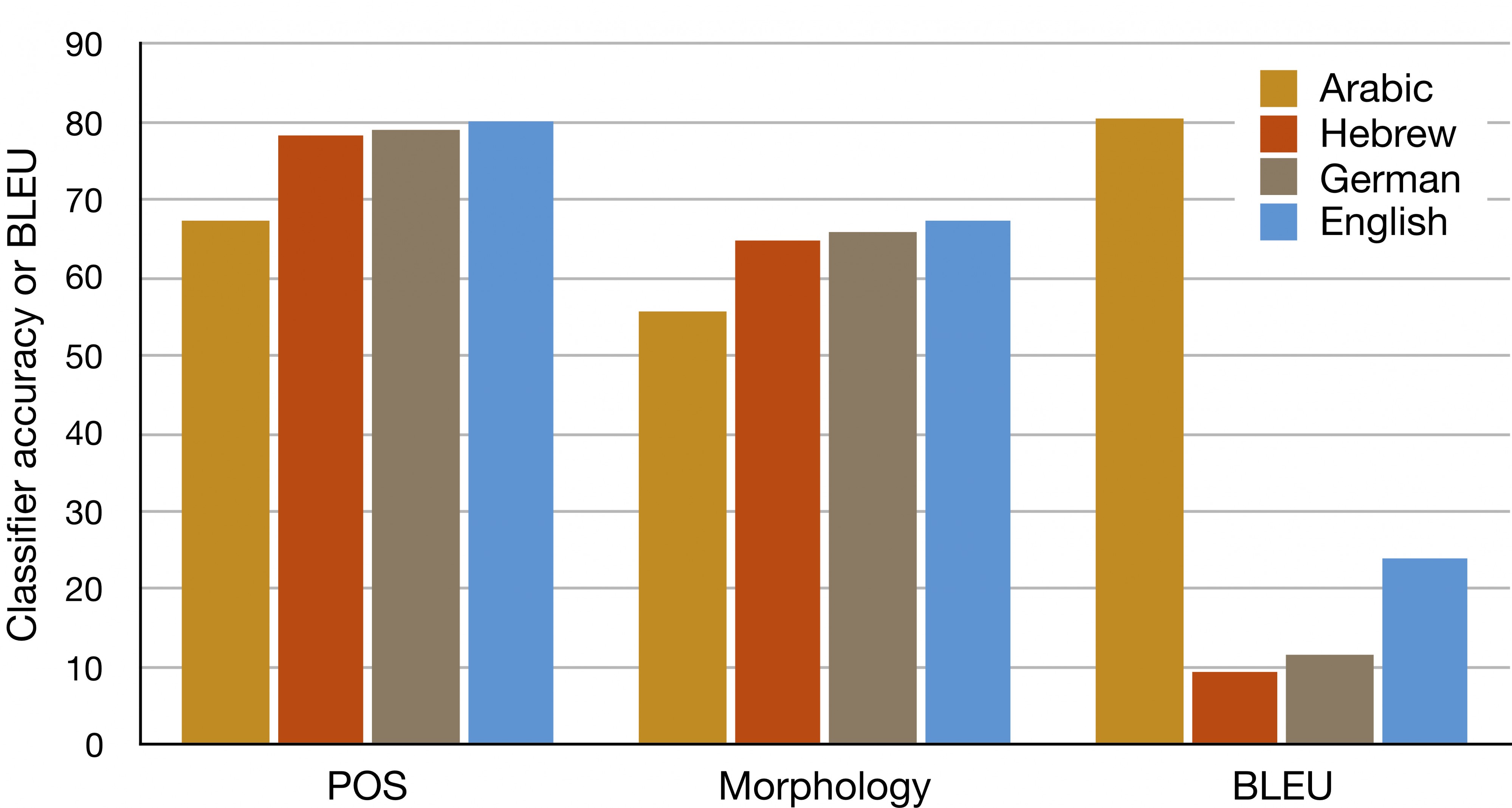}
\caption{Effect of target language on representation quality of the Arabic source. POS and morphological tagging accuracy, and BLEU scores, using encoder-side representations from NMT models trained with different target languages.  These results are obtained with top layer representations of 2-layers word-based NMT models.
The target language has a small effect on source-side representation quality, with better MT models generating better representations, except in the auto-encoder case (Arabic bar). 
}
\label{fig:target-lang}

%\vspace{-10pt}
\end{figure}

Figure \ref{fig:target-lang} shows the results of such an experiment, with models translating from Arabic to several languages: English, Hebrew, German and Arabic itself. These target languages represent a morphologically-poor language (English), a morphologically-rich language with similar morphology to the source language
(Hebrew), and a morphologically-rich language with different morphology (German).
As the figure shows, 
the representations that are learned when translating into English are better for predicting POS or morphology than those learned when translating into German, which are in turn better than those learned when translating into Hebrew. 

How should we interpret these results? English is a morphologically-poor language that does not display many of the morphological properties that are found in the Arabic source. In contrast, German and Hebrew have richer morphologies, so one could expect that translating into them would make the model learn more about morphology. However, Arabic representations learned from the Arabic$\rightarrow$English model are superior in learning morphology. A possible explanation for this phenomenon is that the Arabic$\rightarrow$English model is simply better than the Arabic$\rightarrow$Hebrew and Arabic-German models, as hinted by the BLEU scores. The inherent difficulty in translating Arabic to Hebrew/German may affect the ability to learn good representations of word structure or perhaps more data is required in the case of these languages to learn Arabic representations of the same quality. However, it turns out that an Arabic$\rightarrow$Arabic autoencoder learns to recreate the test sentences extremely well, even though its word representations are actually inferior for the purpose of POS/morphological tagging (Figure \ref{fig:target-lang}). This implies that higher BLEU does not necessarily entail better morphological representations. In other words, a better translation model learns more informative representations, but only when it is actually learning to translate rather than merely memorizing the data as in the autoencoder case. We found these results to be consistent in
other language pairs, i.e., by changing the source from Arabic to German and Czech and also using character models instead of words (See Section~\ref{sec:targetLang} in the supplementary material for more details), however more through study is required along this direction as \namecite{bisazza-tump-2018-lazy} performed a similar experiment on a fine-grained tag level and found contrastive results.

\section{Syntax Results}
\label{sec:results-syntax}

\begin{table}[t]
\centering
				\begin{tabular}{l l rrrrr}
					\toprule
				&	& \multicolumn{1}{c}{de} & \multicolumn{1}{c}{cs} & \multicolumn{1}{c}{ru} & \multicolumn{1}{c}{en}  & \multicolumn{1}{c}{fr} \\
					\midrule
					\multirow{2}{*}{Syntactic Dependency} & MT Classifier & 91.5 & 91.8 & 89.6 & 93.4  & 94.4\\
                     & Majority & 69.0 & 68.6 & 59.4 & 67.1  & 72.4 \\
                     & OOV Rate & 10.3 & 12.9 & 21.7 & 5.9 &  10.9 \\ 
                     \midrule
                   & \namecite{P15-2041} & -- & -- & -- & 93.1  & -- \\
                    \multirow{2}{*}{CCG Tags} & MT Classifier & -- & -- & -- & 91.9  & -- \\
                    	 & Majority & -- & -- & -- & 72.1  & -- \\
                    	 & OOV Rate & -- & -- & -- & 6.9 & --  \\ 
					\bottomrule
				\end{tabular}
\caption{Local majority baseline (most frequent tag/label) and classification accuracy using encoder representations generated by the NMT models, on syntactic tasks. The models are trained on translating each language to English (or German in the English case). The classifier results are far superior to the majority baseline, indicating that NMT representations contain a non-trivial amount of syntactic information.
                }
                \label{tab:locMaj-Syn}
\end{table}

To evaluate the NMT representations from a syntactic perspective, we consider two tasks. 
%The final linguistic property that we want to analyze is \emph{Syntax}. 
First, we made use of CCG supertagging,  which is assumed to capture syntax at the word level. %, to analyze \emph{encoder} representations. 
Second, we used dependency relations between any two words in the sentence for which a dependency edge exists, to investigate how words compose. %carry out further analysis. 
Specifically, we ask the following questions: (i) Do NMT models acquire structural information while they are
being trained on flat sequences of bilingual sentences? (ii) How do representations trained on different translation units (word vs.\ character vs.\ subword units) compare with respect to syntax? and (iii) Do higher layers learn better representations for these kinds of properties than lower layers?

The analysis carried out previously was chiefly based on lexical properties. %, and was therefore limited to unit representations that are learned in NMT models. 
To strengthen our analysis, we further used dependency relations which are available for many different language pairs unlike CCG supertags. Here we concatenate the representations of two words in a relation and ask the classifier to predict their syntactic relation. Table \ref{tab:locMaj-Syn} shows that NMT representations are syntax aware. In both tasks (CCG supertagging and syntactic dependency labeling), the classifier accuracy is much higher compared to the local majority baseline\footnote{For the syntactic dependency majority baseline, we assume the most frequent label of the arc (head-modifier pair). %\todo{YB: This isn't clear; you mean the most frequent tag of the head-modifier pair?} 
When the pair is unseen during test, we ignore the head and fall back to using modifier only. %On the contrary training the classifier without concatenating the representations of head word resulted in a drop of accuracy.
It is non-trivial to train sequence-to-sequence models for the dependency tasks, so we only rely on the majority baseline for comparison.
} 
demonstrating that the representations learned during NMT training learn non-trivial amount of syntactic information.\footnote{We do not have similar baselines for the syntax and semantic tasks as we have for the task of morphology prediction. The reason for this discrepancy is that we used automatic tools for annotating data for POS and morphological tagging, but gold annotated data for syntax and semantic tasks. While state-of-the-art POS and morphological tagging tools are freely available, the same is not true for semantic and CCG tagging. We therefore resorted to use the published numbers as the skyline baseline in this case. For syntactic and semantic dependency labeling tasks an additional complexity is how to train a {\tt seq-to-seq} baseline with such annotations. Remember that the task involves modeling head and modifier word to predict a dependency relation. In the case of semantic dependency, there could be multiple heads for a modifier word.} 
We now proceed to answer the other two questions, namely, the impacts of translation unit and representation depth. 

\subsection{Impact of Translation Unit on Learning Syntax}

\begin{figure*}[t]
\centering
\includegraphics[width=1.0\linewidth]{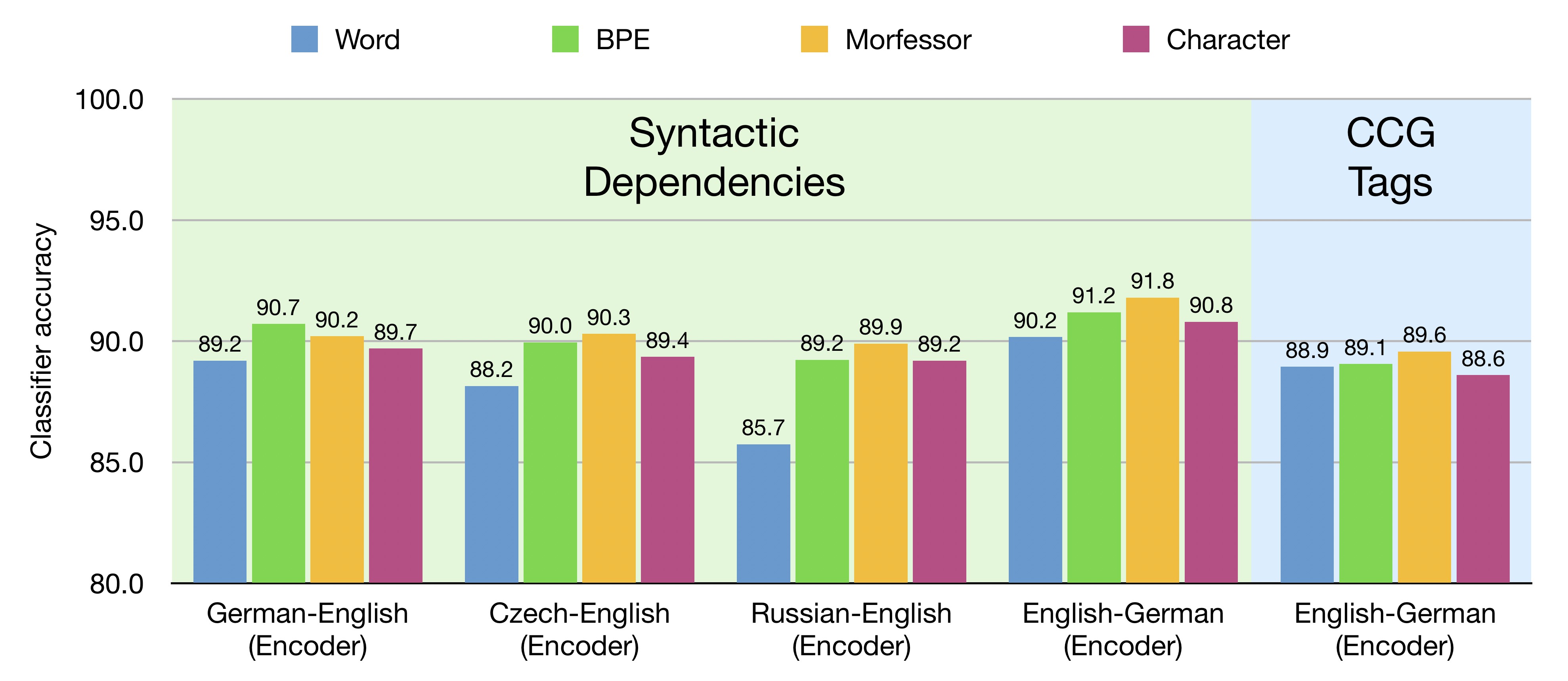}
\caption{Dependency labeling and CCG supertagging accuracy using encoder representations obtained from NMT models trained with different translation units on the source-side the target side is fixed as BPE. 
Subword units generate better representations than character- or word-based ones. 
}
\label{fig:syntax-results-by-repr}
\end{figure*}

% \begin{figure}
% \centering
% 		\includegraphics[width=0.35\linewidth]{figs/syntax-results.eps}
%   	\caption{CCG supertagging for different units in English.}
% 	\label{fig:syn-accuracies}
% \end{figure}

%\sout{We wanted to analyze whether the representations trained from subword units, specifically character units are able to capture long-distance dependencies as well as the word-based models.} 

While character-based models are effective at handling unknown and low-frequency words, they have been found poor at capturing long-distance dependencies. \namecite{sennrich:2017:EACLshort} performed an evaluation based on contrastive translation pairs and found the subword-based system better in capturing long-distance dependencies. Here we directly pit the representations trained on different translation units against each other and compare their performance in predicting syntactic properties. 
%\namecite{sennrich:2017:EACLshort} demonstrated that while character-based models are effective in handling unknown words, they perform worse in capturing long-distance dependencies. However, his evaluation was based on contrastive translation pairs and did not evaluate the representation quality directly.
Figure \ref{fig:syntax-results-by-repr} shows that representations learned from subword units (BPE and Morfessor) consistently outperform the ones learned from character units in both tasks (CCG and syntactic dependency labeling), reinforcing the results found by \citet{sennrich:2017:EACLshort}.  Character-based models, on the other hand, do better than word-based models, which could be attributed to unknown words (in the word-based models).
%Our results in Figure \ref{fig:syntax-results-by-repr} show a similar trend on CCG supertagging, as the representations trained from character-based models gave worse accuracy compared to those trained using word- or subword units. 
%The difference in classifier accuracy is not much, %drastically different, unlike morphology where character-based models significantly outperformed word-based and subword-based representations. This shows that 
We found subword units, particularly those obtained using a morpheme-based segmentation, to give the best results. %\todo{Can we speculate why? }
This could be because the linguistically motivated subword units are more aligned with the syntactic task than the compression-based BPE segmentation. %\todo{YB: the last sentence seems far fetched, I'm not sure Morfessor can be called linguistically motivated}

A possible confound is that character-based models start from a lower linguistic level compared to word or subword models and may require more depth to learn long-range dependencies. To verify this, we trained 3-layered character models for Czech-to-English and English-to-German. We extracted feature representations and trained classifiers to predict syntactic dependency labels. Our results show that using an additional layer does improve the prediction accuracy, giving the same result as subword segmentation (Morfessor) in the case of Czech-to-English, but still worse in the case of English-to-Czech (see Table \ref{tab:threeLayer} in the Appendix for results).

\subsection{Effect of Network Depth}
\label{sec:syntax-depth}

\begin{figure}
  \centering
  \begin{subfigure}{.49\textwidth}
    \centering
      \includegraphics[width=1\linewidth]{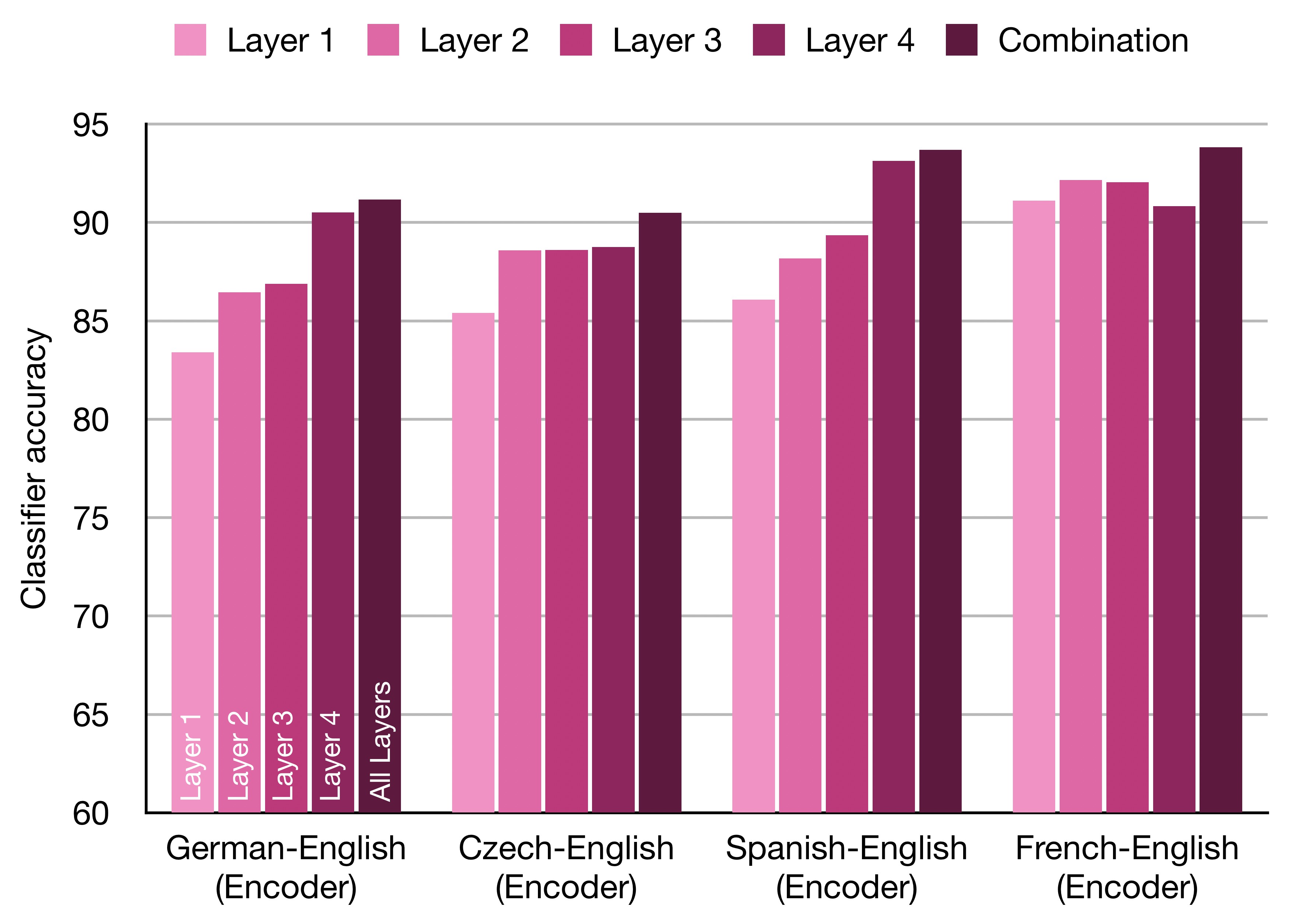}
      \caption {Syntax Dependency}
      \label{fig:layer-effect-all-langs-syntax}
  \end{subfigure}
  \begin{subfigure}{.49\textwidth}
      \centering
      \includegraphics[width=0.75\linewidth]{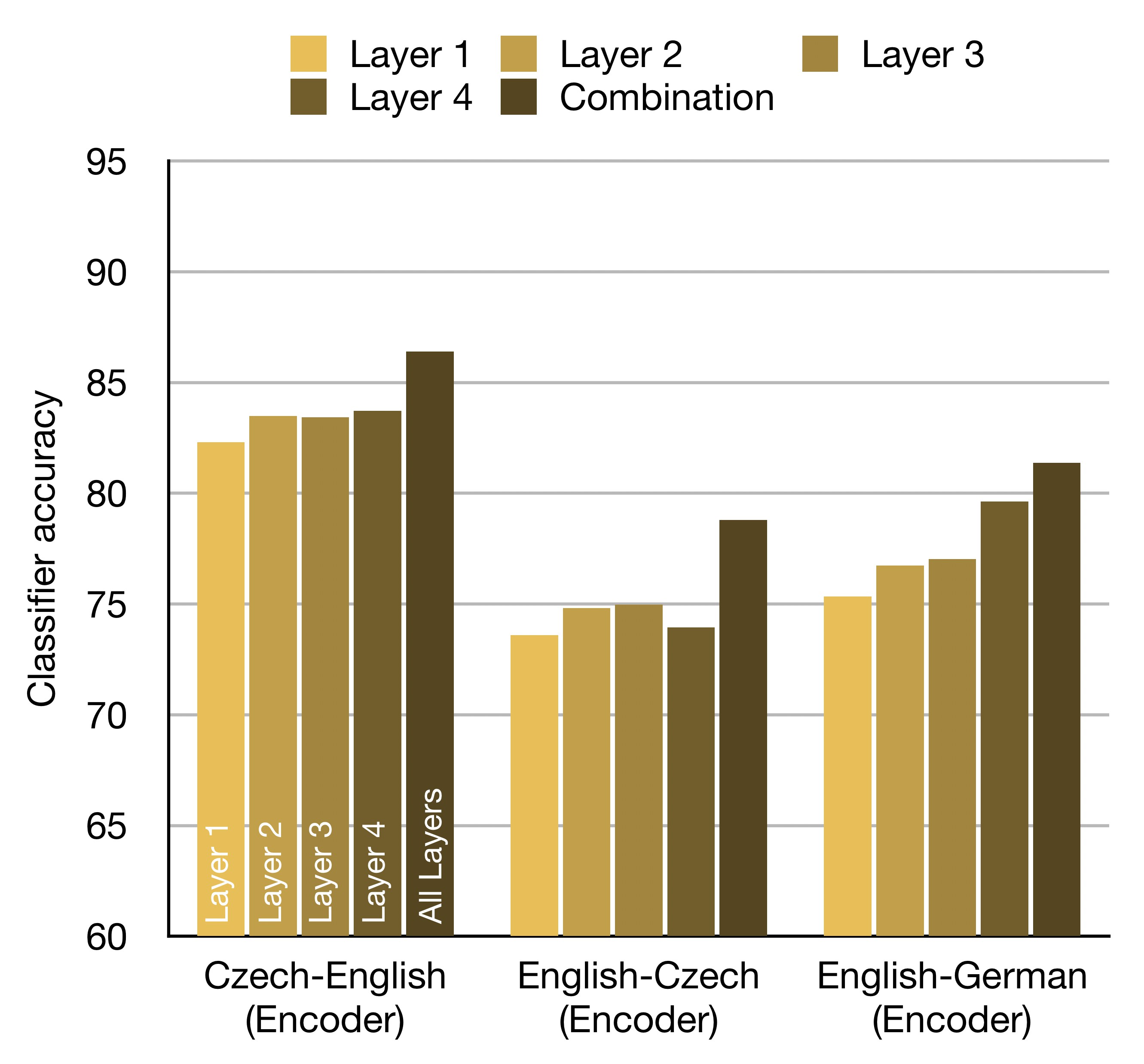}
      \caption {Semantic Dependency}
      \label{fig:layer-effect-all-langs-semantics}
  \end{subfigure}
  \caption{Syntactic and semantic dependency labeling accuracy using using representations from layers 1 to 4, taken from encoders of BPE-based NMT models in different language pairs. 
  Higher layers generate better representations for these tasks in most cases, but a combination of all layers works best, indicating that some relevant information is also captured in the lower layers. 
  }
\end{figure}

We previously found that morphology is predominantly being captured in layer 1 of the NMT models. We now repeat the experiments for syntactic dependencies.  Figure \ref{fig:layer-effect-all-langs-syntax} shows the results of predicting syntactic dependency labels using representations from different layers in the trained models. We found that representations from layer 4 performed better than representations from lower layers except for the French encoder, where layer 3 performs better.  We also repeated this experiment with CCG supertags (see Table \ref{tab:results-ccgtags-4layers} in the supplementary  material) %\todo{YB: Fahim, can you turn the table from the response into a figure and add it as a subfigure in figure 9?} 
and found that higher layers (3 and 4) consistently outperform lower  layers and except for English-Czech, the final layer gives the best accuracy in all cases.\footnote{In their study of NMT and language model representations, \citet{Zhang2018LanguageMT}  noticed that POS is better represented at layer 1 while CCG supertags are sometimes, but not always, better represented at layer 2 (out of 2-layer encoders).}  %\todo{YB: where are these CCG results? ND: there's no figure to refer to, just saying in words} 
These results are consistent with the syntactic dependency results. We  repeated these experiments with the multi-parallel UN corpus by training English-to-\{French, Arabic, Spanish, Russian, and English\} bilingual models. Comparing successive layers (for example, comparing layer 2 versus layer 3), in the majority of the cases, the higher layer performed statistically significantly better than the lower one ($\rho$ < 0.01), according to the approximate randomization test \cite{sigf06}.\footnote{ See Section~\ref{sec:significanceTests} in the supplementary information for the detailed results.}
Similar to the results on morphological tagging, a combination of all layers achieved the best results. See the \textbf{Combination} bar in Figure \ref{fig:layer-effect-all-langs-syntax}. This implies that although syntax is mainly learned at higher layers, syntactic information is at least partly distributed across the network. 

\begin{figure}
\centering
		  \includegraphics[width=0.65\linewidth,trim={0 0.5cm 0 0},clip]{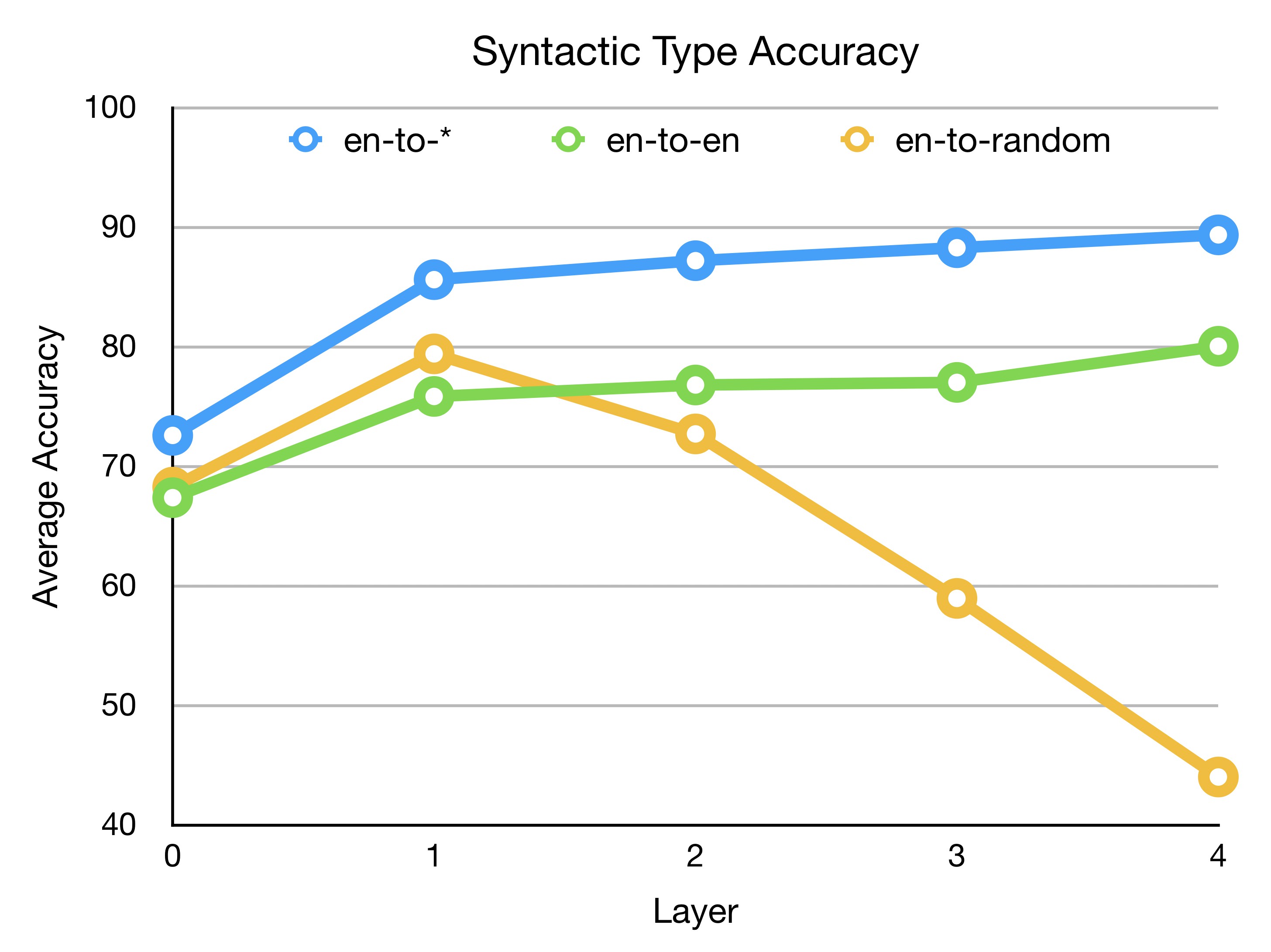}
	  	\caption{Syntactic dependency labeling results with representations from different encoding layers of (word-based) NMT models trained on translating English to other languages (en-to-*, averaged over target languages), compared to an auto-encoder (en-to-en) and to untrained modeled with random weights (en-to-* rand). 
	  	The MT-trained representations improve with each layer, while the random representations degrade after layer 1. The auto-encoder representations also improve but are below the MT-trained ones. These results show that learning to \emph{translate} is important for obtaining good representations.
	  	}
	\label{fig:syn-accuracies2}
\end{figure}

%\todo{YB: I think the paragraph on random encoders can be removed because it seems disconnected with the rest of the paper. We could keep the auto-encoder results and modify the last paragraph accordingly. ND: Done}

One possible concern with these results is that they may be appearing because of the stacked RNN layers, and not necessarily due to the translation task. In the extreme case, perhaps even a random computation that is performed in stacked RNN layers would lead to improved performance in higher layers. This may be especially concerning when predicting relation labels, as this requires combining information about two words in the sentence. To verify that the actual translation task is important, we can look at the performance with random models, initialized in the same manner but not trained at all. Figure \ref{fig:syn-accuracies2} shows that higher layers in random networks generally generate worse representations.  %The syntactic dependencies are taken again from the Universal Dependencies dataset. \label{fn:un}} 
Layer 1 does improve the performance compared to layer 0 (word embeddings without contextual information) showing that some information is captured even in random models. However, after layer 1 the performance degrades drastically, demonstrating that higher layers in random models do not generate informative representations.

The experiment with random weights shows that training the NMT system is important. Does the actual translation task matter? Figure \ref{fig:syn-accuracies2} also shows the results using representations from English-to-English models, that is, an autoencoder scenario. 
%As in the previous tasks, 
%As in the machine translation models, representations from higher layers do not improve morphological tagging, but do improve the prediction of syntactic dependencies. However, 
There is a notable degradation in representation quality when comparing the autoencoder results to those of the machine translation models. For example, the best results for predicting syntactic dependencies with the autoencoder are around 80\% at layer 4. In contrast, the same layer in the translation models produces a score of 88\%. In general, the representations from the machine translation models are always better than those from the autoencoder, and this gap increases as we go higher in the layers. This trend is similar to the results on morphological and semantic tagging with representations from autoencoders that were reported previously. 

\subsection{Analysis}
In this section, we analyze two aspects of how information on syntactic dependencies is captured in different NMT layers: how different types of relations are represented and what the effect of head-modifier distance is.   
The results in this section are obtained using models trained on the United Nations corpus, as described in Section~\ref{sec:nmt-data}.

\subsubsection{Effect of Relation Type} \label{sec:syn-relation-type}

\begin{figure}[t]
\centering
		  \includegraphics[width=0.65\linewidth,trim={0 0.0cm 0 0},clip]{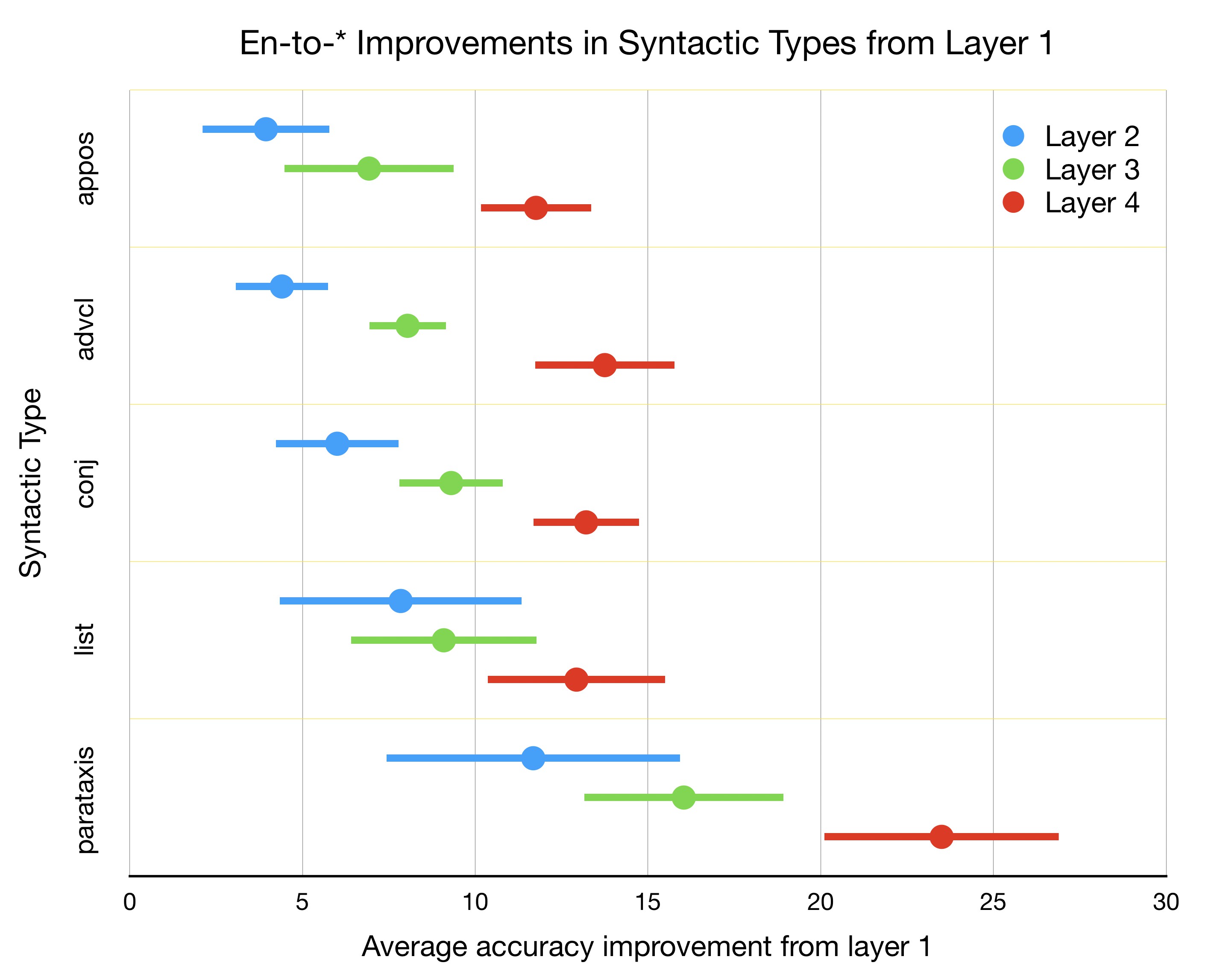}
	  	\caption{Improvement in accuracy of syntactic relation labeling with layers 2/3/4 compared to layer 1. The figure shows the 5 most improved relations when averaging results obtained with encoder representations from (word-based) NMT models trained on translating English to other languages. }
	\label{fig:syn-accuracies3}
\end{figure}

When are higher-layer representations especially important for syntactic relations? Figure \ref{fig:syn-accuracies3} breaks down the performance according to the type of syntactic relations. The figure shows the 5 relations that benefit most from higher layer representations.\footnote{The results shown are with English dependencies using NMT models trained on English to other languages, but the trends are similar for other language pairs.} %\todo{YB: see footnote. We can give the other translation direction too if we want.} 
The general trend is that the quality of the representation improves with higher layers, with up to 20$-$25\% improvement with representations from layer 4 compared to layer 1. The improvement is larger for certain relations: dependent clauses (\texttt{advcl, ccomp}),
loose relations (\texttt{list, parataxis}), and other typically long-range dependencies such
as conjunctions (\texttt{conj}) and appositions (\texttt{appos}). Core nominal arguments like subject (\texttt{nsubj}) and object (\texttt{obj}) also show consistent improvements with higher layers. Relations that do not benefit much from higher layers are mostly function words (\texttt{aux, cop, det}), which are local relations by nature, and the relation between a conjunct and the conjunction (\texttt{cc}), as opposed to the relation between two conjuncts (\texttt{conj}). These relations are local by nature and also typically less ambiguous. For example, the relation between a conjunction \texttt{and} and a noun is always labeled as \texttt{cc}, while a verb and a noun may have a subject or object relation.

\subsubsection{Effect of Relation Distance} 

\begin{figure}[t]
\centering
		  \includegraphics[width=0.65\linewidth,trim={0 0.0cm 0 0},clip]{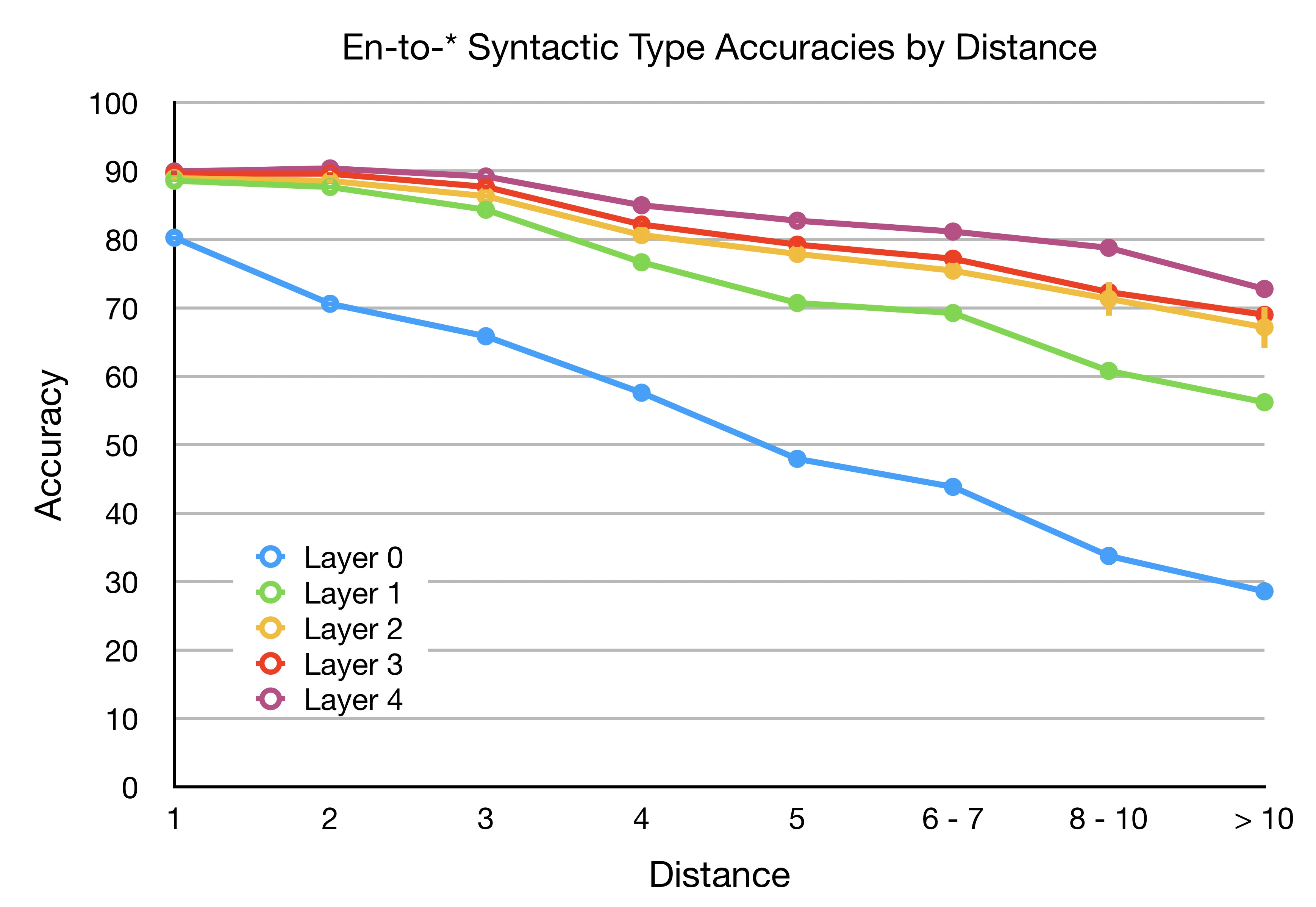}
	  	\caption{Results of syntactic relation labeling at different relation distances using encoder representations from different layers of (word-based) NMT models trained on translating English to other languages (averaged over target language). Long-distance relations are difficult to capture at all layers, but higher layers mitigate more of the drop in performance.
	  	}
	\label{fig:syn-accuracies4}
\end{figure}

In order to quantify the notions of global and local relations, let us consider relation distance. Figure \ref{fig:syn-accuracies4} shows the representation quality as a function of the distance between the words participating in the relation. %\todo{YB: again, the fig shows en-to-* but we can give also  *-to-En} 
Predicting long-distance relations is clearly more difficult than predicting short-distance ones. As the distance between the words in the relation grows, the quality of the representations decreases. When no context is available (layer 0, corresponding to word embeddings), the performance quickly drops with longer distance relations. The drop is more moderate in the hidden layers, but in low layers the effect of relation distance can still be as high as 25\%. Higher layers of the network mitigate this effect and bring the decrease down to under 5\%. Moreover, every layer is performing better than the previous one at each distance group. This indicates that higher layers are much better at capturing long-distance syntactic information.

\section{Semantics Results}
\label{sec:results-semantics}

We now study how information on meaning is captured in NMT models in the context of lexical semantic (SEM) tagging and semantic dependency labeling tasks %, introduced in \namecite{bjerva-plank-bos:2016:COLING} 
(refer to Section \ref{sec:semantics} for details on the tasks). We study %how this sort of semantic information is captured in the neural machine translation system by answering 
the following specific questions: (i) Do NMT systems learn informative semantic representations? (ii) Can a neural network model learn to map a sequence of subwords or character symbols to a meaning representation? %\todo{YB: this is risky, because ``meaning representation'' is associated with representing meaning of sentences ND: It could be words, phrases or sentences} 
(iii) What layers in the model learn more about semantic tags and relations? 

The experiments reported in this section on are  conducted mainly on English, as the semantic tagging task and dataset are recent developments that were initially only available in English. We also experiment and report results for German, for which a new semantic tagging dataset is being developed. However, as the German annotations are very sparse (see Section \ref{sec:supervised-data}), we performed a cross-fold evaluation when reporting results for German. For the semantic dependency labeling, we additionally used Czech data to strengthen the empirical evidence.

\begin{table}[t]
\centering
				\begin{tabular}{l l H r H r H}
					\toprule
				&	& \multicolumn{1}{H}{de} & \multicolumn{1}{c}{cs} & \multicolumn{1}{H}{ru} & \multicolumn{1}{c}{en}  & \multicolumn{1}{H}{fr} \\
					\midrule
					\multirow{3}{*}{Semantic Dependencies} & MT Classifier & -- & 87.8 & -- & 81.5  & --\\
                     & Majority & -- & 63.1 & -- & 57.3  & -- \\
                     & OOV Rate & -- & 12.1 &  -- & 6.3 &  -- \\ 
                     \midrule
                     & \namecite{bjerva-plank-bos:2016:COLING} & -- & -- & -- & 95.2  & -- \\ 
                    \multirow{2}{*}{Semantic Tags} & MT Classifier & -- & -- & -- & 93.4  & -- \\
                    	 & Majority & -- & -- & -- & 84.2  & -- \\
                   & OOV Rate & -- & --  & -- & 4.1  & -- \\ 
                    
					\bottomrule
				\end{tabular}
\caption{Local majority baseline (most frequent tag/label) and classification accuracy using encoder representations generated by the NMT models, on semantic tasks. The models are trained on translating Czech to English (cs column) or English to German (en column). The classifier results are far superior to the majority baseline, indicating that NMT representations contain a non-trivial amount of semantic information.
                }
                \label{tab:locMaj-Sem}
\end{table}

In this section we only report \emph{encoder-side} representation as the the analysis of \emph{decoder-side} representations requires parallel data to generate the hidden representations and no standard tools for annotating the data exist. %\todo{YB: not convincing, as in Morph we annotated the target side. We could in theory do the same here, with Bjerva's semantic tagger implementation or training a TNT tagger, which is what PMB did. Maybe we can argue that there's no standard tool for this task yet.}
Table \ref{tab:locMaj-Sem} shows the results. The classifier achieves 91.4\% on the semantic tagging task and 85\% and 80\% on the task of semantic labeling for Czech and English respectively. All results are significantly better than the local majority baseline
%Figure \ref{fig:sem-accuracies} show the results. The classifiers achieve accuracy above 90\% in English and 85\% in German. These numbers are much better than the local majority baseline %\footnote{Selecting the most frequent tag for each word and the most frequent global tag for the unknown words.} %\alery{YB: we already defined this baseline} 
%(91.4 vs.\ 84.2 in English and 85.3 vs.\ 78.7 in German), 
suggesting that NMT representations learn substantial semantic information.

%\begin{figure}[t]
%\centering
%		  \includegraphics[width=0.65\linewidth,trim={0 0.5cm 0 0},clip]{figs/semantics-results.eps}
%	  	\caption{Semantic tagging for different units in English (EN) and German (DE).}
%	\label{fig:sem-accuracies}
%\end{figure}

\subsection{Impact of Translation Unit on Learning Semantics}

\begin{figure*}[ht]
\centering
\includegraphics[width=0.7\linewidth]{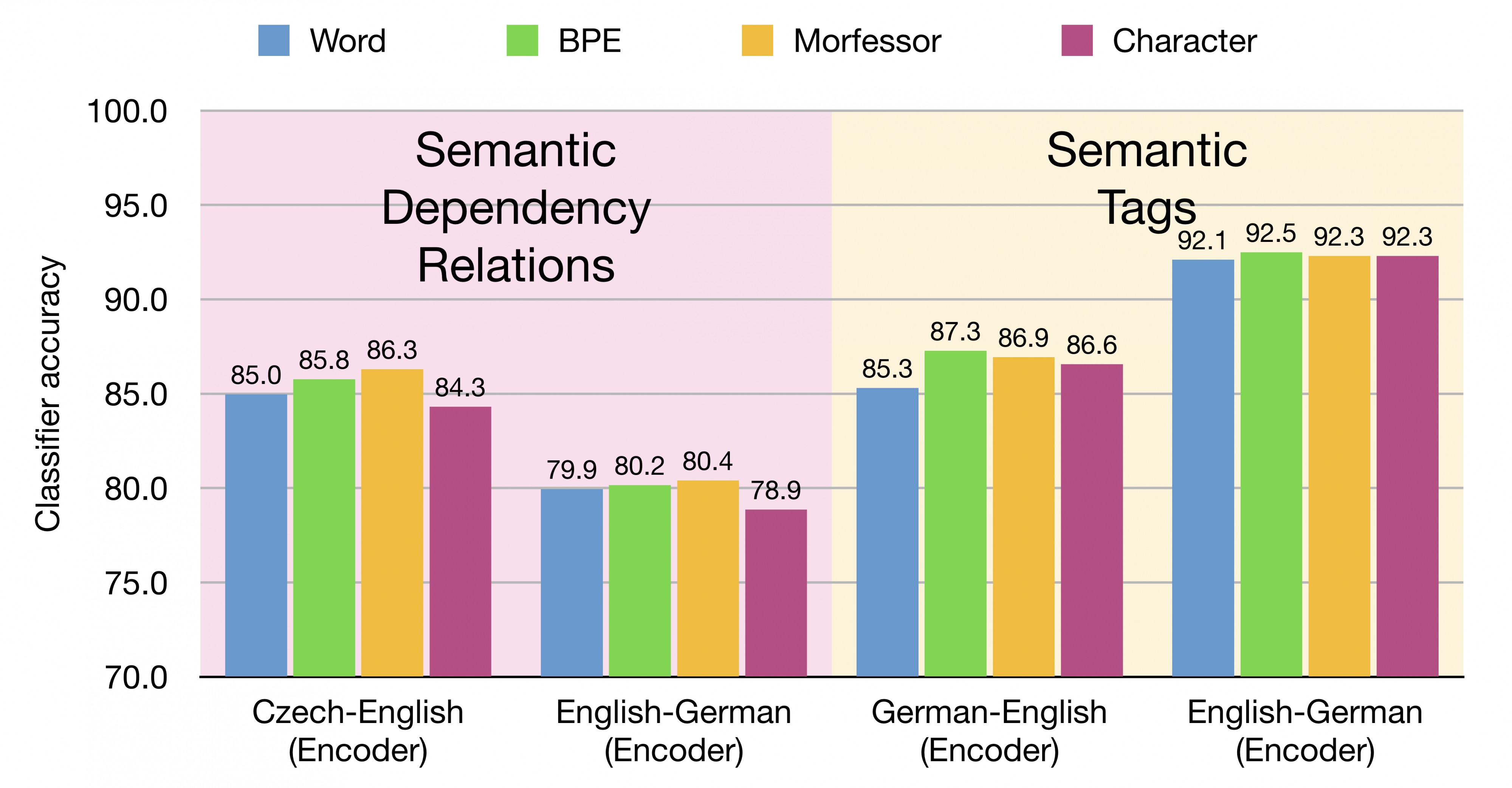}
\caption{Semantic tagging and dependency labeling results using representations of NMT models trained with  different translation units on the source-side the target-side is always BPE.}
%Here the representations are taken from the top encoding layer \alert{of 4-layer models}.  

\label{fig:semantics-results-by-repr}
\end{figure*}

Next we investigate if the representations learned from characters or subword units 
can effectively model semantic information. We trained classifiers using the representations generated from different NMT models that were trained using character or subword units (BPE and Morfessor). Figure \ref{fig:semantics-results-by-repr} summarizes the results on the semantic dependency labeling task and the semantic tagging task. 
In the semantic dependency labeling task, the character-based models perform significantly worse compared to the word-based and subword-based counterparts. We found using subword-based representations, particularly morpheme-based segmentation, to give better performance in most scenarios. 
These results are in contrast with morphological tagging results, where character-based representations were consistently and significantly better compared to their subword counterparts. On comparing the prediction results between subword and character-based representations, we found that in many cases, character-based models failed to predict the label correctly when the head and modifier words are further apart, i.e., in the case of long-distance dependencies. However, this was not always true as in some cases character-based models were able to correctly predict the dependency label for a head that was 12 words apart.

%\alert{ The semantic dependency relations require modeling of words that are far from each other in a sentence. This requires to learn information across long distances. Whereas morphological tagging is mainly dependent on the current word and it's nearby words. The character-based models suffer from learning long range dependencies which is evident from their low performance on the semantic dependency relations task. However, they are able to learn more local properties as in the case of the morphological tagging task. } %\todo{YB: this makes sense, but begs the question of analyzing performance with different representations by dependency distance. ND: We tried this but could not find any conclusive evidence. Sometimes char does good even with long distances and poorly on short distances}

%\todo{YB: can we discuss this more? Can we investigate why BPE is so successful in German? Is it because of noun compounds? Maybe look at accuracy per (coarse) SEM tag}
%A more clear picture is depicted 
On semantic tags, subword-based (BPE and Morfessor) representations and character-based representation achieve comparable results for English. 
However, for German, BPE-based representations performed better than the other representations.

\subsection{Effect of Network Depth}

%\todo{YB: Fahim, can you move what is now figure 9b (semantic dependency results per layer) to here, and make table 9 (sem tagging results per layer) into a similar figure, and have them both as subfigures?}

We found the representations learned in the lower encoding layer to perform better on the task of morphological tagging. Here we investigate the quality of representations at different encoding layers, from the perspective of semantic properties. 

Concerning lexical semantic tagging, as  Table \ref{tab:results-semtags-4layers} shows, representations from layers 2 and 3 do not consistently improve  performance above layer 1. However, representations from layer 4 lead to small but significant improvement with all target languages, according to the approximate randomization test.\footnote{These results are obtained using models trained on the United Nations multi-parallel corpus.} %\cite{sigf06}. %\todo{YB: move this ref earlier} 
We observed a similar pattern in the case of semantic dependency labeling task (see Figure \ref{fig:layer-effect-all-langs-semantics}), where higher layers (layer 4 in the case of Czech-English and English-German and layer 3 in the case of English-Czech) gave better accuracy. Intuitively, higher layers have a more global perspective because they have access to higher representations of the word and its context, while lower layers have a more local perspective. Layer 1 has access to context but only through one hidden layer which may not be sufficient for capturing semantics. It appears that higher representations are necessary for learning even relatively simple lexical semantics and, especially, predicate-argument relations.

\subsection{Analysis of Lexical Semantics}

In this section, we analyze three aspects of lexical semantic information as represented in the semantic tagging datasaet. First, we categorize semantic tags into coarse-grained categories and compare the classification quality within and across categories. 
Second, we perform a qualitative analysis of discourse relations and when they are better represented in different NMT layers.
Third, we compare the quality of encoder representations when translating into different target languages. 
The results in this section are obtained using models trained on the United Nations corpus, as described in Section~\ref{sec:nmt-data}.

\subsubsection{Semantic Tag Level Analysis} \label{sec:sem-tags-analysis}

\begin{figure}[t]
\centering
\includegraphics[width=0.65\linewidth]{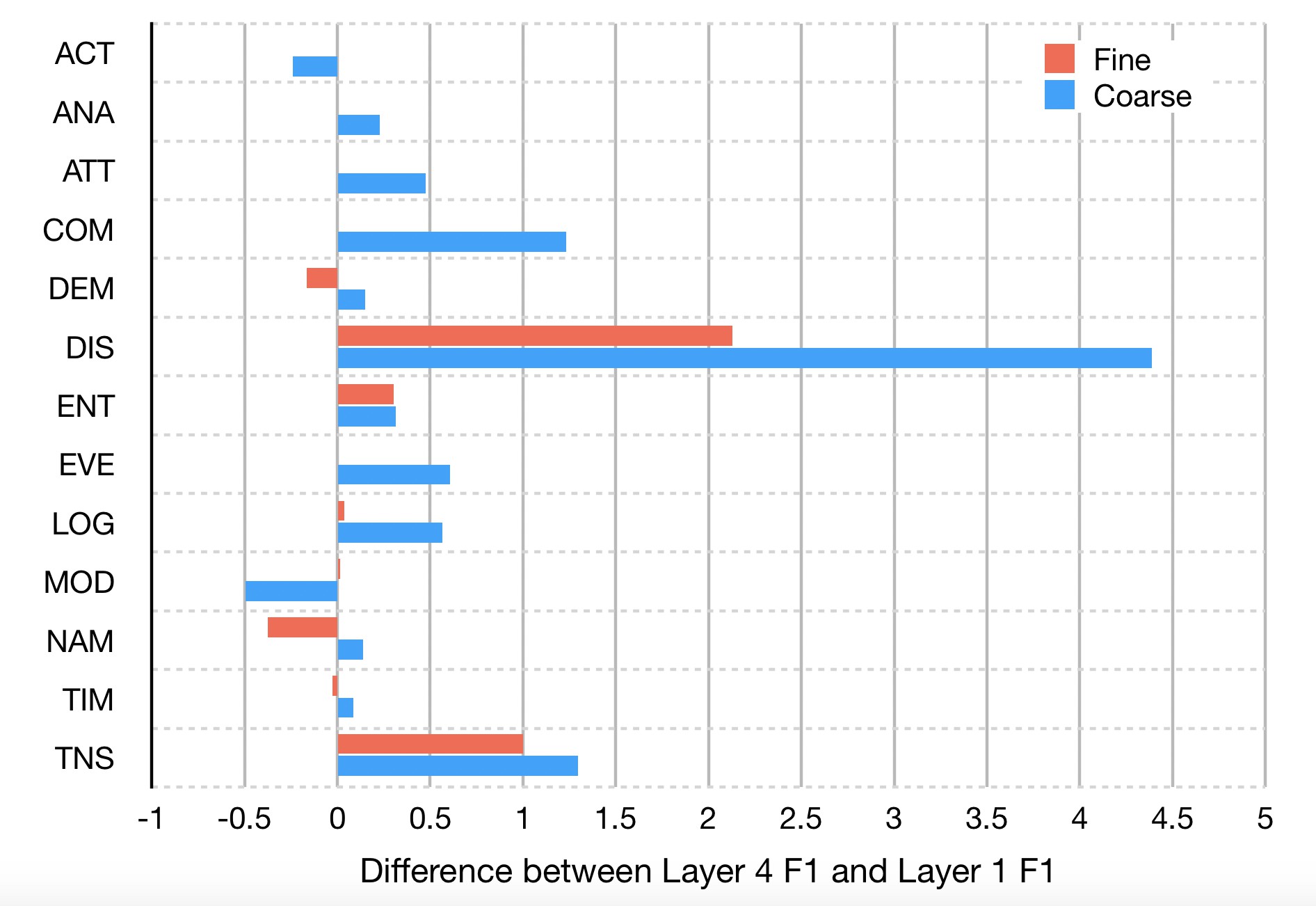}
\caption{Difference in semantic tagging F$_1$ when using representations from layer 4 compared to layer 1, showing F$_1$ when directly predicting coarse tags (blue) and when predicting fine-grained tags and averaging inside each coarse tag (red). 
The representations are taken from encoders of (word-based) NMT models trained on translating English to other languages (averaged over target language).  
}
\label{fig:coarse-layer1-4-f1}
\end{figure}

The SEM tags are grouped in coarse-grained categories such as events, names, time, and logical expressions. Figure \ref{fig:coarse-layer1-4-f1} shows the change in F$_1$ score (averaged over target languages) when moving from layer 1 to layer 4 representations. The blue bars describe the differences per coarse tag when directly predicting  coarse tags. The red bars show the same differences when predicting fine-grained tags and micro-averaging inside each coarse tag. The former shows the differences between the two layers at distinguishing among coarse tags. The latter gives an idea of the differences when distinguishing between  fine-grained tags within a coarse category. The first observation is that in the majority of cases there is an advantage for classifiers trained with layer 4
representations, i.e., higher layer representations are better suited for learning  the SEM tags, at both coarse and fine-grained levels.

Considering specific tags, higher layers of the NMT model are especially 
better at capturing semantic information such as \emph{discourse relations} (\texttt{DIS} tag:  subordinate vs.\ coordinate vs.\ apposition 
relations), semantic properties of nouns (\emph{roles} vs.\ \emph{concepts}, within the \texttt{ENT} tag), \emph{events} and \emph{predicate tense} (\texttt{EVE} and \texttt{TNS} tags), \emph{logic relations} and \emph{quantifiers} (\texttt{LOG} tag: 
disjunction, conjunction, implication, 
existential, universal, 
etc.), and \emph{comparative constructions} (\texttt{COM} tag: equatives, comparatives, and superlatives). These  examples represent semantic concepts and relations that require a level of abstraction going beyond the lexeme or word form, and thus might be better represented in higher layers in the deep network.

\begin{figure*}[t]
\centering
\footnotesize
\begin{tabular}{l|cc|p{10cm}}
\toprule
& L1 & L4 & \\
\midrule
1 & REL & \emph{SUB} & Zimbabwe 's President 
Robert Mugabe 
has freed three men who were jailed for murder and sabotage \underline{\emph{as}} they battled South Africa 's anti-apartheid African National Congress in 1988 . \\
2 & REL & \emph{SUB} & The military says the battle erupted \underline{\emph{after}} gunmen fired on U.S. troops and Afghan police investigating a reported beating of a villager . \\
3 & IST & \emph{SUB} & Election authorities had previously told Haitian-born Dumarsais Simeus that he was not eligible to run \underline{\emph{because}} he holds U.S. citizenship . \\	
\midrule
4 & AND & \emph{COO} & Fifty people representing 26 countries took the Oath of Allegiance this week ( Thursday ) \underline{\emph{and}} became U.S. citizens in a special ceremony at the Newseum in Washington , D.C. \\
5 & AND & \emph{COO} & But rebel groups said on Sunday they would not sign \underline{\emph{and}} insisted on changes . \\
6 & AND & \emph{COO} & A Fox asked him , `` How can you pretend to prescribe for others , when you are unable to heal your own lame gait \underline{\emph{and}} wrinkled skin ? '' \\
\midrule
7 & NIL & \emph{APP} & But Syria 's president \underline{\emph{,}} Bashar al-Assad , has already rejected the commission 's request [...] \\ %to interview him . \\
8 & NIL & \emph{APP} & Hassan Halemi \underline{\emph{,}} head of the pathology department at Kabul University where the autopsies were carried out , said hours of testing Saturday confirmed [...] \\ %the identities of teachers Jun Fukusho and Shinobu Hasegawa . \\
9 & NIL & \emph{APP} & Mr. Hu made the comments Tuesday during 
a meeting with Ichiro Ozawa \underline{\emph{,}} the leader of Japan 's main opposition party . \\
\midrule
10 & \emph{AND} & COO & [...] %In Washington , D.C. , 
abortion opponents will march past the U.S. Capitol \underline{\emph{and}} end outside the Supreme Court . \\
11 & \emph{AND} & COO & Van Schalkwyk said no new coal-fired power stations would be approved unless they use technology that captures \underline{\emph{and}} stores carbon emissions . \\
12 & \emph{AND} & COO & A MEMBER of the Kansas Legislature meeting a Cake of Soap was passing it by 
without recognition , 
but the Cake of Soap insisted on stopping \underline{\emph{and}} shaking hands . \\
\bottomrule
\end{tabular}
\caption{Examples of cases of disagreement between layer 1 (L1) and layer 4 (L4) representations when predicting semantic tags. The correct tag is \emph{italicized} and the relevant word is  \underline{\emph{underlined}}.}
\label{fig:error-analysis}
\end{figure*}

%One negative example that stands out in Figure~\ref{fig:coarse-layer1-4-f1} is the prediction of the \texttt{MOD} tag,  corresponding to \emph{modality} (necessity, possibility, and negation). It seems that such semantic concepts should be better represented in higher layers following our previous hypothesis. Still, layer 1 is better than layer 4 in this case. One possible explanation is that words  tagged as \texttt{MOD} form a closed class, with only a few and mostly unambiguous words (``no'', ``not'', ``should'', ``must'', ``may'', ``can'', ``might'', etc.). It is enough for the classifier to memorize these  words in order to predict  this class with high F$_1$, and this is something that occurs better in lower layers. One final case worth mentioning is the \texttt{NAM} category, which stands for different types of named entities (person, location, organization, artifact, etc.). In principle, this seems a clear case of semantic abstractions  suited for higher layers, but the results from layer 4 are not significantly better than those from layer 1. This might be signaling a limitation of the NMT system at learning this type of semantic classes. Another factor might be the fact that many named entities are out of vocabulary words for the NMT system.

\subsubsection{Analyzing Discourse Relations}

Now we analyze specific cases of disagreement between predictions using representations from layer 1 and layer 4. We focus on discourse relations, as they show the largest improvement when going from layer 1 to layer 4 representations (\texttt{DIS} category in Figure~\ref{fig:coarse-layer1-4-f1}). 
Intuitively, identifying discourse relations requires a relatively large context so it is expected that higher layers would perform better in this case. 

There are three discourse relations in the SEM tags annotation scheme: subordinate (\texttt{SUB}), coordinate (\texttt{COO}), and apposition (\texttt{APP}) relations. For each of those, Figure~\ref{fig:error-analysis} (examples 1--9) shows the first three cases in the test set where layer 4 representations correctly predicted the tag but layer 1 representations were wrong. Examples 1--3 have subordinate conjunctions (\emph{as}, \emph{after}, \emph{because}) connecting a main and an embedded clause, which layer 4 is able to correctly predict. Layer 1 mistakes these as attribute tags (\texttt{REL}, \texttt{IST}) that are usually used for prepositions. In examples 4--5, the coordinate conjunction \emph{and} is used to connect sentences/clauses, which layer 4 correctly tags as \texttt{COO}. Layer 1 wrongly predicts the tag \texttt{AND}, which is used for conjunctions connecting shorter expressions like words (e.g., ``murder \emph{and} sabotage'' in example 1). Example 6 is probably an annotation error, as \emph{and} connects the phrases ``lame gait'' and ``wrinkled skin'' and should be tagged as \texttt{AND}. In this case, layer 1 is actually correct. 
In examples 7--9, layer 4 correctly identifies the comma as introducing an apposition, while layer 1 predicts \texttt{NIL}, a tag for punctuation marks without semantic content (e.g., end-of-sentence period).  
As expected, in most of these cases identifying the discourse function requires a fairly large context.

Finally, we show in examples 10--12 the first three occurrences of \texttt{AND} in the test set, where layer 1 was correct and layer 4 was wrong. Interestingly, two of these (10-11) are clear cases of \emph{and} connecting clauses or sentences, which should have been annotated as \texttt{COO}, and the last (12) is a conjunction of two gerunds. The predictions from layer 4 in these cases thus appear justifiable. 

\subsubsection{Effect of Target Language}

\begin{table}[t]
\centering
%\begin{tabular}{l|lllll|l} %|l}
\begin{tabular}{l|p{0.7cm}p{0.7cm}p{0.7cm}p{0.7cm}|p{0.7cm}} %|l}
\toprule
$k$ & \multicolumn{1}{c}{Ar} & \multicolumn{1}{c}{Es} & \multicolumn{1}{c}{Fr} & \multicolumn{1}{c}{Ru}  & \multicolumn{1}{|c}{En} \\ %& \multicolumn{1}{c}{Avg} \\
\midrule
\multicolumn{6}{c}{SEM Tagging Accuracy} \\
\midrule
0 & 81.9$^{*}$ & 81.9$^{*}$ & 81.8$^{*}$ & 81.8$^{*}$  & 81.2$^{*}$ \\ %& 81.8 \\
1 & 87.9 & 87.7 & 87.8 & 87.9  & 84.5 \\ %& 87.8 \\
2 & 87.4$^{*}$ & 87.5$^{*}$ & 87.4$^{*}$ & 87.3$^{*}$  & 83.2$^{*}$ \\ %& 87.4 \\
3 & 87.8 & 87.9$^{*}$ & 87.9$^{**}$ & 87.3$^{*}$ &  82.9$^{*}$ \\ %& 87.6 \\
4 & 88.3$^{*}$ & 88.6$^{*}$ & 88.4$^{*}$ & 88.1$^{*}$  & 82.1$^{*}$ \\ %& 88.2 \\
\midrule
\multicolumn{6}{c}{BLEU} \\
\midrule
& 32.7 & 49.1 & 38.5 & 34.2  & 96.6 \\ %& \\
\bottomrule
\end{tabular}
\caption{Semantic tagging accuracy  using features  from the $k$-th encoding layer
of 4-layered NMT models  trained with different target languages. 
``En'' column is an English autoencoder.
BLEU scores are given for reference. Statistically significant differences from layer 1  
are shown at $p<0.001$$^{(*)}$ and $p<0.01$$^{(**)}$. See text for details. 
}
\label{tab:results-semtags-4layers}
\end{table}

Does translating into different languages make the  NMT system learn different  source-side representations? We previously found a fairly consistent effect of the target language on the quality of encoder representations for POS and morphological tagging, with differences of $\sim$2-3\% in accuracy. Here we examine if such an effect exists in SEM tagging. We trained 4-layered English-to-\{Arabic,Russian,French,Spanish,English\} models with the multi-parallel UN corpus. Table \ref{tab:results-semtags-4layers} shows results using features obtained by training NMT systems on different target languages (the English source remains fixed).  There are very small differences with different target languages ($\sim$0.5\%). While the differences are small, they are mostly statistically significant. %\todo{YB: this is the first time we mention stat.\ significance, so we should explain how we calculate it. Also, it's confusing that we only do this for part of the experiments} 
For example, at layer 4, all the pairwise comparisons with different target languages are  statistically significant ($p<0.001$). Again, we note that training an English autoencoder results in much worse representations compared to MT models. In contrast, the autoencoder has excellent sentence recreation capability (96.6 BLEU). This indicates that learning to translate (to any foreign language) is important for obtaining useful representations for semantic tagging, as it is for morphological tagging. 
%\todo{YB: in the IJCNLP paper, we had a paragraph explaining that the small differences are due to a larger training dataset, not because of SEM vs Morph. I think this is important to note. ND: I am not convinced that this is due to training dataset. In my recent experiments where I trained bilingual models with 2.5M sentences, I still found differences in Morph to be more}

\subsection{Analysis of Semantic Dependencies}

In the previous sections, we analyzed how the layer depth impacts the representations from the perspective of specific syntactic relations (Section~\ref{sec:syn-relation-type}) and lexical semantic tags (Section~\ref{sec:sem-tags-analysis}). We found that higher layers tend to better represent properties that are more global, loose, and abstract compared to lower layers.  Does the same hold for semantic dependencies? 

% \begin{figure}
%   \centering
% 	\includegraphics[width=0.65\linewidth]{new-figs/types_semantics_en_to_all_top5.png}
% 	\caption{Improvement in accuracy of semantic relation labeling with layers 2/3/4 compared to layer 1. The figure shows the 5 most improved relations when averaging results obtained with encoder representations from (word-based) NMT models trained on translating English to other languages. }
% 	\label{fig:semdeprel-types}
% \end{figure}

\begin{figure}
  \centering
	\includegraphics[width=\linewidth]{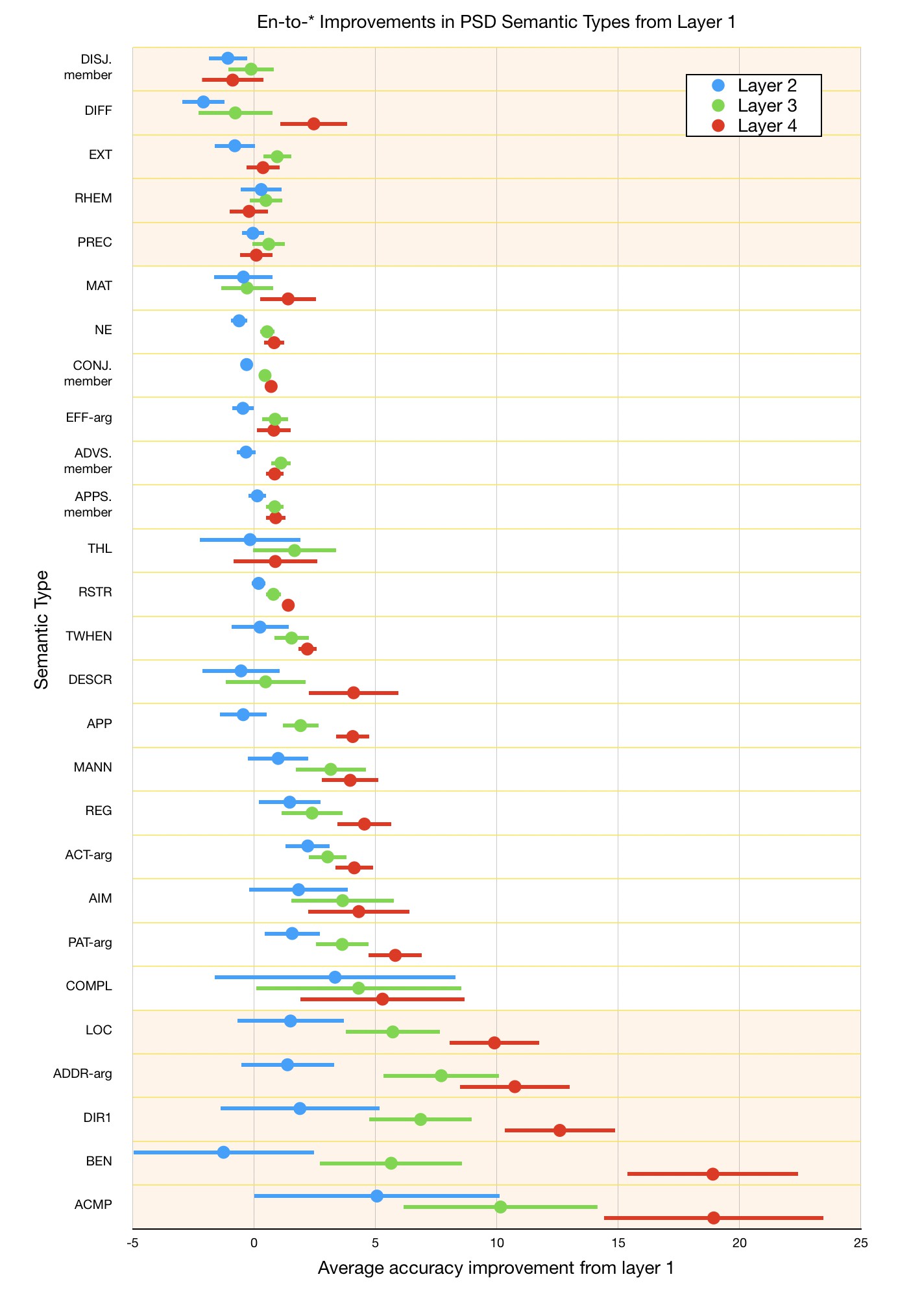}
	\caption{Improvement in accuracy of semantic relation labeling with layers 2/3/4 compared to layer 1, when averaging results obtained with encoder representations from (word-based) NMT models trained on translating English to other languages.  The 5 least/most improved relations are highlighted.}
	\label{fig:semdeprel-types}
\end{figure}

%Figure~\ref{fig:semdeprel-types} shows the top 5 relations that benefit most from higher layer representations.
Figure~\ref{fig:semdeprel-types} shows the improvement from higher layer representations. The 5 most improved relations are highlighted. 
 The following are examples from the PSD manual~\cite{tectogrammatical-reference-book}. We give in parentheses the average distance in words between head and modifier, per relation. Four of the top five relations are looser kinds of relations that syntactically correspond to adjuncts:
  accompaniment (\texttt{ACMP}; ``He works without his glasses''; distance 6.93);
  to whose advantage something happens (\texttt{BEN}; ``He did it for their sake''; distance 4.40);   direction (\texttt{DIR1}; ``He made a step from the wall''; distance 4.41); and location (\texttt{LOC};  ``a match in a foreign country''; distance 4.77);
  The fifth is an addressee argument (\texttt{ADDR-arg}; ``He gave the child a toy.''; distance 3.69). 
  These relations also have longer distances between head and modifier compared to the overall average distance (3.34 words).

  In contrast, the relations that benefit the least (highlighted at the top of Figure~\ref{fig:semdeprel-types}) include  a disjunctive that captures the relation between the disjunction ``or'' and a word in a list  (\texttt{DISJ.member}; distance 2.22); expressing difference (\texttt{DIFF}; ``The goods were delivered four days later''; distance 2.54 ); a degree specifier for expressing extent (\texttt{EXT}; e.g., the relation between ``about'' or ``almost'' and a quantity; distance 1.87); a rhematizer that often connects a negation word to its negated verb (\texttt{RHEM}; ``Cray Research did not want to...'', example from the PSD dataset; distance 2.17); %a temporal expression of duration (\texttt{THL}; ``He wrote it in two hours''; distance 3.61); and an argument that is the effect of an event (\texttt{EFF-arg}; ``He considered Paul a professional''; distance 7.71). 
   and linking the clause to the preceding text (\texttt{PREC}; ``Hence, I'm happy''; distance 7.73). 
  Of these, the first three actually drop in performance when using (some of the) higher layer representations, which may be explained by their more local, tight relation. This also partly accords with the distances, which are below average for the bottom 4. The \texttt{PREC} relation is an exception, with a relatively large distance but almost no benefit from higher layers. This can be explained by its use for words linking the clause to the preceding text, which are limited and easy to memorize (``However'', ``Nevertheless'',  ``Moreover'', etc.). As these cases are assigned the main verb as a the \texttt{PREC} head, they  may span over large distances, but their closed-class nature enables simple memorization even at low layers. 
  However, relation distance does not explain the entire benefit from higher layers, as some of the most distant relations are not the ones that benefit most from higher layers. Still, similar to the syntactic dependencies, the representations from higher layers benefit more in case of looser, less tight semantic relations.

\section{Comparison Against Multilingual Models}
\label{sec:multilingual}

%\begin{table}[t]
%\small
%			\begin{tabular}{l|c|c|c|c|c|c|c|c}
%				\toprule
%				& \multicolumn{1}{c}{de-en} & \multicolumn{1}{c}{cs-en} & \multicolumn{1}{c}{fr-en} & \multicolumn{1}{c}{es-en} & \multicolumn{1}{c}{en-de} & \multicolumn{1}{c}{en-de} & \multicolumn{1}{c}{en-fr} & \multicolumn{1}{c}{en-es} \\
%				\midrule
%				\textbf{bi}  & 33.4 & 29.6 & 41.4 & 37.8 & 29.3 & 22.3 & 40.8 & 37.2 \\
 %               \textbf{multi}  & 32.9 & 30.0 & 41.4 & 37.2 & 27.6 & 21.8 & 39.5 & 35.3 \\
  %              \textbf{multi*} & 34.7 & 31.4  & 43.5 & 38.8 & 30.0  & 23.7 & 42.4 & 37.6 \\
	%			\bottomrule
	%		\end{tabular}
     %   \caption{BLEU scores across language pairs - bi = bilingual model, multi = multilingual model, multi* = rich multilingual model.}
	%	\label{tab:multlingual-bleu-scores}
 % \end{table}

Languages share orthography, morphological patterns, and grammar. Learning to translate between several pairs simultaneously can help improve the translation quality of the underlying languages and also enable translation for language pairs with little or no parallel data. \citet{johnson2016google} exploited a remarkably simple idea of appending a language code to each training sentence, in a shared encoder-decoder framework. The models are trained with all multilingual data consisting of multiple language pairs at once.
%\todo{YB: add more information on the multilingual training setup that we have ND: Done} 
In their projection  of a small corpus of 74 triples of semantically identical cross-language (Japanese, Korean, English) phrases to 3D space via t-SNE, they found preliminary evidence that the shared architecture in their multilingual NMT systems learns a universal interlingua. We use this idea with our machinery to investigate how effective the multilingual representations are in learning morphology, syntax, and semantics, compared to their corresponding bilingual models.
%\todo{YB: add more details on their finding? ND: Done} 

\begin{figure}
  \centering
	\includegraphics[width=1\linewidth]{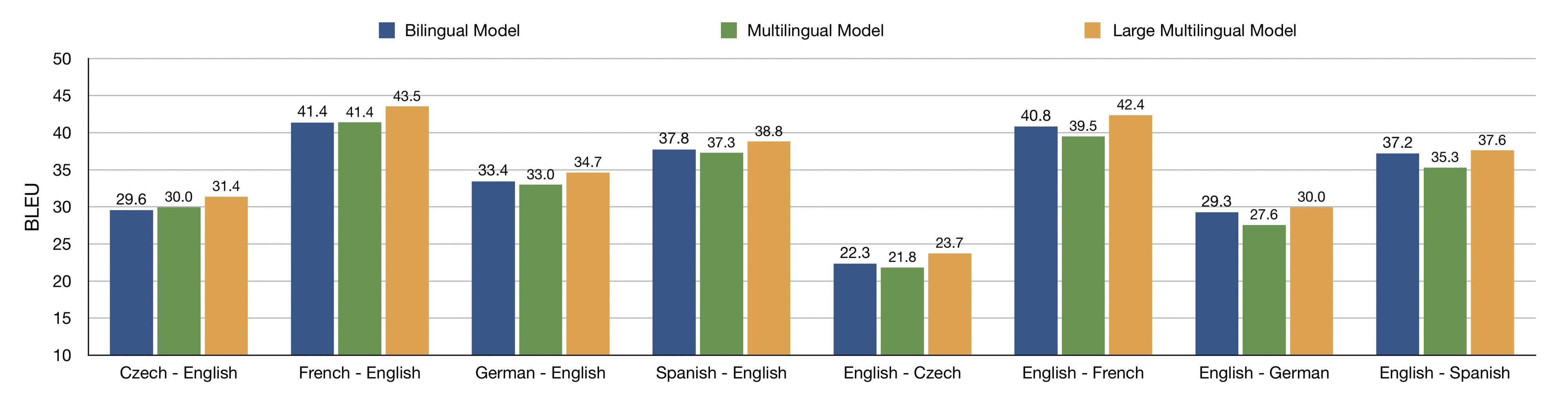}
	\caption{Comparing bilingual and multilingual systems in terms of translation quality. The multilingual models are many-to-one (*-to-English) or one-to-many (English-to-*) during training, but run on a specific language pair during testing.
	}
	\label{fig:multilingual-bleu}
\end{figure}

We trained 4-layer multilingual systems with many-to-one and one-to-many configurations with English on one side (encoder/decoder) and German, Spanish, French, and Czech on the other side. The models were trained with BPE subword units. We trained two versions of the multilingual model, one with the same parameters as the bilingual models and another with word embeddings and LSTM states with 1024 dimensions (as opposed to the default 500 dimensions). Our goal was to investigate the effect of increasing model parameters on translation quality and representation quality in terms of the understudied linguistic phenomenon. The systems were trained on NEWS+TED+Europarl, approximately 2.5M sentences per language pair, and a total of 10M sentences for training the multilingual models.

%%\todo{YB: give more details on the multilingual models, such as which language pairs exactly are contained. And what are the bilingual baselines. ND: Done}

\begin{table}[t]
\small
			\begin{tabular}{l cccccccc}
				\toprule
				& \multicolumn{4}{c}{cs-en} & \multicolumn{4}{c}{en-cs} \\
                \cmidrule(lr){2-5} \cmidrule(lr){6-9}
 & \multicolumn{1}{c}{BLEU} & Morphology & Syntax & Semantics & \multicolumn{1}{c}{BLEU} & Morphology & Syntax & Semantics \\
				\midrule
				bi  & 29.6 & 78.9 & 90.6 & 86.4 & 22.3 & 81.8 & - & 78.8 \\
                bi*  & 29.8 & 77.7 & 91.2 & 84.6  & 22.6 & 82.5 & - & 79.3 \\
                multi  & 30.0 & 82.1 & 91.8 & 86.8 & 21.8 & 81.8 & - & 81.2 \\
                multi*  & 31.4 & 84.0 & 88.7 & 87.8 & 23.7 & 84.6 & - & 81.4 \\

				\bottomrule
			\end{tabular}
        \caption{Comparing bilingual and multilingual Czech$\leftrightarrow$English models across translation quality and classifying different linguistic properties.  bi = bilingual model, bi* = larger bilingual model, multi = multilingual model, multi* = larger multilingual model. Larger models have twice the number word embedding and LSTM state dimensions. The larger models improve the substantially the multilingual systems, but only slightly improve the bilingual systems. Morphology = Morphological Tagging, Syntax = Syntactic Dependency Labeling, Semantics = Semantic Dependency Labeling
        }
		\label{tab:rich-Czech}
  \end{table}

Figure \ref{fig:multilingual-bleu} shows BLEU scores comparing bilingual and multilingual translation systems across different language pairs. %\todo{YB: I'm confused about the meaning of multilingual in a specific X-Y pair ND: The model is multilingual, we just input one language to it and tell it to output a particular language} 
We see that the many-to-one multilingual system, i.e., *-to-English, is on par or slightly behind bilingual systems when trained with the same number of parameters as the bilingual models. In contrast, the one-to-many multilingual system, i.e., English-to-*, is significantly worse compared to its bilingual counterparts. The reason for this discrepancy could be that generation is a harder task compared to encoding, especially when translating into morphologically-rich languages: %\todo{I'm not following the line of reasoning here. Harder than what? We should be comparing multi to bi, right? Also the transition into talking about out-of vs from English is confusing ND: I am discussing what *-to-En models are on par with the bilingual models, but En-to-* models are not. It has traditionally been the case that translating into English is easier than translating out of English especially into a morphologically rich language} 
An average difference of -1.35 is observed when translating out-of-English compared to -0.13 when translating into-English. The  larger %\todo{YB: I'd use ``larger'' instead of ``richer'', which isn't well defined} 
multilingual models (with twice as many parameters) restored the baseline performance, in fact showing significant improvements in many cases. %On average the into-English systems improved over bilingual systems by +1.6 BLEU scores and out-of-English systems improved by +1.0.
We also trained two of the bilingual baselines (Czech $\leftrightarrow$ English) by doubling the parameters. While the large multilingual system gave an improvement of +1.4 (see Table \ref{tab:rich-Czech}) over the baseline multilingual system by doubling the parameters size, the bilingual system only obtained an improvement of +0.2 by increasing the model size. A similar pattern was observed in the opposite direction where increasing the model size gave a BLEU improvement of +1.9 in the multilingual system, but only +0.3 in the case of the bilingual system. These results show that multilingual systems benefit from other language pairs being trained in tandem and suggest that the underlying representations are richer than the ones trained using bilingual models. We now proceed to analyze these representations in light of the understudied linguistic properties.

\begin{figure}
	\centering
	\includegraphics[width=1\linewidth]{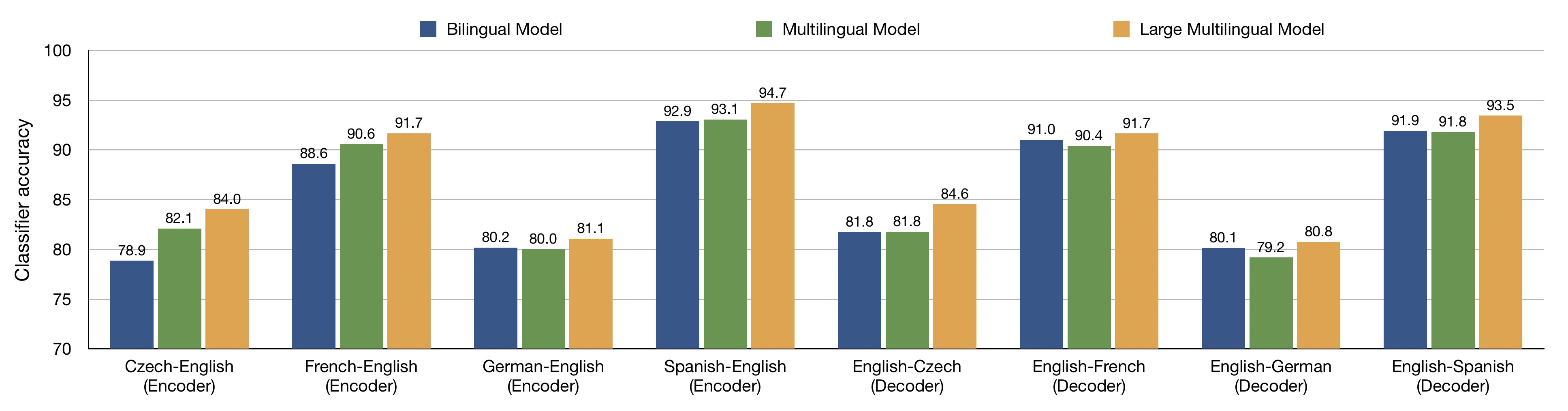}
	\caption{Comparing bilingual and multilingual models in terms of morphological tagging accuracy.  Same-size multilingual models benefit the encoder representations compared to bilingual models, but not decoder representations; larger multilingual models benefit also the decoder.}
	\label{fig:multilingual-morph}
\end{figure}

\begin{figure}[t]
\centering
  \begin{subfigure}{0.49\textwidth}
    \centering
	\includegraphics[width=1\linewidth]{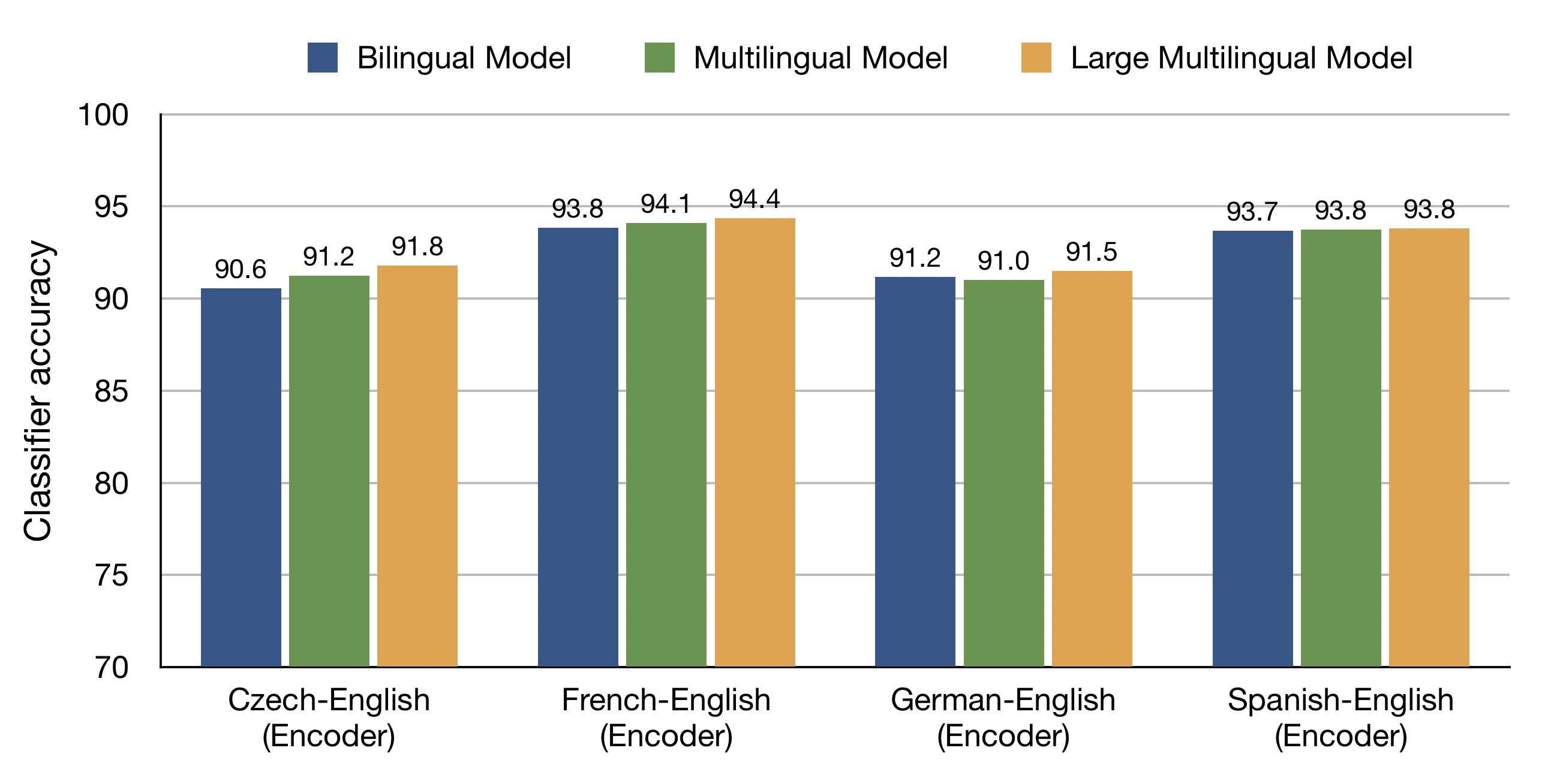}
	\caption{Syntactic dependency labeling. }
	\label{fig:multilingual-syntax}
  \end{subfigure}
  \begin{subfigure}{0.49\textwidth}
   \centering
	\includegraphics[width=0.8\linewidth]{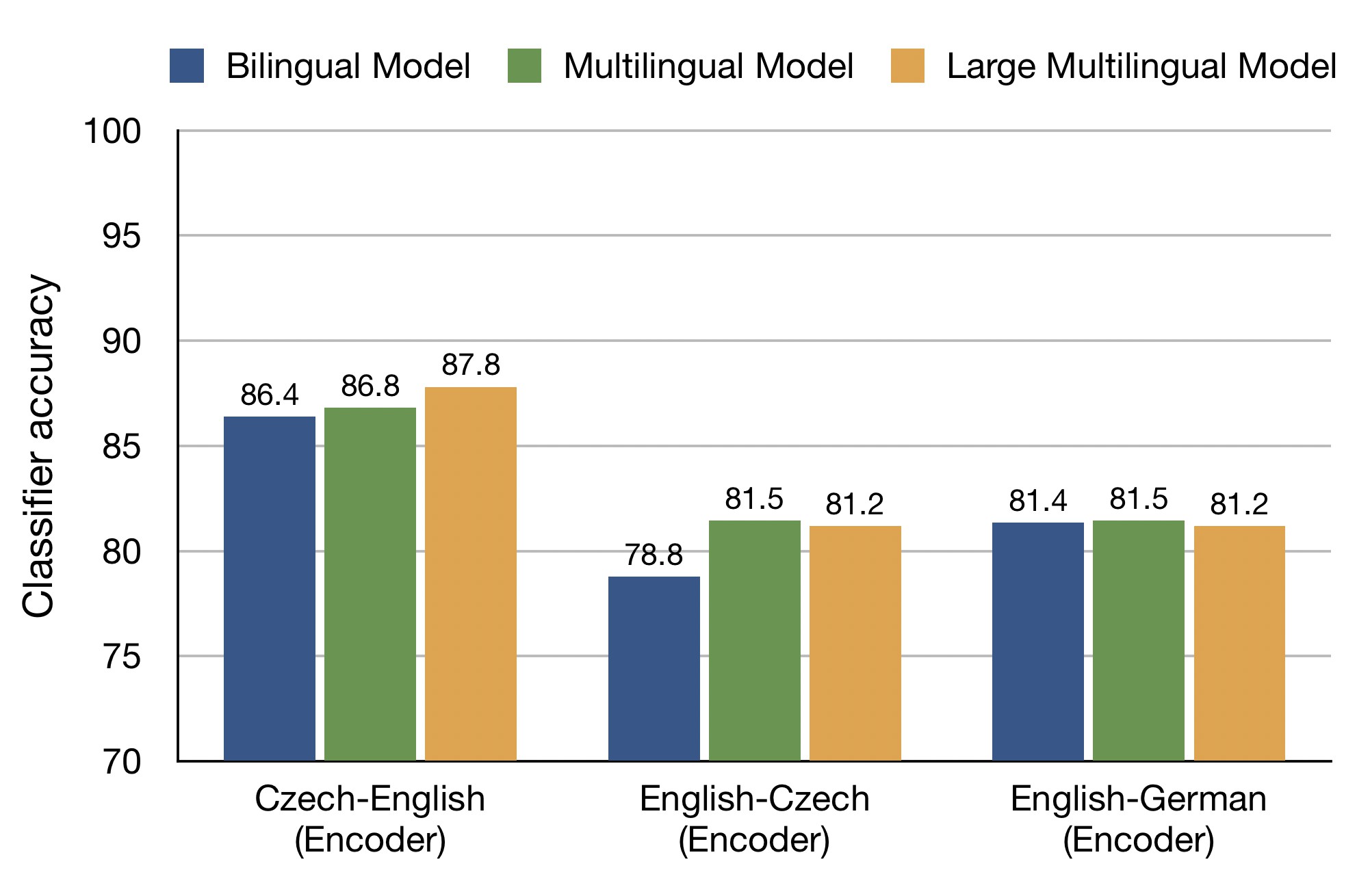}
	\caption{Semantic dependency labeling. }
	\label{fig:multilingual-semantics}
  \end{subfigure}  
  \caption{Comparing bilingual and multilingual models with syntactic and semantic dependencies. Multilingual models consistently improve the NMT representations.
  }
\end{figure}

Figures \ref{fig:multilingual-morph}, \ref{fig:multilingual-syntax}, and \ref{fig:multilingual-semantics} show that the representations learned from multilingual models, despite sharing the encoder and decoder representations among 4 languages, can still effectively learn the same amount of morphology, syntax, and semantics as learned by their bilingual counterparts. In almost all cases, the multilingual representations are either better or at par with the bilingual models. Using a larger multilingual model (double the parameters size) gave consistent improvements in accuracy which resonate with the improvement in translation quality. 

Focusing on morphology (Figure \ref{fig:multilingual-morph}), a multilingual encoder generates representations that are better than its bilingual counterpart in most cases, while a multilingual decoder slightly degrades representation quality compared to the bilingual decoder. However, using a larger multilingual model leads to substantial improvements. These results mirror the patterns shown in terms of translation quality (Figure \ref{fig:multilingual-bleu}). 
In the case of syntactic and semantic dependencies (Figures \ref{fig:multilingual-syntax} and \ref{fig:multilingual-semantics}), even the default-sized multilingual model works better than the bilingual model, and the larger multilingual model leads to additional small improvements in syntactic dependencies and in Czech semantic dependencies (the Czech-English bars in Figure \ref{fig:multilingual-semantics}).  The larger multilingual model does not improve in the case of English semantic dependencies (English-Czech/German bars in Figure \ref{fig:multilingual-semantics}), even though the corresponding BLEU scores do improve  with a larger multilingual model (Figure \ref{fig:multilingual-bleu}).\footnote{One speculation for this might be that translating into several target languages does not add much semantic information on the source side because this kind of information is more language-agnostic, but at this point there is insufficient evidence for this kind of claim. %\alert{YB: this is very speculative but may be an interesting explanation}
}

%Again 
In an effort to probe whether increasing the number of parameters in a bilingual model would result in similar performance improvements, Table \ref{tab:rich-Czech} shows results across different properties in Czech$\leftrightarrow$English language pairs. We consistently found that increasing the model size does not lead to the same improvements as we observed in the case of multilingual models. These results reinforce that multilingual NMT models learn richer representations compared to the bilingual model and benefit from the shared properties across different language pairs.

% \begin{figure}
% \centering
% \begin{minipage}{.49\textwidth}
 
% \end{minipage}%
% \begin{minipage}{.49\textwidth}
	
% \end{minipage}
% \end{figure}

\section{Discussion} \label{sec:discussion}

In this section, we discuss the overall patterns that arise from the experimental results from several angles. First, we discuss how to assess the overall quality of the learned NMT representations with regards to other baselines and upper bounds. Second, we consider NMT representations from the perspective of contextualized word representations and contrast them to recent popular representations. 
Third, we reflect on the methodological approach taken in this work, and what it may or may not tell us about how the NMT model exploits language representations. Fourth, we briefly discuss the relation of our results to other NMT architectures. Finally, we touch upon the role of analysis work in understanding and improving NMT models. 

%\begin{itemize}
%    \item explanation, discussion, reflection (Reviewer C weakness 1)
%    \item How much information is there in the representations relative to baselines and upper bounds (Reviewer C weakness 2)
%    \item Relevance of the results to contextualized word representations (ELMo, BERT, CoVE). See also ICLR paper on edge probing. And Yonatan's NAACL paper on CWRs. (Reviewer C weakness 3). And our AAAI paper - where LM embedding did as good as MT in the case of POS/Morph but poorly in SEM task. 
    %\item Relevance to newer NMT architectures (Reviewer C required change)
    %\item Connections between multilingual experiments and more recent work on multilingual models, such as the multilingual BERT, and maybe \url{https://arxiv.org/pdf/1903.07091.pdf}
    %\item correlation vs causation; intervention experiments 
    %\item How insights can inform new models and improve SOTA
%\end{itemize}

\subsection{Assessing Representation Quality} 

The analyses presented in this work shed light on the quality of different language representations in NMT, with a particular focus on comparing various NMT components and design choices (layers, translation units, etc.). Our questions, therefore, have mostly been from a comparative perspective: how good are certain NMT representations compared to others with respect to certain linguistic tasks. 
But just how good are the NMT representations overall? Answers to this question may depend on the use case. One could, for example, evaluate the utility of NMT representations to improve state-of-the-art performance by plugging them as additional features in some strong model. Indeed, \citet{NIPS2017_7209} found this approach to yield state-of-the-art results in several language understanding tasks. 
It is also important to consider the quality of the NMT representations for the understudied tasks here in comparison to other baselines and competitive systems. Throughout the paper, we have compared the results to a majority baseline, arguing that NMT representations obtain substantial improvements.
Here we compare, for each linguistic task, the best performing NMT representations with  several baselines and upper bounds. 
We compare with the local majority baseline (most frequent tag/label for each word according to the training data, and the globally most frequent tag/label for words not seen in training) and with a classifier trained on word embeddings that are pre-trained on the source-side of the MT training data. We also train an encoder-decoder on converting text to tags, by automatically annotating the source side of the MT parallel data. Then we use this encoder-decoder to tag the test set of the supervised data and evaluate its quality. Finally, we generate representations from the encoder of this encoder-decoder model and train a classifier on them to predict the tags.  This setting aims to mimic our main scenario, except that we generate representations with an encoder-decoder specially trained on the linguistic task that we evaluate, rather than representations generated by an NMT model. 

%Figure~\ref{fig:nmt-baselines-bounds} 
Table \ref{tab:wordClassifier} shows the results.  A classifier trained on NMT representations performs far better than the majority baseline, as we have already confirmed. %\todo{YB: here would be good to compare with a classifier trained on word embeddings that are pre-trained on the source side of the MT data, ND: Lets do it for the camera-ready} 
A similar classifier trained on representations from a task-specific encoder-decoder performs even better. This indicates that training on a specific task leads to representations more geared towards that task, as may be expected. In fact, a similar behavior has been noted with other contextual word representations~\cite{liu:2019:NAACL}. 
Still, the representations do not contain all available information (or, not all information may be extracted by a simple classifier), as the task-specific encoder-decoder performs  better than a classifier trained on its representations.

%Interestingly, the NMT representations close \alert{much/little/half} of the gap between word embeddings and task-specific encoder representations. 
%\alert{YB: insert here a comparison of error reduction after we have all results. How much of the gap between a classifier trained on word embeddings and one trained on task-specific encoder representations is closed by the classifier trained on NMT representations} 

\subsection{Contextualized Word Representations} 
The representations generated by NMT models may be thought of as contextualized word representations (CWRs), as they capture context via the NMT encoder or decoder. We have already mentioned one work exploiting this idea, known as CoVE~\cite{NIPS2017_7209}, which used NMT representations as features in other models to perform various NLP tasks. 
Other prominent contextualizers include ELMo~\cite{Peters2018DeepCW}, which trains two separate, forward and backward LSTM language models (with a character CNN building block) and concatenates their representations across several layers; GPT~\cite{Radford2018IL} and GPT-2~\cite{radford2019language}, which use transformer language models based on self-attention~\cite{NIPS2017_7181}; and BERT~\cite{devlin2018bert}, which uses a bidirectional transformer model trained on masked language modeling (filling the blanks). 
All these generate representations that feed into task-specific classifiers, potentially with fine-tuning the contextualizer weights.\footnote{See \citet{peters2019tune} for an evaluation of when it is worth to fine-tune.}

How do NMT representations compare to CWRs trained from raw text? Directly answering this question is beyond the scope of this work, and is also tricky to perform for two reasons. First, CWRs like ELMo, BERT, and GPT require very large amounts of data, on the order of billions of words, which is far beyond what typical NMT systems are trained on (our largest NMT systems are trained on an order of magnitude fewer words). Second, at present, various CWRs are trained in incomparable settings in terms of data, number of parameters, and infrastructure. %Moreover, the code for their initial training is not always available. 
It seems that a comparison of common CWRs themselves is necessary before they are compared to NMT representations.

There is, however, indirect information that may tell us something about how CWRs trained on raw texts behave in comparison to NMT representations. 
\citet{bowman2018looking} compared sentence encoders trained on a variety of tasks, including the CoVE translation representations, and evaluated on language understanding tasks. They found language modeling pre-training to perform best, but cautioned that without fine-tuning, many of the results are not far above trivial baselines. They also found that grammar-related tasks benefit more from such pre-training than meaning-oriented tasks.
\citet{Peters2018DissectingCW} compared the ELMo LSTM with similar systems based on convolutions or transformer-style self-attention.  They found that all architectures learn hierarchical representations: the word embedding layer focuses on morphology, low encoding layers focus on local syntax, and high encoding layers carry more semantic information. 
These results are mostly in line with our findings concerning representation depth, although we have not noticed a clear separation between syntactic and semantic properties. 
\citet{Zhang2018LanguageMT} compared representations from NMT and bidirectional language models on POS tagging and CCG supertaggging. They found the language model representations to consistently outperform those from NMT. In other work, we have found that language model representations are of similar quality to NMT ones in terms of POS and morphology, but are behind in terms of semantic tagging~\cite{dalvi:2019:AAAI}.

\citet{Tenney2019What} compared representations from CoVE, ELMo, GPT, and BERT on a number of classiciation tasks, partially overlapping with the ones we study. 
They found that CWRs trained on raw texts outperform the MT representations of CoVE; however, as noted above these models are all trained in very different setups and cannot be fairly compared. 
Another interesting finding is that learning a weighted mix of layers works better than any one layer, and also better than concatenating. This again indicates that some layers are better than others for different tasks, consistent with our results. 
Concerning different tasks, \citet{Tenney2019What} found that CWRs are especially helpful (compared to a lexical baseline) with syntactic 
tasks, such as dependency and constituent labeling, and less helpful with certain semantic tasks like capturing fine-grained semantic attributes and pronoun resolution. They did notice improvements with semantic roles, which are related to our predicate-argument relations, where we also noticed significant improvements at higher layers. 
Finally, \citet{liu:2019:NAACL} compared ELMo, GPT, and BERT on various classification tasks in terms of their linguistic knowledge and transferability. They found that simple classifier trained on top of the (frozen) representations led to state-of-the-art results in many cases, but failed on tasks requiring fine-grained linguistic knowledge like conjunct identification. They observed that the first layer of LSTM-based CWRs performs better than other layers, while in transformer-based models the intermediate layers are the best. Considering different pre-training tasks, higher layer representations were more task-specific (and less general) in LSTM models, but not in  transformer models. In our investigation, the top layers of the (LSTM) NMT models were better for syntactic and semantic tasks. One possible explanation for this could be that translation is more aligned with the syntactic and semantic properties than language modeling. 

The development and analysis of CWRs is still ongoing. At present, NMT representations appear to be weaker than those obtained by contextualizers trained on raw texts, at least when the latter are trained on much larger amounts of data. It remains to be seen whether NMT representations can complement raw-text CWRs in certain scenarios. 

\subsection{On the Impact of Language Representation on Translation Output}

Our methodological approach evaluates whether various linguistic properties are decodable from learned NMT representations. Our assumption was that the quaity of a trained classifier can serve as a proxy to the quality of the original model, for a given task.
However, it is not clear whether the NMT model really ``cares'' about the linguistic properties, in the sense that it relies on them for performing the translation tasks. In essence, we only provide correlational evidence, not causal evidence. This is a limitation of much of the work using classification tasks to analyze neural networks,  as explained by \citet{belinkov2019analysis}. 
One avenue for addressing this question in causal terms is to define interventions: change something in the representation and test whether and how it impacts the output translation. 
\citet{indivdualneuron:iclr} perform such intervention experiments in NMT. They identify individual neurons that capture certain morphological properties---gender, number, and tense---and modify their activations. They evaluate how such intervention affects the output translations, finding that tense is fairly well modified, but gender and number are not as affected. 
Following similar ideas may be a fruitful area for further investigation of various linguistic properties and how much NMT systems depend on them when producing a translation.

\subsection{Why Analyze?} 
There are various motivations for work on interpretability and analysis of neural network models in NLP and other domains. There are also questions concerning  their necessity. While this article does not aim to solve this debate,\footnote{See \citet{belinkov2019analysis} and references therein for considerations in the context of NLP.}  we would like to highlight a few potential benefits of the analysis. 
First, several of our results may serve as guidelines for improving the quality of NMT systems and their utility for other tasks. The results on using different translation units suggest that their choice may depend on what properties one would like to capture. This may have implications for using MT systems in different languages (morphologically-rich vs.\ poor, free vs.\ fixed word order) or genre (short, simple sentences vs.\ long, complex ones). 
The results on representation depth suggest that using NMT representations for contextualization may benefit from combining layers, maybe with task-wise weighting. One could also imagine performing multi-task learning of MT and other tasks, with auxiliary losses integrated in different layers. 
The results on multilingual systems indicate that such systems may lead to better representations, but often require greater capacity. Inspecting language representations in a zero-shot MT scenario~\cite{TACL1081,arivazhagan2019missing} may also yield new insights for improving such systems. 

\subsection{Other NMT Architectures}
\label{sec:nmt-arch}
The NMT models analyzed in this work are all based on recurrent LSTM encoder-decoder models with attention. While this is the first successful NMT architecture, and still a dominant one, it is certainly not the only one. Other sucecssful architectures include fully convolutional~\cite{pmlr-v70-gehring17a} and fully-attentional, transformer encoder-decoder models \cite{NIPS2017_7181}.  There are also non-autoregressive models which are promising in terms of efficiency~\cite{gu2018nonautoregressive}.
At present, NMT systems based on transformer components appear to be the most successful. Combinations of transformer and recurrent components may also be helpful \cite{P18-1008}. 

The generalization of the particular results in this work to other architectures is a question of study. 
Recent efforts to analyze transformer-based NMT models include attempts to extract syntactic trees from self-attention weights~\cite{W18-5444,W18-5431} and evaluating representations from the transformer encoder~\cite{W18-5431}. The latter found that lower layers tend to focus on POS and shallow syntax, while higher layers are more focused on semantic tagging. These results are in line with our findings.  However, more work is needed to understand the linguistic representational power of various NMT architectures. We expect the questions themselves, and the methods, to remain  an active field  of investigation with newer architectures and systems.%\todo{YB: these two sentences may be moved to the conclusion} 

\section{Conclusion and Future Work}
\label{sec:conclusion}

In this article, we presented a comprehensive analysis of the representations learned during NMT training from the perspective of core linguistic phenomena, namely morphology, syntax, and semantics. We evaluated the representation quality on the tasks of morphological, syntactic and semantic tagging and using syntactic and semantic dependency labeling. Our results show that the representations learned during neural MT training learn a non-trivial amount of linguistic information.
We found that different properties are represented to varying extents in different components of the NMT models. The main insights are:
\begin{itemize}
\item Comparing representations at different layer depths, we found that word morphology is learned at the lower layer in the LSTM encoder-decoder model, whereas  non-local linguistic phenomena in syntax and semantics are better represented at the higher layers. For example, we found that higher layers better at predicting clause-level syntactic dependencies, or second and third semantic
arguments, in contrast to short-range dependencies which do not benefit much from higher layers.
%\todo{YB: may recap a bit more here}
%We analyzed and compared representations learned from different translation units. 
\item Comparing representations with different translation units, we found that representations learned using characters perform best at capturing word morphology, and therefore provide a more viable option when translating morphologically rich languages such as Czech. They are more robust towards handling unknown and low frequency words. 
\item In contrast, representations learned from subword units are better at capturing syntactic and semantic information that requires learning non-local dependencies. Character-based representations, on the other hand, are poor at handling long-range dependencies and therefore inferior when translating syntactically divergent language pairs such as German-English. %\todo{YB: say something about no free lunch here?} 
\item We found morpheme segmented units to give better representations than the ones learned using non-linguistic BPE units. The former outperformed the latter in most scenarios, even giving slightly better translation quality. 
\item We found that multilingual models benefit from the shared properties across different language pairs and learn richer representations compared to the bilingual model.

\end{itemize}

%\todo{YB: consider making this future work a call to others rather than saying that we intend to do this}
%In the future, we would like to expand the analysis in two directions. First, in terms of the studied linguistic properties, we would like to move beyond words and word relations to explore phrase and sentence structures.  Second, in terms of the studied NMT models, the current study focused on models based on LSTMs. We would like to explore other architectures such as Transformers \cite{NIPS2017_7181} which recently set a new state-of-the-art compared to both recurrent and convolutional models~\cite{pmlr-v70-gehring17a}.

Future work can expand the analysis into many directions. For instance, in terms of the studied linguistic properties, moving beyond words and relations to explore phrase and sentence structures could be an interesting frontier to explore. The current study focused on NMT models based on LSTMs. Analyzing other architectures such as Transformers \cite{NIPS2017_7181}, which recently set a new state-of-the-art compared to both recurrent and convolutional models~\cite{pmlr-v70-gehring17a}, would be an exciting direction to pursue.

\section*{Acknowledgements}
This work was funded by the QCRI, HBKU, as part of the collaboration with the MIT, CSAIL.
Yonatan Belinkov was also partly supported by the Harvard Mind, Brain, and Behavior Initiative.

\starttwocolumn
\bibliography{thesis,refs}
\onecolumnnew
\newpage
\appendix
\section{Supplementary Material}

\subsection{Character-based Models}
\label{sec:charCNN}

The character-based models reported in this paper were trained using bidirectional LSTM models only. We simply segmented words into characters and marked word boundaries. However, we did try charCNN \cite{kim2015character,costajussa-fonollosa:2016:P16-2} models in our preliminary experiments. The model is a CNN with a highway network over characters and trains an LSTM on top of it. In our results, we found the charCNN variant to perform poorly (see Table \ref{tab:bleu-scores}), compared to the simple char-based LSTM model both in translation quality and comparing classifier accuracy. We therefore left it out and focused on char-based LSTM model.. 

 \begin{table*}[h]
 	\centering
 	%\footnotesize
 	\begin{tabular}{l c c c c}
 		\toprule
 		 & \multicolumn{2}{c}{BLEU} & \multicolumn{2}{c}{Accuracy} \\
 		\cmidrule(lr){2-3} \cmidrule(lr){4-5}
 		& de-en & cs-en & de-en & cs-en \\
 	    \midrule
 		char $\rightarrow$ bpe & 34.9 & 29.0 & 79.3 & 81.7 \\
 		charCNN $\rightarrow$ bpe &32.3 & 28.0 & 79.0 & 79.9 \\
 		\bottomrule
 	\end{tabular}
 	    \caption{BLEU scores and morphological classifier accuracy across language pairs, comparing (encoders of) fully character-based LSTM and charCNN LSTM models.}
 	\label{tab:bleu-scores}
\end{table*}

\subsection{Effect of Target Language}
\label{sec:targetLang}

Our results showed that the representations that are learned when translating into English are better for predicting POS or morphology than those learned when translating into German, which are in turn better than
those learned when translating into Hebrew. The inherent difficulty in translating Arabic to (morphologically rich) Hebrew/German languages may affect the ability to learn good representations of word structure, or perhaps more data is required in the case of these languages to learn Arabic representations of
the same quality. We found these results to be
consistent in other language pairs, i.e., by changing the source from Arabic to German and Czech and when training character-based models instead of word-based models. See Tables \ref{tab:target} and \ref{tab:different_language} for these results.

\begin{table*}[h]
\centering
\begin{tabular}{l l cccc}
\toprule
& & ar & he & de & en \\ 
\midrule 
\multirow{3}{*}{Word} & POS & 67.21 & 78.13 & 78.85 & 80.21 \\
& Morphology & 55.63 & 64.87 & 65.91 & 67.18 \\ 
& BLEU & 80.43 & 9.51 & 11.49 & 23.8 \\ 
\midrule 
\multirow{3}{*}{Char} & POS & 87.72 & 92.67 & 93.05 & 93.63 \\ 
& Morphology & 75.21 & 80.50 & 80.61 & 81.49 \\ 
& BLEU & 75.48 & 11.15 & 12.86 & 27.82 \\ 
\bottomrule 
\end{tabular}
\caption{Effect of changing the target language on POS and morphological tagging with classifiers trained on the encoder side of both word-based and character-based (here: charCNN) models. The source language, for which classification is done, is always Arabic.} 
\label{tab:target}
\end{table*}

 \begin{table}[h]
%\footnotesize
\centering
\begin{tabular}{l rrr}
\toprule
 %& \multicolumn{3}{c}{POS Tagging Accuracy}  \\
& \multicolumn{3}{c}{Target language}  \\
%\midrule
%\backslashbox{Source}{Target} &
Source language  & 
\multicolumn{1}{c}{English} & \multicolumn{1}{c}{Arabic} & \multicolumn{1}{c}{Self} \\
\cmidrule(lr){1-1} \cmidrule(lr){2-4}
German & 93.5 & 92.7 & 89.3 \\
Czech & 75.7 & 75.2 & 71.8  \\
\bottomrule
\end{tabular}
\caption{Impact of changing the target language on POS tagging accuracy with classifiers trained on the encoder side. 
%Studying the impact of changing target language, 
%POS accuracy (Layer 2), 
Self = German/Czech in rows 1/2 respectively.}
\label{tab:different_language}
\end{table}

\subsection{Three Layered Character-based Models}

In order to probe whether character-based models require additional depth in the network to capture the same amount of information, we carried out further experiments training 3-layered character models for Czech-to-English and English-to-German. We extracted feature representations and trained classifiers to predict syntactic dependency labels. Our results show that using an additional layer does improve the prediction accuracy, giving the same result as subword segmentation (Morfessor) in the case of Czech-to-English, but still worse in the case of English-to-Czech (see Table \ref{tab:threeLayer}).

 \begin{table*}[h]
 	\centering
 	%\footnotesize
 	\begin{tabular}{l r r r r r}
 		\toprule
 	 & \multicolumn{2}{c}{cs-en} & \multicolumn{3}{c}{en-de}  \\
 	    \cmidrule(lr){2-3} \cmidrule(lr){4-6}
 	 & Syn Dep & Sem Dep & Syn Dep & Sem Dep  & Sem tags  \\
 	 \midrule
 		char (layer 2) & 89.3 & 84.3 & 90.3 & 78.9 & 92.3  \\
 		char (layer 3)  & 90.2 & 85.2 & 91.1 & 79.6 & 92.7  \\
 		best subword  & 90.3 & 86.3 & 91.4 & 80.4 & 93.2  \\
 		\bottomrule
 	\end{tabular}
 	    \caption{Results on syntactic and semantic tagging and labeling with representations obtained from char-based models trained with an extra layer.}
 	\label{tab:threeLayer}
\end{table*}

\subsection{Layer-wise Experiments Using CCG Tags}

Along with the syntactic dependency labeling task, we found higher layers to give better classifier accuracy also in the CCG tagging task. See Table \ref{tab:results-ccgtags-4layers} for the results. 

\begin{table}[h]
\centering
\begin{tabular}{l llll}
\toprule
$k$ & \multicolumn{1}{c}{de} & \multicolumn{1}{c}{cs} & \multicolumn{1}{c}{fr} & \multicolumn{1}{c}{es} \\ 
\midrule
1 & 88.15 & 84.90 & 87.70 & 87.55 \\
2 & 88.70 & 86.20  & 88.60 &  88.10\\
3 & 88.80 & 86.60 & 88.10 & 88.35 \\
4 & 89.50 & 85.10 & 88.80 & 88.90 \\
\midrule
All & 91.60 & 89.90 & 91.30 &  91.20\\
\bottomrule
\end{tabular}
\caption{CCG tagging accuracy 
using features 
from the $k$-th encoding layer 
of 4-layered English-to-* NMT models trained with different target languages (German, Czech, French, and Spanish).}
\label{tab:results-ccgtags-4layers}
\end{table}

\subsection{Statistical Significance Results}
\label{sec:significanceTests}
Table~\ref{tab:significanceTests} shows statistical significance results for syntactic dependency labeling experiments from Section~\ref{sec:syntax-depth}. 

\begin{table}[h]
\footnotesize
\centering
\begin{tabular}{l | lllll | lllll | lllll}
\toprule
& \multicolumn{5}{c}{English-Arabic} & \multicolumn{5}{c}{English-Spanish} & \multicolumn{5}{c}{English-French}  \\
\cmidrule(lr){2-6} \cmidrule(lr){7-11} \cmidrule(lr){12-16}
$k$ & 0 &  1 & 2 &  3  & 4 & 0 & 1 &  2  & 3  & 4 & 0 & 1 &  2  & 3  & 4 \\
\midrule
0 &   & \ddag &  &   &   &  & \ddag &    &   &  &  & \ddag &    &   &  \\
1 & \ddag  &  & \ddag & \ddag  & \ddag  & \ddag &  &  \ddag  & \ddag  & \ddag & \ddag &  & \ddag   &  \ddag & \ddag \\
2 &   & \ddag &  & ns  & \ddag  &  & ns &    & \ddag  & \ddag &  & \dag &   &  \ddag & \ddag \\
3 &   & \ddag & ns &   & \ddag  &  & \ddag & \ddag    &   & \ddag &  & \ddag & \ddag   &   &  \ddag \\
4 &   & \ddag & \dag & ns  &   &  & \ddag & \ddag   & \ddag  &  &  & \ddag & \ddag    & \ddag  &  \\
\midrule
& \multicolumn{5}{c}{English-Russian} & \multicolumn{5}{c}{English-English} &   \\
\midrule
$k$ & 0 &  1 & 2 &  3  & 4 & 0 & 1 &  2  & 3  & 4  %& 0 & 1 &  2  & 3  & 4 
\\
\midrule
0 &   & \ddag &  &   &   &  & \ddag &    &   &  %& - & - & -  &  - & - 
\\
1 & \ddag  &  & \dag & \ddag  &  \ddag & \ddag &  & \dag   & \dag  & \ddag %& - & - & -   & -  & - 
\\
2 &   & \ddag &  & \ddag  & \ddag  &  & \dag &  & ns  & \ddag %& - & - & -  &  - & - 
\\
3 &   & \ddag & ns &   & \dag  &  & \dag & ns   &   & \ddag %& - & - & -  &  - & - 
\\
4 &   & \ddag & \ddag & \ddag  &   &  & \ddag &  \ddag  & \ddag  &  %& - & - & -  &  - & - 
\\
\bottomrule
\end{tabular}
\caption{Statistical significance results for syntactic dependency labeling from Section 7.2. The cells above the main diagonal are for the translation direction A$\rightarrow$B and below it are for the direction B$\rightarrow$A. $ns = p > 0.05, \dag = p < 0.01, \ddag = p < 0.001$. Comparisons at empty cells are not shown.}
\label{tab:significanceTests}
\end{table}

%\subsection{Distributedness of Open Class Categories}

%Table \ref{tab:open-class-percentage} shows the distribution of neurons that are responsible for a few tags. We can see that for several closed class categories (\texttt{TO}, months of year: \texttt{MOY}), the distribution is heavily skewed towards a single layer, while for open class categories (verbs: \texttt{VB}, \texttt{NNS}: singular nouns, etc.), the distribution is less skewed towards a particular layer.
%\begin{table}[]
%	\centering
%	\begin{tabular}{l r r}
%		\toprule
%		Tag  & Layer 1    & Layer 2 \\
%		\toprule
%		TO   &   100.00\%  &    0.00\% \\
%		ORG  &   100.00\%  &    0.00\% \\
%		LOC  &   100.00\%  &    0.00\% \\
%		MOY  &    80.00\%  &   20.00\% \\
%		\midrule
%		VBZ  &    87.50\%  &   12.50\% \\
%		VBD  &    83.78\%  &   16.22\% \\
%		EXS  &    75.00\%  &   25.00\% \\
%		VB   &    66.67\%  &   33.33\% \\
%		PST  &    50.00\%  &   50.00\% \\
%		NNS  &    52.94\%  &   47.06\% \\
%		NN   &    25.00\%  &   75.00\% \\
%		\bottomrule
%	\end{tabular}
%	\caption{Percentage of neurons responsible for some tags in layer 1 and 2 in a two layer NMT model.}
%	\label{tab:open-class-percentage}
%\end{table}
\end{document}